\documentclass{article} 
\usepackage{nips15submit_e,times}
\usepackage{hyperref}
\usepackage{url}

\usepackage{amsmath}
\usepackage{amsfonts}
\usepackage{amsthm}
\usepackage{amssymb}
\usepackage{dsfont}
\usepackage{algorithm}
\usepackage{algorithmic}
\usepackage{graphicx} 
\usepackage{subfigure}
\usepackage[english]{babel}  

\DeclareMathOperator*{\argmin}{arg\,min}
\DeclareMathOperator*{\argmax}{arg\,max}

\newcommand{\bs}[1]{\boldsymbol{#1}}
\allowdisplaybreaks[2]
\newtheorem{theorem}{Theorem}

\newtheorem{lemma}{Lemma}

\title{Algorithms with Logarithmic or Sublinear Regret for\\
 Constrained Contextual Bandits}

\author{
Huasen Wu \\
University of California at Davis\\
\texttt{hswu@ucdavis.edu}
\And
R.~Srikant \\
University of Illinois at Urbana-Champaign \\
\texttt{rsrikant@illinois.edu}
\And
Xin Liu \\
University of California at Davis\\
\texttt{liu@cs.ucdavis.edu}
\And
Chong Jiang \\
University of Illinois at Urbana-Champaign\\
\texttt{jiang17@illinois.edu}
}

\nipsfinalcopy 

\begin{document}

\maketitle
\vspace{-0.5cm}
\begin{abstract}
We study contextual bandits with budget and time constraints, referred to as \emph{constrained contextual bandits}.
The time and budget constraints significantly complicate the exploration and exploitation tradeoff because they introduce complex coupling among contexts over time.
To gain insight, we first study unit-cost systems with known context distribution. When the expected rewards are known,
we develop an approximation of the oracle, referred to Adaptive-Linear-Programming (ALP), which achieves near-optimality and only requires the ordering of expected rewards. With these highly desirable features,  we  then combine ALP with the upper-confidence-bound (UCB) method in the general case where the expected rewards are unknown {\it a priori}. We show that the proposed UCB-ALP algorithm achieves logarithmic regret except for certain boundary cases.
Further, we design algorithms and obtain similar regret bounds for  more general systems with unknown context distribution and heterogeneous costs.  To the best of our knowledge, this is the  first work that shows how to achieve logarithmic regret in constrained contextual bandits. 
Moreover, this work also sheds light on the study of computationally efficient algorithms for general constrained contextual bandits.
\end{abstract}


\section{Introduction} \label{sec:intro}
The contextual bandit problem \cite{Langford2007NIPS,Lu2010ICAIS:CMAB,Zhou2015CMAB:Survey} is an important extension of the classic multi-armed bandit (MAB) problem \cite{Auer2002ML:UCB}, where  the agent can observe a set of features, referred to as \emph{context}, before making a decision. 
After the random arrival of a context, the agent chooses an action and receives a random reward with expectation depending on both the context and action.
To maximize the total reward, {the agent needs to make a careful tradeoff between taking the best action based on the historical performance (exploitation) and discovering the potentially better alternative actions under a given context (exploration).}
This model has attracted much attention as it fits the personalized service requirement in many applications such as clinical trials, online recommendation, and online hiring in crowdsourcing.
Existing works try to reduce  the regret of contextual bandits by leveraging the structure of the context-reward models such as linearity \cite{Li2010WWW:LinUCB} or similarity \cite{Slivkins2014JMLR:ContMAB}, and more recent work \cite{Agarwal2014ICML:CMAB} focuses on computationally efficient algorithms with minimum regret. For Markovian context arrivals, algorithms such as UCRL \cite{Auer2007UCB4RL} for more general reinforcement learning problem can be used to achieve logarithmic regret.

However, traditional contextual bandit models do not capture an important characteristic of real systems: in addition to time, there is usually a cost associated with the  resource consumed by each action and the total cost is limited by a budget in many applications. Taking crowdsourcing \cite{Badanidiyuru2012EC:OnlineProcurement} as an example, the budget constraint for a given set of tasks will limit the number of workers that an employer can hire. Another example is the clinical trials \cite{Lai2012SA}, where each treatment is usually costly and the budget of a trial is limited.
Although budget constraints have been studied in non-contextual bandits where logarithmic or sublinear regret is achieved \cite{Tran2012AAAI:MAB_BF,Badanidiyuru2013FOCS,Jiang2013CDC,Slivkins2013TR,Qin&Liu2015IJCAI:TS,Combes&Srikant2015Sigmetrics}, as we will see later, these results are inapplicable in the case with observable contexts.

In this paper, we study contextual bandit problems with budget and time constraints, referred to as \emph{constrained contextual bandits}, where  the agent is given a budget $B$ and a time-horizon $T$.
In addition to a reward, a cost is incurred  whenever an action is taken under a context. The bandit process ends when the agent runs out of either budget or time.
The objective of the agent is to maximize the expected total reward subject to the budget and time constraints. We are interested in the regime where $B$ and $T$ grow towards infinity proportionally.

The above constrained contextual bandit problem can be viewed as a special case of Resourceful Contextual Bandits (RCB)
\cite{Badanidiyuru2014COLT}. In \cite{Badanidiyuru2014COLT}, RCB is studied under more general settings with possibly infinite contexts, random costs, and multiple budget constraints.
A Mixture\_Elimination algorithm is proposed and shown to achieve  $O(\sqrt{T})$ regret.
However, the benchmark for the definition of regret in \cite{Badanidiyuru2014COLT} is restricted to within a finite policy set. Moreover, the Mixture\_Elimination algorithm suffers high complexity  and the design of computationally efficient algorithms for such general settings is still an open problem.

To tackle this problem,  motivated by certain applications, we restrict the set of parameters in our model as follows: we assume finite discrete contexts, fixed costs, and a single budget constraint. This simplified model is justified  in many scenarios such as clinical trials \cite{Lai2012SA} and rate selection in wireless networks \cite{Combes2014Infocom}.
More importantly, these simplifications allow us to design easily-implementable algorithms that achieve $O(\log T)$ regret (except for a set of parameters of zero Lebesgue measure, which we refer to as boundary cases), where the regret is defined more naturally as the performance gap between the proposed algorithm and \emph{the oracle}, i.e., the optimal algorithm with known statistics.

Even with simplified assumptions considered in this paper, the exploration-exploitation tradeoff is still challenging
due to the budget and time constraints. The key  challenge comes from the complexity of the oracle algorithm.
With budget and time constraints, the oracle algorithm cannot simply take the action that maximizes the instantaneous reward. In contrast, it needs to balance between the instantaneous and long-term rewards based on the current context and the remaining budget.
In principle, dynamic programming (DP) can be used to obtain this balance. However, using DP in our scenario incurs difficulties in both algorithm design and analysis: first, the implementation of DP is computationally complex due to the curse of dimensionality; second, it is difficult to obtain a benchmark for regret analysis, since the DP algorithm is implemented in a recursive manner and its expected total reward is hard to be expressed in a closed form; third, it is difficult to extend the DP algorithm to the case with unknown statistics, due to the difficulty of evaluating the impact of estimation errors on the performance of DP-type algorithms.

To address these difficulties, we first study approximations of the oracle algorithm when the system statistics are known. Our key idea is to approximate the oracle algorithm with linear programming (LP)  that relaxes the hard budget constraint to an average budget constraint. When fixing the average budget constraint at $B/T$, this LP approximation provides an upper bound on the expected total reward, which serves as a good benchmark in regret analysis. Further, we propose an Adaptive Linear Programming (ALP) algorithm that adjusts the budget constraint to the \emph{average remaining budget} $b_\tau/\tau$, where $\tau$ is the remaining time and $b_\tau$ is the remaining budget. Note that although the idea of approximating a DP problem with an LP problem has been widely studied in literature (e.g.,  \cite{Badanidiyuru2014COLT,Veatch2013MOR:ALP}), the design and analysis of ALP here is quite different. In particular, we show that ALP achieves $O(1)$ regret, i.e., its  expected total reward is within a constant independent of $T$ from the optimum, except for certain boundaries. This ALP approximation and its regret analysis make an important step towards achieving logarithmic regret for constrained contextual bandits.


Using the insights from the case with known statistics, we study algorithms for constrained contextual bandits with unknown expected rewards. Complicated interactions between information acquisition and decision making arise in this case. Fortunately,  the ALP algorithm has a highly desirable property that it only requires the ordering of the expected rewards and can tolerate certain estimation errors of system parameters.
This property allows us to combine ALP with estimation methods that can efficiently provide a correct rank of the expected rewards. In this paper, we propose a UCB-ALP algorithm by combining ALP with
the upper-confidence-bound (UCB) method \cite{Auer2002ML:UCB}. We show that  UCB-ALP achieves $O(\log T)$ regret except for certain boundary cases, where its regret is $O(\sqrt{T})$.
We note that UCB-type algorithms are proposed in \cite{Agrawal2014EC} for non-contextual bandits with concave rewards and convex constraints, and further extended to linear contextual bandits.
However, \cite{Agrawal2014EC} focuses on static contexts\footnote{After the online publication of our preliminary version, two recent papers \cite{Agrawal2015TR:CB,Agrawal2015TR:LCB} extend their previous work \cite{Agrawal2014EC} to the dynamic context case, where they focus on possibly infinite contexts and achieve $O(\sqrt{T})$ regret, and  \cite{Agrawal2015TR:CB} restricts to a finite policy set as \cite{Badanidiyuru2014COLT}.} and achieves $O(\sqrt{T})$ regret in our setting since it uses a fixed budget constraint in each round.
In comparison, we consider random context arrivals and use an adaptive budget constraint to achieve logarithmic regret. To the best of our knowledge, this is the first work that shows how to achieve logarithmic regret in constrained contextual bandits. Moreover, the proposed UCB-ALP algorithm is quite computationally efficient and we believe these results shed light on addressing the open problem of general constrained contextual bandits.

Although the intuition behind  ALP and UCB-ALP  is natural, the rigorous analysis of their regret is non-trivial since we need to consider many interacting factors such as action/context ranking errors,  remaining budget fluctuation, and randomness of context arrival.
We evaluate the  impact of these factors using a series of novel techniques, e.g., the method of showing concentration properties under adaptive algorithms and the method of bounding estimation errors under random contexts.
For the ease of exposition, we study the ALP and UCB-ALP algorithms in unit-cost systems with known context distribution in Sections~\ref{sec:oracle_solution} and \ref{sec:ucb_dp}, respectively. Then we discuss the generalization to systems with unknown context distribution in Section~\ref{sec:unknown_dist} and with heterogeneous costs in Section~\ref{sec:het_cost}, which are much more challenging and the details can be found in the supplementary material.

\section{System Model}
We consider a contextual bandit problem with a context set $\mathcal{X} = \{1,2,\ldots, J\}$ and an action set $\mathcal{A} = \{1,2,\ldots, K\}$. At each round $t$, a context $X_t$ arrives independently with identical distribution $\mathbb{P}\{X_t = j\} = \pi_j$, $j \in \mathcal{X}$, and each action $k \in \mathcal{A}$ generates a non-negative reward $Y_{k, t}$.
Under a given context $X_t = j$, the reward $Y_{k,t}$'s are independent random variables in $[0,1]$. The conditional expectation $\mathbb{E}[Y_{k,t}|X_t = j] = u_{j,k}$ is unknown to the agent. Moreover, a cost is incurred if action $k$ is taken under context $j$.
To gain insight into constrained contextual bandits, we consider fixed and known costs in this paper, where the cost is $c_{j,k} > 0$ when action $k$ is taken under context $j$.
Similar to traditional contextual bandits, the context $X_t$ is observable  at the beginning of round $t$, while only the reward of the action taken by the agent is revealed at the end of round $t$.

At the beginning of round $t$, the agent observes the context $X_t$ and takes an action $A_t$ from $\{0\}\cup \mathcal{A}$, where ``$0$'' represents a {\it dummy} action that the agent skips the current context. Let $Y_t$ and $Z_t$ be the reward and cost for the agent in round $t$, respectively. If the agent takes an action $A_t = k > 0$, then the reward is $Y_t = Y_{k,t}$ and the cost is $Z_t = c_{X_t, k}$. Otherwise, if the agent takes the dummy action $A_t = 0$, neither  reward nor cost is incurred, i.e., $Y_t = 0$ and $Z_t = 0$. In this paper, we focus on contextual bandits with a known time-horizon $T$ and limited budget $B$. The bandit process ends when the agent runs out of the budget or at the end of time $T$.

A contextual bandit algorithm $\Gamma$ is a function that maps the historical observations $\mathcal{H}_{t-1} = (X_1, A_1, Y_1$; $X_2, A_2, Y_2$; $\ldots; X_{t-1}, A_{t-1}, Y_{t-1})$ and the current context $X_t$ to an action $A_t \in \{0\}\cup \mathcal{A}$.
The objective of the algorithm is to maximize the expected total reward $U_{\Gamma}(T, B)$ for a given time-horizon $T$ and a budget $B$, i.e.,
\begin{eqnarray}
\text{maximize}_{\Gamma}&&  U_{\Gamma}(T, B) = \mathbb{E}_{\Gamma}\big[\sum_{t = 1}^T Y_t\big] \nonumber \\
\text{subject to~~} && \sum_{t = 1}^T Z_t \leq B,\nonumber
\end{eqnarray}
where the expectation is taken over the distributions of contexts and rewards. Note that we consider a ``hard'' budget constraint, i.e., the total costs should not be greater than $B$ under any realization.

We measure the performance of the algorithm $\Gamma$ by comparing it with the  oracle, which is the optimal algorithm with known statistics, including the knowledge of $\pi_j$'s, $u_{j,k}$'s, and $c_{j,k}$'s. Let $U^*(T,B)$ be the expected total reward obtained by the oracle algorithm. Then, the regret of the algorithm $\Gamma$ is defined as
\begin{eqnarray}
R_{\Gamma}(T, B) = U^*(T,B) - U_{\Gamma}(T, B). \nonumber
\end{eqnarray}
The objective of the algorithm is then to minimize the regret. We are interested in the asymptotic regime where the time-horizon $T$ and the budget $B$ grow to infinity proportionally, i.e., with a fixed ratio $\rho = B/T$.

\section{Approximations of the Oracle} \label{sec:oracle_solution}
In this section, we study approximations of the oracle, where the statistics of bandits are known to the agent. This will provide a benchmark for the regret analysis and insights into the design of constrained contextual bandit algorithms.

As a starting point, we focus on unit-cost systems, i.e., $c_{j,k} = 1$ for each $j$ and $k$, from Section~\ref{sec:oracle_solution} to  Section~\ref{sec:unknown_dist}, which will be relaxed in Section \ref{sec:het_cost}.
In unit-cost systems, the quality of action $k$ under context $j$ is fully captured by its expected reward $u_{j,k}$. Let $u_j^*$ be the highest expected reward under context $j$, and  $k_j^*$ be the best
action for context $j$, i.e.,
$u_j^* = \max_{k \in \mathcal{A}} u_{j,k}$ and $k_j^* = \argmax_{k \in \mathcal{A}} u_{j,k}$.
For ease of exposition, we assume that the best action under each context is unique, i.e., $u_{j,k} < u_j^*$ for all $j$ and $k \neq k_j^*$.
Similarly, we also assume  $u_1^* > u_2^*  > \ldots > u_J^*$ for simplicity.

With the knowledge of $u_{j,k}$'s, the agent  knows the best action $k_j^*$ and its expected reward $u_j^*$ under any context $j$.
In each round $t$, the task of the oracle is deciding whether to take  action $k_{X_t}^*$ or not depending on  the remaining time $\tau = T- t + 1$  and the remaining  budget $b_\tau$.

The special case of two-context systems ($J = 2$) is trivial, where the agent just needs to procrastinate for the better context (see Appendix~\ref{app:two_contex} of the supplementary material).
When considering more general cases with $J > 2$, however, it is computationally intractable to exactly characterize the oracle solution.
Therefore, we resort to approximations  based on linear programming (LP).

\subsection{Upper Bound: Static Linear Programming}
We propose an upper bound for the expected total reward $U^*(T,B)$ of the oracle by relaxing the hard constraint to an average constraint
and solving the corresponding constrained LP problem. Specifically, let $p_j \in [0,1]$ be the probability that the agent takes action $k_j^*$ for context $j$,
and $1-p_j$ be the probability that the agent skips context $j$ (i.e., taking action $A_t = 0$).
Denote the probability vector as $\bs{p} = (p_1, p_2, \ldots, p_J)$. For a time-horizon $T$
and budget $B$, consider the following LP problem:
\vspace{-0.1cm}
\begin{eqnarray}
(\mathcal{LP}_{T,B})~~\text{maximize}_{\boldsymbol{p}} && \sum_{j = 1}^J p_j \pi_j u_j^*,  \label{eq:staticLP_obj} \\
\text{subject to~~} && \sum_{j=1}^J p_j \pi_j \leq B/T, \label{eq:staticLP_const}\\
&&\bs{p} \in [0,1]^J. \nonumber
\end{eqnarray}
Define the following threshold as a function of the average budget $\rho = B/T$:
\vspace{-0.1cm}
\begin{eqnarray} \label{eq:context_threhold}
\tilde{j}(\rho) = \max\{j: \sum_{j' = 1}^j\pi_{j'} \leq \rho\}
\end{eqnarray}
with the convention that $\tilde{j}(\rho) = 0$ if $\pi_1 > \rho$.
We can verify that the following solution is optimal for $\mathcal{LP}_{T,B}$:
\begin{eqnarray}\label{eq:LP_solution}
p_j(\rho) =
\begin{cases}
1, &\text{if $1 \leq j \leq  \tilde{j}(\rho)$},\\
\frac{\rho - \sum_{j' = 1}^{\tilde{j}(\rho)}\pi_{j'}}{\pi_{\tilde{j}(\rho)+1}}, & \text{if $j = \tilde{j}(\rho)+1$},\\
0, & \text{if $j > \tilde{j}(\rho)+1$}.\\
\end{cases}
\end{eqnarray}
Correspondingly, the optimal value of  $\mathcal{LP}_{T,B}$ is
\begin{eqnarray}\label{eq:LP_single_step_value}
{v}(\rho) = \sum_{j =1}^{ \tilde{j}(\rho)} \pi_j u_j^*
+ p_{\tilde{j}(\rho) + 1}(\rho) \pi_{\tilde{j}(\rho) + 1} u_{\tilde{j}(\rho) + 1}^*.
\end{eqnarray}
This optimal value ${v}(\rho)$ can be viewed as the maximum expected reward in a single round with average budget $\rho$.
Summing over the entire horizon, the total expected reward becomes $\widehat{U}(T,B) = T{v}(\rho)$, which is an upper bound of $U^*(T,B)$.
\begin{lemma}\label{thm:upper_bound}
For a unit-cost system with known statistics, if the time-horizon is $T$ and the budget is $B$,
then  $\widehat{U}(T,B) \geq U^*(T,B)$.
\end{lemma}

The proof of Lemma~\ref{thm:upper_bound} is available in Appendix~\ref{app:proof_upper_bound} of the supplementary material. With Lemma~\ref{thm:upper_bound}, we can bound the regret of any algorithm by comparing its performance
with the upper bound $\widehat{U}(T,B)$  instead of $U^*(T,B)$.
Since $\widehat{U}(T,B)$ has a simple expression,
as we will see later, it significantly reduces the complexity of regret analysis.

\subsection{Adaptive Linear Programming} \label{subsec:adaptive_lp}
Although the solution \eqref{eq:LP_solution} provides an upper bound on the expected reward,
using such a fixed algorithm will not achieve good performance as the ratio $b_{\tau}/\tau$, referred to as {\emph{average remaining budget}},
 fluctuates over time.
We propose an Adaptive Linear Programming (ALP) algorithm that adjusts the threshold and randomization probability according to the
instantaneous value of $b_{\tau}/\tau$.

Specifically, when the remaining time is $\tau$ and the remaining budget is $b_\tau = b$, we consider
an LP problem $\mathcal{LP}_{\tau, b}$ which is the same as $\mathcal{LP}_{T,B}$ except that
$B/T$ in Eq.~\eqref{eq:staticLP_const} is replaced with $b/\tau$.
Then, the optimal solution for $\mathcal{LP}_{\tau,b}$ can be obtained by replacing $\rho$ in Eqs.~\eqref{eq:context_threhold},
 \eqref{eq:LP_solution}, and \eqref{eq:LP_single_step_value} with $b/\tau$. The ALP algorithm then makes decisions based on this optimal solution.

\textbf{ALP Algorithm:} At each round $t$ with remaining budget $b_{\tau} = b$, obtain $p_j(b/\tau)$'s by solving $\mathcal{LP}_{\tau, b}$; take action $A_t = k^*_{X_t}$ with probability $p_{X_t}(b/\tau)$, and $A_t = 0$ with probability $1 - p_{X_t}(b/\tau)$.

The above ALP algorithm only requires the ordering of the expected rewards instead of their accurate values. This highly desirable feature allows us to combine ALP with classic MAB algorithms such as UCB \cite{Auer2002ML:UCB} for the case without knowledge of expected rewards. Moreover, this simple ALP algorithm achieves very good performance within a constant distance from the optimum, i.e., $O(1)$ regret, except for certain boundary cases. Specifically, for $1 \leq j \leq J$, let $q_j$ be the cumulative probability defined as
$q_j = \sum_{j'=1}^j \pi_{j'}$ with the convention that $q_{0} = 0$.
The following theorem states  the near optimality of ALP.
\begin{theorem} \label{thm:alp_regret}
Given any fixed $\rho \in (0,1)$, the regret of ALP satisfies:\\
1) (Non-boundary cases) if $\rho \neq q_j$ for any $j \in \{ 1,2,\ldots, J-1\}$, then $R_{\rm ALP}(T,B) \leq \frac{u_1^* - u_J^*}{1 - e^{-2\delta^2}}$,
where $\delta = \min\{\rho - q_{\tilde{j}(\rho)},
q_{\tilde{j}(\rho)+1} -\rho\}$.\\
2) (Boundary cases) if $\rho = q_j$ for some $j \in \{1,2,\ldots, J-1\}$, then $R_{\rm ALP}(T,B) \leq \Theta^{\rm (o)} \sqrt{T} + \frac{u_1^* - u_J^*}{1 - e^{-2(\delta')^2}}$, where $\Theta^{\rm (o)} =  2(u_1^* - u_J^*) \sqrt{\rho (1-\rho)}$ and $\delta' = \min\{\rho - q_{\tilde{j}(\rho) - 1}, q_{\tilde{j}(\rho) +1} - \rho\}$.
\end{theorem}

Theorem~\ref{thm:alp_regret} shows that ALP achieves $O(1)$ regret except for certain boundary cases, where it still achieves $O(\sqrt{T})$ regret. This implies that the regret due to the linear relaxation is negligible in most cases. Thus, when the expected rewards are unknown, we can achieve low regret, e.g., logarithmic regret, by combining ALP with appropriate information-acquisition mechanisms.

{\it \textbf{Sketch of Proof:}} Although the ALP algorithm seems fairly intuitive, its regret analysis is non-trivial. The key to the proof is to analyze the evolution of the remaining budget $b_\tau$ by mapping ALP to ``sampling without replacement''.
Specifically, from Eq.~\eqref{eq:LP_solution}, we can verify that when the remaining time is $\tau$ and the remaining budget is $b_{\tau} = b$,
the system consumes one unit of budget with probability $b/\tau$, and consumes nothing with probability $1-b/\tau$.
When considering the remaining budget, the ALP algorithm can be viewed as  ``sampling  without replacement''.
Thus, we can show that $b_\tau$ follows the hypergeometric distribution \cite{Dubhashi2009Concentration} and has the following properties:
\begin{lemma}\label{thm:alp_to_hypergeo}
Under the ALP algorithm, the remaining budget $b_{\tau}$ satisfies:\\
1) The expectation and variance of  $b_{\tau}$ are $\mathbb{E}[b_{\tau}] = \rho \tau$
and  ${\rm Var}(b_{\tau}) = \frac{T-\tau}{T-1}\tau\rho(1-\rho)$, respectively.\\
2) For any positive  number $\delta$ satisfying $0 < \delta <  \min\{\rho, 1-\rho\}$, the tail distribution of $b_{\tau}$ satisfies
\begin{eqnarray}
\mathbb{P}\{b_{\tau} < (\rho - \delta) \tau\} \leq e^{-2\delta^2 \tau}~~\text{and}~~
\mathbb{P}\{b_{\tau} > (\rho + \delta) \tau\} \leq e^{-2\delta^2 \tau}. \nonumber
\end{eqnarray}
\end{lemma}


Then, we prove Theorem~\ref{thm:alp_regret} based on Lemma~\ref{thm:alp_to_hypergeo}. Note that the expected total reward under ALP is $U_{\rm ALP}(T,B) = \mathbb{E}\big[\sum_{\tau = 1}^{T}v(b_{\tau}/\tau)\big]$,
where $v(\cdot)$ is defined in \eqref{eq:LP_single_step_value} and the expectation is taken over  the distribution of $b_{\tau}$.
For the non-boundary cases, the single-round expected reward satisfies $\mathbb{E}[v(b_\tau/\tau)] = {v}(\rho)$ if the threshold $\tilde{j}(b_\tau/\tau) = \tilde{j}(\rho)$  for all possible $b_\tau$'s. The regret then is bounded by a constant because the probability of the event $\tilde{j}(b_\tau/\tau) \neq \tilde{j}(\rho)$ decays exponentially due to the concentration property of $b_\tau$.  For the boundary cases, we show the conclusion by relating the regret with the variance of $b_\tau$.
Please refer to Appendix~\ref{app:proof_alp_regret} of the supplementary material for details.

\section{UCB-ALP Algorithm for Constrained Contextual Bandits} \label{sec:ucb_dp}
Now we get back  to the constrained contextual bandits, where  the expected rewards are unknown to the agent.
We assume the agent knows the context distribution as \cite{Badanidiyuru2014COLT}, which will be relaxed in Section~\ref{sec:unknown_dist}.
Thanks to the desirable properties of ALP, the maxim of ``optimism under uncertainty'' \cite{Auer2007UCB4RL} is still applicable and ALP can be extended  to the  bandit settings when combined with estimation policies that can quickly provide correct ranking with high probability.
Here, combining ALP with the UCB method \cite{Auer2002ML:UCB}, we propose a UCB-ALP algorithm for constrained contextual bandits.

\subsection{UCB: Notations and Property}
Let $C_{j,k}(t)$ be the number of times that action  $k \in \mathcal{A}$ has been taken under context $j$ up to round $t$. If $C_{j,k}(t-1) > 0$, let $\bar{u}_{j,k}(t)$ be the empirical reward of action $k$ under context $j$, i.e.,
$\bar{u}_{j,k}(t) = \frac{1}{C_{j,k}(t-1)}\sum_{t' = 1}^{t-1} {Y_{t'} \mathds{1}({X_{t'} = j, A_{t'} = k})}$,
where $\mathds{1}(\cdot)$ is the indicator function.
We define the UCB of $u_{j,k}$ at $t$ as
$\hat{u}_{j,k}(t) = \bar{u}_{j,k}(t) + \sqrt{\frac{{\log t}}{2C_{j,k}(t-1)}}$ for $C_{j,k}(t-1) > 0$, and $\hat{u}_{j,k}(t) = 1$ for $C_{j,k}(t-1) = 0$. Furthermore, we define the UCB of the maximum expected reward under context $j$ as
$
\hat{u}_{j}^*(t) = \max_{k \in \mathcal{A}}\hat{u}_{j,k}(t)
$.
As suggested in \cite{Garivierkl2011COLT:KL-UCB}, we use a smaller coefficient in the exploration term $\sqrt{\frac{{\log t}}{2C_{j,k}(t-1)}}$ than the traditional UCB algorithm \cite{Auer2002ML:UCB} to achieve better performance.

We present the following property of UCB that is important in regret analysis.
\begin{lemma} \label{thm:context_action_pair_errorprob}
For two context-action pairs, $(j, k)$ and $(j', k')$, if $u_{j,k} < u_{j',k'}$, then for any $t \leq T$,
\begin{eqnarray}
\mathbb{P}\{\hat{u}_{j,k}(t) \geq \hat{u}_{j',k'}(t)|C_{j,k}(t-1) \geq \ell_{j,k}\} \leq 2t^{-1},
\end{eqnarray}
where $\ell_{j,k} =  \frac{2\log T}{(u_{j',k'} - u_{j,k})^2}$.
\end{lemma}

Lemma~\ref{thm:context_action_pair_errorprob} states that for two context-action pairs, the ordering of their expected rewards can be identified correctly with high probability, as long as the suboptimal pair has been executed for sufficient times (on the order of $O(\log T)$). This property has been widely applied in the analysis of UCB-based algorithms \cite{Auer2002ML:UCB,Jiang2013CDC}, and its proof can be found in \cite{Jiang2013CDC,Golovin2009Lecture} with a minor modification on the coefficients.

%
%

\subsection{UCB-ALP Algorithm}
We propose a UCB-based adaptive linear programming (UCB-ALP) algorithm, as shown in Algorithm~\ref{alg:ucb_alp}.
As indicated by the name, the UCB-ALP algorithm maintains UCB estimates of expected rewards for all context-action pairs
and then implements the ALP algorithm based on these estimates. Note that the UCB estimates  $\hat{u}_j^*(t)$'s may be non-decreasing in $j$. Thus, the solution of $\mathcal{LP}_{\tau,b}$ based on $\hat{u}_j^*(t)$ depends on the actual ordering of $\hat{u}_j^*(t)$'s and may be different from Eq.~\eqref{eq:LP_solution}. We use $\hat{p}_{j}(\cdot)$ rather than $p_j(\cdot)$ to indicate this difference.

\begin{algorithm}[htbp]
\caption{UCB-ALP}
\label{alg:ucb_alp}
\begin{algorithmic}
\STATE {\bfseries Input:} Time-horizon $T$, budget $B$, and context distribution $\pi_j$'s;

\STATE {\bfseries Init:} $\tau = T$, $b = B$; \\
~~~~~~$C_{j,k}(0) = 0$, $\bar{u}_{j,k}(0) = 0$, $\hat{u}_{j,k}(0) = 1$, $\forall j \in \mathcal{X}$ and $\forall k\in \mathcal{A}$; $\hat{u}^*_{j}(0) = 1$, $\forall j \in \mathcal{X}$;

\FOR{$t = 1$ {\bfseries to} $T$}
\STATE{$k^{*}_j(t) \gets \argmax_{k} \hat{u}_{j,k}(t), ~\forall j$;}
\STATE{$\hat{u}_{j}^*(t) \gets \hat{u}_{j,k_{j}^*(t)}^*(t)$;}
\IF{$b > 0$}
\STATE{ Obtain the probabilities $\hat{p}_{j}(b/\tau)$'s by solving $\mathcal{LP}_{\tau,b}$ with $u_j^*$ replaced by $\hat{u}_j^*(t)$;}
\STATE{ Take action $k^{*}_{X_t}(t)$ with probability $\hat{p}_{X_t}(b/\tau)$;}
\ENDIF
\STATE{Update $\tau$, $b$, $C_{j,k}(t)$, $\bar{u}_{j,k}(t)$,  and $\hat{u}_{j,k}(t)$.}
\ENDFOR
\end{algorithmic}
\end{algorithm}
\vspace{-0.2cm}

\subsection{Regret of UCB-ALP}
We study the regret of UCB-ALP in this section. Due to space limitations, we only present a sketch of the analysis. Specific representations of the regret bounds and proof details can be found in the supplementary material.

Recall that $q_j = \sum_{j'=1}^j \pi_{j'}$ ($1 \leq j \leq J$) are the boundaries defined in Section~\ref{sec:oracle_solution}.
We show that as the budget $B$ and the time-horizon $T$ grow to infinity in proportion, the proposed UCB-ALP algorithm achieves logarithmic regret except for the boundary cases.

\begin{theorem}\label{thm:regret_ucb_alp}
Given $\pi_j$'s, $u_{j,k}$'s and a fixed $\rho \in (0,1)$, the regret of UCB-ALP satisfies:\\
1) (Non-boundary cases) if $\rho \neq q_j$ for any $j \in \{1, 2, \ldots, J-1\}$, then the regret of UCB-ALP is $R_{\rm{UCB-ALP}}(T, B) = O\big(JK\log T\big)$.\\
2) (Boundary cases) if $\rho = q_j$ for some $j \in \{ 1,2, \ldots, J-1\}$, then the regret of UCB-ALP is $R_{\rm{UCB-ALP}}(T, B) = O\big(\sqrt{T} + JK\log T \big)$.
\end{theorem}
Theorem~\ref{thm:regret_ucb_alp} differs from Theorem~\ref{thm:alp_regret} by an additional term $O(JK\log T)$. This term results from using UCB to learn the ordering of expected rewards. Under UCB, each of the $JK$ content-action pairs should be executed roughly $O(\log T)$ times to obtain  the correct ordering.
For the non-boundary cases, UCB-ALP is order-optimal because obtaining the correct action ranking under each context will result in $O(\log T)$ regret \cite{Lai1985AAM}.
Note that our results do not contradict  the lower bound in \cite{Badanidiyuru2014COLT} because we consider discrete contexts and actions, and focus on instance-dependent regret. For the boundary cases, we keep both the $\sqrt{T}$ and $\log T$ terms  because the constant in the $\log T$ term is typically much larger than that in the $\sqrt{T}$ term. Therefore, the $\log T$ term may dominate the regret particularly when the number of context-action pairs is large for medium $T$.
It is still an open problem if one can achieve regret lower than $O(\sqrt{T})$ in these cases.


{\it \textbf{Sketch of Proof:}} We bound the regret of UCB-ALP by comparing its performance  with the benchmark $\widehat{U}(T,B)$. The analysis of this bound is challenging due to the close interactions among different sources of regret and the randomness of context arrivals. We first partition the regret according to the sources and then bound each part of regret, respectively.

\textbf{Step 1: Partition the regret.}
By analyzing the implementation of UCB-ALP, we show that its regret is bounded as
\vspace{-0.1cm}
\begin{equation}
R_{\rm {UCB-ALP}}(T, B) \leq R_{\rm {UCB-ALP}}^{({\rm a})}(T, B) + R_{\rm {UCB-ALP}}^{({\rm c})}(T, B),\nonumber
\end{equation}
where the first part $R_{\rm {UCB-ALP}}^{({\rm a})}(T, B) = \sum_{j=1}^J \sum_{k\neq k_j^*} (u_j^* - u_{j,k})\mathbb{E}[C_{j,k}(T)]$ is the regret from action ranking errors within a context, and the second part $R_{\rm {UCB-ALP}}^{({\rm c})}(T, B) = \sum_{\tau  = 1}^T \mathbb{E}\big[v(\rho) -\sum_{j = 1}^J \hat{p}_j(b_{\tau}/\tau) \pi_j u_j^*\big]$  is the regret from the fluctuations of $b_{\tau}$ and context ranking errors.

\textbf{Step 2: Bound each part of regret. }
For the first part, we can show that $R^{\rm (a)}_{\rm{UCB-ALP}}(T, B) = O(\log T)$ using similar techniques for
traditional UCB methods \cite{Golovin2009Lecture}.
The major challenge of regret analysis for UCB-ALP then lies in the evaluation of the second part $R_{\rm {UCB-ALP}}^{({\rm c})}(T, B)$.

We first verify that the evolution of $b_\tau$ under UCB-ALP is similar to that under ALP and Lemma~\ref{thm:alp_to_hypergeo} still holds under UCB-ALP. With respect to context ranking errors, we note that unlike classic UCB methods,  not all context ranking errors contribute to the regret due to the threshold structure of ALP. Therefore, we carefully categorize the context ranking results based on their contributions. We briefly discuss the analysis for the non-boundary cases here. Recall that $\tilde{j}(\rho)$ is the threshold for the static LP problem $\mathcal{LP}_{T,B}$. We define the following events that capture all possible ranking results based on UCBs:
\begin{align}
&\mathcal{E}_{\rm rank,0}(t) = \big\{\forall j \leq \tilde{j}(\rho), \hat{u}^*_j(t) > \hat{u}^*_{\tilde{j}(\rho) +1}(t); \forall j > \tilde{j}(\rho) +1,\hat{u}^*_j(t) < \hat{u}^*_{\tilde{j}(\rho) +1}(t)\big\},\nonumber\\
&\mathcal{E}_{\rm rank,1}(t)  = \big\{\exists j \leq \tilde{j}(\rho), \hat{u}^*_j(t) \leq \hat{u}^*_{\tilde{j}(\rho) +1}(t); \forall j > \tilde{j}(\rho) +1,\hat{u}^*_j(t) < \hat{u}^*_{\tilde{j}(\rho) +1}(t)\big\},\nonumber\\
&\mathcal{E}_{\rm rank,2}(t)  = \big\{\exists j > \tilde{j}(\rho) +1,\hat{u}^*_j(t) \geq \hat{u}^*_{\tilde{j}(\rho) +1}(t)\big\}. \nonumber
\end{align}
The first event  $\mathcal{E}_{\rm rank,0}(t)$ indicates a roughly correct context ranking, because under $\mathcal{E}_{\rm rank,0}(t)$ UCB-ALP obtains a correct solution for $\mathcal{LP}_{\tau, b_{\tau}}$ if  $b_{\tau}/\tau \in [q_{\tilde{j}(\rho)}, q_{\tilde{j}(\rho) + 1}]$. The last two events $\mathcal{E}_{\rm rank,s}(t)$, $s = 1, 2$, represent two types of context ranking errors: $\mathcal{E}_{\rm rank,1}(t)$ corresponds to ``certain contexts with above-threshold reward having lower UCB'', while $\mathcal{E}_{\rm rank,2}(t)$ corresponds to ``certain contexts with below-threshold reward having higher UCB''.
Let $T^{(s)} = \sum_{t = 1}^T \mathds{1}(\mathcal{E}_{{\rm rank}, s}(t))$ for $0 \leq s \leq 2$.
We can show that the expected number of context ranking errors satisfies $\mathbb{E}[T^{(s)}] = O(JK\log T)$, $s = 1,2$, implying that $R_{\rm {UCB-ALP}}^{({\rm c})}(T, B) = O(JK\log T)$. Summarizing the two parts, we have $R_{\rm {UCB-ALP}}(T, B) = O(JK\log T)$ for the non-boundary cases. The regret for the boundary cases can be bounded using similar arguments.

{\textbf{Key Insights from UCB-ALP:}} Constrained contextual bandits involve complicated interactions between information acquisition and decision making. UCB-ALP alleviates these interactions by approximating the oracle with ALP for decision making. This approximation  achieves near-optimal performance while tolerating certain estimation errors of system statistics, and thus enables the combination with estimation methods such as UCB in unknown statistics cases. Moreover, the adaptation property of UCB-ALP guarantees the concentration property of the system status, e.g., $b_\tau/\tau$. This allows us to separately study the impact of action or context ranking errors and conduct rigorous analysis of regret. These insights can be applied in algorithm design and analysis for constrained contextual bandits under more general settings.

\section{Bandits with Unknown Context Distribution} \label{sec:unknown_dist}
When the context distribution is unknown, a reasonable heuristic  is to replace the probability $\pi_j$ in ALP with its empirical estimate, i.e., $\hat{\pi}_j(t) = \frac{1}{t}{\sum_{t' = 1}^t \mathds{1}(X_{t'} = j)}$.  We refer to this modified ALP algorithm as Empirical ALP (EALP), and its combination with UCB as UCB-EALP.

The empirical distribution provides a maximum likelihood estimate of the context distribution and the  EALP and UCB-EALP algorithms achieve similar performance as ALP and UCB-ALP, respectively, as observed in numerical simulations. However, a rigorous analysis for EALP and UCB-EALP is much more challenging  due to the dependency introduced by the empirical distribution. To tackle this issue, our rigorous analysis focuses on a truncated version of EALP where we stop updating the empirical distribution after a given round. Using the method of bounded averaged differences based on  coupling argument, we obtain the concentration property of  the average remaining budget $b_\tau/\tau$, and show that this truncated EALP algorithm achieves $O(1)$ regret except for the boundary cases. The regret of the corresponding UCB-based version can by bounded similarly as UCB-ALP.

\section{Bandits with Heterogeneous Costs} \label{sec:het_cost}
The insights obtained from unit-cost systems can also be used to design algorithms for heterogeneous cost systems where the cost $c_{j,k}$ depends on $j$ and $k$. We generalize the ALP algorithm to approximate the oracle, and adjust it to the case with unknown expected rewards. For simplicity, we assume the context distribution is known here, while the empirical estimate can be used to replace the actual context distribution if it is unknown, as discussed in the previous section.

With heterogeneous costs, the quality of an action $k$ under a context $j$ is roughly captured by its \emph{normalized expected reward}, defined as $\eta_{j,k} = u_{j,k}/c_{j,k}$. However, the agent cannot only focus on the ``best'' action, i.e., $k_j^* = \argmax_{k \in \mathcal{A}} \eta_{j,k}$, for context $j$.
This is because there may exist another action $k'$ such that $\eta_{j,k'} < \eta_{j,k_{j}^*}$, but $u_{j,k'} > u_{j,k_j^*}$ (and of course, $c_{j,k'} > c_{j, k_j^*}$). If the budget  allocated to context $j$ is sufficient, then the agent may take action $k'$ to maximize the expected reward.
Therefore, to approximate the oracle, the ALP algorithm in this case needs to solve an LP problem accounting for all context-action pairs with an additional constraint that only one action can be taken under each context.
By investigating the structure of ALP in this case and the concentration of the remaining budget, we show that  ALP achieves $O(1)$ regret in non-boundary cases, and $O(\sqrt{T})$ regret in boundary cases. Then, an $\epsilon$-First ALP algorithm is proposed for the unknown statistics case where an exploration stage is implemented first and then an exploitation stage is implemented according to ALP.



\section{Conclusion} \label{sec:conclusion}
In this paper, we study computationally-efficient algorithms that achieve logarithmic or sublinear regret for constrained contextual bandits. Under simplified yet practical assumptions, we show that the close interactions between the information acquisition and decision making in constrained contextual bandits can be decoupled by adaptive linear relaxation. When the system statistics are known, the ALP approximation achieves near-optimal performance, while tolerating certain estimation errors of system parameters. When the expected rewards are unknown, the proposed UCB-ALP algorithm leverages the advantages of ALP and UCB, and achieves $O(\log T)$ regret except for certain boundary cases, where it achieves $O(\sqrt{T})$ regret. Our study provides an efficient approach of  dealing with the challenges introduced by budget constraints and could potentially be extended to more general constrained contextual bandits. 

\textbf{Acknowledgements:}
This research was supported in part by NSF Grants CCF-1423542, CNS-1457060, CNS-1547461, and AFOSR MURI Grant FA 9550-10-1-0573. 


\bibliography{OnlineOpt,mypublications_0802,math}
\bibliographystyle{unsrt_abbrv}
\clearpage
\newpage
\pagebreak

\appendix
\section*{Appendices}


%

\section{Proof of Lemma~\ref{thm:upper_bound}: Upper Bound}\label{app:proof_upper_bound}




We prove Lemma~\ref{thm:upper_bound} by comparing $\widehat{U}(T,B)$ with the expected total reward under any feasible algorithm satisfying the budget constraint.

Let $C_j$ be the number of rounds that an action is taken  under context $j$ for any realization under any feasible algorithm with known statistics. Let $p_j = \mathbb{E}[C_j]/(\pi_j T)$, which satisfies $0 \leq p_j \leq 1$. Then the expected total reward becomes $\sum_{j = 1}^J u_j^* \mathbb{E}[C_j] = T\sum_{j = 1}^J p_j \pi_j u_j^*$. Further, because the hard budget constraint is met for all realizations, i.e., $\sum_{j = 1}^J C_j \leq  B$, we have $\sum_{j = 1} p_j \pi_j = \sum_j E[C_j]/T \leq B/T$. Thus, the expected total reward obtained by any feasible algorithm, including the oracle algorithm, is upper bounded by $\widehat{U}(T,B)$. 

\section{Proof of Theorem~\ref{thm:alp_regret}: Near Optimality of ALP}\label{app:proof_alp_regret}

\subsection{Proof of Lemma~\ref{thm:alp_to_hypergeo}: Evolution of Remaining Budget}

The evolution of the remaining budget $b_\tau$ is critical  for evaluating the expected total reward under ALP. We prove Lemma~\ref{thm:alp_to_hypergeo} by casting ALP to a sampling problem without replacement.

From the implementation of ALP, we can verify that when the remaining time is $\tau$ and remaining budget is $b_{\tau} = b$,
the system consumes one unit of budget with probability $b/\tau$, and consumes nothing with probability $1-b/\tau$.
Thus, when focusing on the remaining budget,
we can view the ALP algorithm as a sampling problem without replacement as follows.

\textbf{Mapping ALP to Sampling without Replacement:} Consider $T$ balls in an urn,
including $B$ black balls and $T-B$ white balls. Running ALP is equivalent to randomly drawing a ball without replacement.
Taking an action $A_t > 0$ is equivalent to drawing  a black ball and taking the dummy action $A_t = 0$ is equivalent to drawing a white ball.
The event that $b_{\tau} = b$ is equivalent to the event that the agent draws $T- \tau$ balls, and the number of drawn black balls is $B-b$.

Based on the above mapping and using its symmetric property, we know that $b_\tau$ follows the hypergeometric distribution \cite{Dubhashi2009Concentration} and complete the proof of Lemma~\ref{thm:alp_to_hypergeo}.

\subsection{Part 1: Non-Boundary Cases}\label{app:proof_alp_regret_non_boundary}
According to Lemma~\ref{thm:upper_bound}, $\widehat{U}(T,B)$ is an upper bound on the total expected reward. Thus,
\begin{eqnarray} \label{eq:alp_regret}
U^*(T,B) - U_{\rm ALP}(T,B)
\leq  \widehat{U}(T,B) - U_{\rm ALP}(T,B) =  \sum_{\tau = 1}^T \big\{{v}(\rho)-\mathbb{E}[{v}(b_\tau/\tau)]\big\}.
\end{eqnarray}
To evaluate the gap between the single-round values, we define an auxiliary function $\tilde{v}(b/\tau)$ for a given $\rho$ as follows:
\begin{eqnarray}\label{eq:approx_single_round_alp}
\tilde{v}(b/\tau) = \sum_{j = 1}^{\tilde{j}(\rho) }\pi_j u_j^*
 +\pi_{\tilde{j}(\rho) + 1} \tilde{p}_{\tilde{j}(\rho) + 1}(b/\tau) u_{\tilde{j}(\rho) + 1}^*,
\end{eqnarray}
where
\begin{eqnarray}
\tilde{p}_{\tilde{j}(\rho) + 1}(b/\tau) = \frac{b/\tau - \sum_{j = 1}^{\tilde{j}(\rho)} \pi_j} {\pi_{\tilde{j}(\rho) +1}}. \nonumber
\end{eqnarray}
This auxiliary function bridges the gap of single-round values, $v(\rho)$ and $\mathbb{E}[v(b_\tau/\tau)]$, as follows:

First, we note that $\tilde{v}(b/\tau)$ uses the same threshold $\tilde{j}(\rho)$ as in $v(\rho)$.
The only difference between $\tilde{v}(b/\tau)$ and $v(\rho)$ is that $\tilde{v}(b/\tau)$ uses the instantaneous
average budget $b/\tau$ instead of the fixed average budget $\rho$. Considering all possible $b$'s and according to Lemma~\ref{thm:alp_to_hypergeo},
we have
\begin{eqnarray}\label{eq:virtual_and_upperbound}
{v}(\rho) - \mathbb{E}[{\tilde{v}}(b_{\tau}/\tau)] = \big\{\rho - \mathbb{E}[b_{\tau}/\tau]\big\}u_{\tilde{j}(\rho) + 1}^* = 0.
\end{eqnarray}

Second, compared with $v(b/\tau)$, the difference of the auxiliary function $\tilde{v}(b/\tau)$ comes
from the event of $\tilde{j}(b/\tau) \neq \tilde{j}(\rho)$, which only occurs when $b/\tau < q_{\tilde{j}(\rho)}$
or $b/\tau > q_{\tilde{j}(\rho)+1}$.
Because $\rho \neq q_j$, $1 \leq j \leq J-1$,  there exists a positive number $\delta = \min\{\rho -q_{\tilde{j}(\rho)}, q_{\tilde{j}(\rho)+1} - \rho\}$
 such that  for all $\rho - \delta \leq \rho' < \rho + \delta$, the threshold under $\rho'$ is the same as that under $\rho$, i.e.,
$
\tilde{j}(\rho') = \tilde{j}(\rho). \nonumber
$
Therefore, for all $b$ satisfying $\rho - \delta \leq b/\tau \leq \rho + \delta$,
$v(b/\tau) = \tilde{v}(b/\tau)$.

If $b/\tau < \rho - \delta$, then
\begin{eqnarray} \label{eq:single_step_gap_left}
&&\tilde{v}(b/\tau) -  v(b/\tau) \nonumber \\
&=&  \bigg[ \sum_{j = \tilde{j}(b/\tau)  +1 }^{\tilde{j}(\rho)} \pi_j u_j^*
+ \big(\frac{b}{\tau} - q_{ \tilde{j}(\rho)}\big) u_{\tilde{j}(\rho) + 1}^*  \bigg]  - \big(\frac{b}{\tau} - q_{ \tilde{j}(b/\tau)}\big) u_{\tilde{j}(b/\tau)  +1}^* \nonumber \\
&\leq &  \bigg[ u_1^* \sum_{j = \tilde{j}(b/\tau)  +1 }^{\tilde{j}(\rho)} \pi_j
+ \big(\frac{b}{\tau} - q_{ \tilde{j}(\rho)}\big) u_{\tilde{j}(\rho) + 1}^*  \bigg]  - \big(\frac{b}{\tau} - q_{ \tilde{j}(b/\tau)   }\big) u_{\tilde{j}(\rho) + 1}^* \nonumber \\
&\leq& (u_1^* - u_{\tilde{j}(\rho) + 1}^*)\sum_{j = \tilde{j}(b/\tau)  +1 }^{\tilde{j}(\rho)} \pi_j     \nonumber \\
&\leq & q_{\tilde{j}(\rho)}(u_1^* -  u_J^*).
\end{eqnarray}

Similarly, if $b/\tau > \rho + \delta$, then
\begin{eqnarray} \label{eq:single_step_gap_right}
&& \tilde{v}(b/\tau) -  v(b/\tau) \nonumber \\
 &=&
\big(\frac{b}{\tau} - q_{ \tilde{j}(\rho)}\big) u_{\tilde{j}(\rho) +1}^* - \bigg[ \sum_{j = \tilde{j}(\rho)+1}^{\tilde{j}(b/\tau)} \pi_j u_j^* + \big(\frac{b}{\tau} - q_{ \tilde{j}(b/\tau) }\big) u_{\tilde{j}(b/\tau)+1}^*  \bigg] \nonumber \\
&\leq& (1 - q_{\tilde{j}(\rho)}) (u_1^* - u_J^*).
\end{eqnarray}

Summing all the above three cases ($\rho - \delta \leq b/\tau \leq \rho + \delta$, $b/\tau < \rho - \delta$, and $b/\tau > \rho + \delta$) and using Eq.~\eqref{eq:virtual_and_upperbound}, we have
\begin{eqnarray} \label{eq:virtual_and_alp}
&&v(\rho) - \mathbb{E}[v(b_\tau/\tau)] \nonumber\\
&=& v(\rho)- \mathbb{E}[\tilde{v}(b_\tau/\tau)] + \mathbb{E}[\tilde{v}(b_\tau/\tau) - v(b_\tau/\tau)]  \nonumber\\
& =&  \sum_{ b < \tau(\rho - \delta)~{\rm or}~b >\tau(\rho + \delta)} \mathbb{P}(b_\tau = b)[\tilde{v}(b/\tau) -v(b/\tau)] \nonumber\\
&\leq & q_{\tilde{j}(\rho)}(u_1^* - u_J^*) \mathbb{P}\{b_{\tau} < \tau(\rho - \delta)\} \nonumber \\
&&+ (1-q_{\tilde{j}(\rho)}) (u_1^* - u_J^*) \mathbb{P}\{b_{\tau} > \tau(\rho + \delta)\}     \nonumber \\
&\leq & (u_1^* - u_J^*) e^{-2\delta^2\tau}.
\end{eqnarray}

Part 1 of Theorem~\ref{thm:alp_regret} then follows by substituting Eq.~\eqref{eq:virtual_and_alp} into Eq.~\eqref{eq:alp_regret}.

\subsection{Part~2: Boundary Cases}\label{app:proof_alp_regret_all}
The proof of Part~2 of Theorem~\ref{thm:alp_regret} is similar to that of Part~1. Specifically, when  $\rho = q_{\tilde{j}(\rho)}$, let $\delta' = \min\{\rho - q_{\tilde{j}(\rho)-1}, q_{\tilde{j}(\rho)+1} - \rho\}$.   From the proof of Part 1,
we know that
 \begin{eqnarray}
v(\rho)- \mathbb{E}[\tilde{v}(b_\tau/\tau)] = 0.
\end{eqnarray}

In addition, if $\rho \leq b/\tau \leq \rho + \delta'$, then $\tilde{j}(b/\tau) =  \tilde{j}(\rho)$ and $v(b/\tau) = \tilde{v}(b/\tau)$.

If $\rho - \delta' \leq b/\tau <\rho$, we have $\tilde{j}(b/\tau) = \tilde{j}(\rho) - 1$, and
\begin{eqnarray}
&&\tilde{v}(b/\tau) - v(b/\tau) \nonumber \\
& = & \pi_{\tilde{j}(\rho)} u^*_{\tilde{j}(\rho)} + (\frac{b}{\tau} - q_{\tilde{j}(\rho)})u^*_{\tilde{j}(\rho)+ 1}  - (\frac{b}{\tau} - q_{\tilde{j}(b/\tau)})u^*_{\tilde{j}(b/\tau) + 1} \nonumber \\
& = & (\pi_{\tilde{j}(\rho)} + q_{\tilde{j}(\rho) - 1} - \frac{b}{\tau})u^*_{\tilde{j}(\rho)} + (\frac{b}{\tau} - q_{\tilde{j}(\rho)})u^*_{\tilde{j}(\rho)+ 1} \nonumber \\
& = & (\rho - \frac{b}{\tau})u^*_{\tilde{j}(\rho)} + (\frac{b}{\tau} - \rho)u^*_{\tilde{j}(\rho)+ 1} \nonumber \\
&\leq & \big|\frac{b}{\tau} - \rho\big|(u_1^* - u^*_J).
\end{eqnarray}

Moreover, we still have \eqref{eq:single_step_gap_left} if  $b/\tau < \rho - \delta'$, and \eqref{eq:single_step_gap_right} if $b/\tau > \rho + \delta'$.

Compared with the proof of Part~1, we know that the only difference relies on the case of $\rho - \delta' \leq b/\tau <\rho$. Thus, summing all the above cases and using the results in the analysis of Part~1, we have
\begin{eqnarray}
\mathbb{E}[\tilde{v}(b_\tau/\tau) - v(b_{\tau}/\tau)] \leq (u_1^* - u_J^*)\big\{\mathbb{E}[|b_{\tau}/\tau - \rho|] + e^{-2(\delta')^2\tau}\big\} \leq  (u_1^* - u_J^*)\big[\sqrt{\frac{{\rm Var}(b_{\tau})}{\tau^2}} + e^{-2(\delta')^2\tau}\big] . \nonumber
\end{eqnarray}
Consequently,
\begin{eqnarray}
 U^*(T,B) -  U_{\rm ALP}(T,B) &\leq & \widehat{U}(T,B) -  U_{\rm ALP}(T,B) \nonumber \\
&=&  \sum_{\tau = 1}^T \big\{v(\rho)-\mathbb{E}[v(b_\tau/\tau)]\} \nonumber \\
&=& \sum_{\tau = 1}^T \big\{v(\rho)- \mathbb{E}[\tilde{v}(b_\tau/\tau)] \big\} + \sum_{\tau = 1}^T  \mathbb{E}[ \tilde{v}(b_\tau/\tau) - v(b_{\tau}/\tau) ]  \nonumber\\
&\leq& (u_1^* - u_J^*)\sum_{\tau = 1}^T\big[\sqrt{\frac{{\rm Var}(b_{\tau})}{\tau^2}} + e^{-2(\delta')^2\tau}\big] \nonumber\\
& = &(u_1^* - u_J^*)\sum_{\tau = 1}^T\big[\sqrt{\frac{(T-\tau)\rho(1-\rho)}{(T-1)\tau}}+ e^{-2(\delta')^2\tau}\big]  \nonumber\\
& \leq &(u_1^* - u_J^*) \sqrt{\rho(1-\rho)} \sum_{\tau = 1}^T\sqrt{\frac{1}{\tau}}  + \frac{u_1^* - u_J^*}{1-  e^{-2(\delta')^2}}\nonumber\\
&\leq &2\sqrt{\rho(1-\rho)} (u_1^* - u_J^*) \sqrt{T} + \frac{u_1^* - u_J^*}{1- e^{-2(\delta')^2}}. \nonumber
\end{eqnarray}

\section{Proof of Theorem~\ref{thm:regret_ucb_alp}: Regret of UCB-ALP}
We bound the regret of UCB-ALP by comparing its performance  with the benchmark $\widehat{U}(T,B)$. To obtain this upper bound, we first partition the regret according to the sources and then bound each part of the regret, respectively.

Before presenting the proof, we first introduce a notation that will be widely used later. For contexts $j$ and $j'$, and an action $k$, let $\Delta_{j,k}^{(j')}$ be the difference between the expected reward for action $k$ under context $j$ and the highest expected reward under context $j'$, i.e., $\Delta_{j,k}^{(j')} = u_{j'}^* - u_{j,k}$.
When $j' = j$, $\Delta_{j,k}^{(j)}$ is the difference of expected reward between the suboptimal action $k$ and the best action under context $j$.

\subsection{Step 1: Partition the Regret}

Note that the total reward of the oracle solution $U^*(T,B)  \leq \widehat{U}(T,B)$. Thus, we can bound the regret of UCB-ALP by comparing its total expected reward $U_{\rm UCB-ALP}(T,B)$ with $\widehat{U}(T,B)$, i.e.,
\begin{eqnarray}
&&R_{\rm {UCB-ALP}}(T, B) \nonumber \\
&=& U^*(T,B) - U_{\rm UCB-ALP}(T,B) \nonumber \\
& \leq &\widehat{U}(T,B) - U_{\rm UCB-ALP}(T,B) \nonumber\\
&= & T v(\rho) - \sum_{j = 1}^J \sum_{k =1}^K u_{j,k} \mathbb{E}[C_{j,k}(T)].
\end{eqnarray}
The total expected reward of UCB-ALP can be further divided as
\begin{eqnarray}
&& U_{\rm UCB-ALP}(T,B) \nonumber \\
& =&  \sum_{j = 1}^J u_j^* \mathbb{E}\big[\sum_{k=1}^K C_{j,k}(T)\big] - \sum_{j=1}^J \sum_{k=1}^K \Delta_{j,k}^{(j)}\mathbb{E}[C_{j,k}(T)] \nonumber \\
& =&  \sum_{j = 1}^J u_j^*\mathbb{E}\big[C_{j}(T)\big] - \sum_{j=1}^J \sum_{k=1}^K \Delta_{j,k}^{(j)}\mathbb{E}[C_{j,k}(T)], \nonumber
\end{eqnarray}
where $C_j(T) = \sum_{k=1}^K C_{j,k}(T)$ is the total number that actions have been taken under context $j$ up to round $T$.

Consequently, the regret of UCB-ALP can be bounded as
\begin{eqnarray}\label{eq:regret_ucb_alp__nonboundary_details}
R_{\rm {UCB-ALP}}(T, B) \leq R_{\rm {UCB-ALP}}^{({\rm a})}(T, B) + R_{\rm {UCB-ALP}}^{({\rm c})}(T, B),
\end{eqnarray}
where
\begin{eqnarray}
R_{\rm {UCB-ALP}}^{({\rm a})}(T, B) = \sum_{j=1}^J \sum_{k=1}^K \Delta_{j,k}^{(j)}\mathbb{E}[C_{j,k}(T)], \nonumber
\end{eqnarray}
and
\begin{eqnarray}
R_{\rm {UCB-ALP}}^{({\rm c})}(T, B) = \sum_{\tau  = 1}^T \mathbb{E}\bigg[v(\rho) -\sum_{j = 1}^J \hat{p}_j(b_{\tau}/\tau) \pi_j u_j^*\bigg]. \nonumber
\end{eqnarray}

Eq.~\eqref{eq:regret_ucb_alp__nonboundary_details} clearly shows that the regret of the UCB-ALP algorithm can be divided into two parts: the first part $R_{\rm {UCB-ALP}}^{({\rm a})}(T, B)$ is from taking  suboptimal actions under a given context; the second part $R_{\rm {UCB-ALP}}^{({\rm c})}(T, B)$ is from the deviation of remaining budget $b_{\tau}$ and context ranking errors.

\subsection{Step 2: Bound Each Part of Regret}

\subsubsection{Step 2.1: Bound of $R_{\text{UCB-ALP}}^{\rm (a)}(T, B)$}
For the regret from action ranking errors,
we show in Lemma~\ref{thm:regret_withincontext} that $R_{\text{UCB-ALP}}^{\rm (a)}(T, B) = O(\log T)$ using similar techniques for traditional UCB methods \cite{Golovin2009Lecture}.
\begin{lemma}\label{thm:regret_withincontext}
Under UCB-ALP, the regret due to the action ranking errors within context $j$ satisfies
\begin{eqnarray} \label{eq:regret_ucb_alp_regret_withincontext}
&& R_{\text{UCB-ALP}}^{\rm (a)}(T, B) \nonumber\\
&\leq& \sum_{j=1}^J \sum_{k \neq k_j^*} \bigg[\big(\frac{2}{\Delta_{j,k}^{(j)}} + 2\Delta_{j,k}^{(j)}\big) \log T + 2 \Delta_{j,k}^{(j)}\bigg].
\end{eqnarray}
\end{lemma}
\begin{proof}
For $k \neq k_j^*$, let $\ell_{j,k}^{(j)} = \frac{2\log T}{(\Delta_{j, k}^{(j)})^2}$. According to Lemma~\ref{thm:context_action_pair_errorprob}, we have
\begin{eqnarray}
&& \mathbb{E}[C_{j,k}(T)] \nonumber \\
&\leq& \ell_{j,k}^{(j)} + \sum_{t = 1}^T \mathbb{P}\{X_t = j, A_t = k, C_{j,k}(t-1) \geq \ell_{j,k}^{(j)}\} \nonumber \\
&\leq & \ell_{j,k}^{(j)} + \sum_{t = 1}^T \mathbb{P}\{X_t = j, A_t = k | C_{j,k}(t-1) \geq \ell_{j,k}^{(j)}\} \nonumber \\
&\leq & \ell_{j,k}^{(j)} + \sum_{t = 1}^T  2t^{-1}. \nonumber
\end{eqnarray}
The conclusion then follows by the facts that $\sum_{t = 1}^T t^{-1} \leq 1 + \log T$ and $R_{\text{UCB-ALP}}^{\rm (a)}(T, B) =  \sum_{j = 1}^J\sum_{k \neq k_j^*} \Delta_{j,k}^{(j)} \mathbb{E}[C_{j,k}(T)]$.
\end{proof}

\subsubsection{Step 2.2: Bound of $R_{\text{UCB-ALP}}^{\rm (c)}(T, B)$}
Next, we show that the second part $R_{\rm {UCB-ALP}}^{({\rm c})}(T, B) = O(\log T)$.
\textbf{We first present the proof for the non-boundary cases}, and discuss the boundary cases later.

Note that we have separately considered the regret due to action ranking errors in  $R_{\text{UCB-ALP}}^{\rm (a)}(T, B)$ and we only need to consider the best action of each context for $R_{\text{UCB-ALP}}^{\rm (c)}(T, B)$. Thus, we define $v^*_{\rm UCB-ALP}(\tau, b_{\tau})$ as follows:
\begin{equation}
v^*_{\rm UCB-ALP}(\tau, b_{\tau})= \sum_{j = 1}^J \tilde{p}_j(b_{\tau}/\tau) \pi_j u_j^*.\nonumber
\end{equation}
Let $\Delta v_\tau$ be the single-round difference between  UCB-ALP and the upper bound, i.e.,
\begin{eqnarray}
\Delta v_\tau = v(\rho) -v^*_{\rm UCB-ALP}(\tau, b_{\tau}).\nonumber \
\end{eqnarray}
Then $R_{\rm {UCB-ALP}}^{({\rm c})}(T, B) = \sum_{\tau = T}^1 \mathbb{E}[\Delta v_\tau]$. We study the expectation $\mathbb{E}[\Delta v_{\tau}]$ under all possible situations. For a random variable $X$ and event $\mathcal{E}$, let $\mathbb{E}[X,\mathcal{E}] = \mathbb{E}[X\mathds{1}(\mathcal{E})]$. Then, the expectation $\mathbb{E}[X] = \mathbb{E}[X,\mathcal{E}] + \mathbb{E}[X,\urcorner\mathcal{E}]$. Therefore,
\begin{eqnarray} \label{eq:correct_rank_single_round_regret}
\mathbb{E}[\Delta v_\tau]  =  \sum_{s = 0}^2\mathbb{E}[\Delta {v}_{\tau},  \mathcal{E}_{{\rm rank}, s}(T - \tau + 1)].
\end{eqnarray}

We first consider the case of $s = 0$ and convert the expectation value into other two cases. Considering all possible value of $b_\tau$, we have
\begin{eqnarray} \label{eq:diff_case0}
&& \mathbb{E}[\Delta {v}_{\tau},  \mathcal{E}_{{\rm rank}, 0}(T - \tau + 1)]\nonumber \\
& = & \sum_{b = 0}^{B} \mathbb{E}[\Delta {v}_{\tau}| b_{\tau} = b,  \mathcal{E}_{{\rm rank}, 0}(T - \tau + 1)] \mathbb{P}\{b_{\tau} = b,  \mathcal{E}_{{\rm rank}, 0}(T - \tau + 1)\}.
\end{eqnarray}

For the probability, we have
\begin{align}\label{eq:diff_case0_prob}
\mathbb{P}\{b_{\tau} = b, \mathcal{E}_{{\rm rank}, 0}(T - \tau + 1)\}
= \mathbb{P}\{b_{\tau} = b\} - \sum_{s= 1}^2 \mathbb{P}\{b_{\tau} = b, \mathcal{E}_{{\rm rank}, s}(T - \tau + 1)\}.
\end{align}

For the conditioned expectation, we note that  $\mathcal{E}_{{\rm rank}, 0}(T - \tau + 1)$ provides a roughly correct context rank in the sense that if $b_{\tau}/\tau$ is close to $\rho$, then $v^*_{\rm UCB-ALP}(\tau, b_{\tau}) = v(b_\tau/\tau)$, where $v(b_\tau/\tau)$ is the single round value with the correct context rank.
Specifically, letting $\delta = \frac{1}{2}\min\{\rho - q_{\tilde{j}(\rho)},q_{\tilde{j}(\rho)+1}- \rho\}$.
If $b \in [\rho - \delta, \rho + \delta]$, then $v^*_{\rm UCB-ALP}(\tau, b) = v(b/\tau)$, and thus,
\begin{eqnarray} \label{eq:diff_case0_exp}
\mathbb{E}[\Delta {v}_{\tau}| b_{\tau} = b,  \mathcal{E}_{{\rm rank}, 0}(T - \tau + 1)] =  v(\rho) - v(b/\tau).
\end{eqnarray}

Combining Eqs.~\eqref{eq:diff_case0} $\sim$ \eqref{eq:diff_case0_exp} and using the facts that $v(\rho) \geq 0$ and $v^*_{\rm UCB-ALP}(\tau, b) \geq 0$, we have
\begin{align}  \label{eq:correct_rank_regret}
&\quad\mathbb{E}[\Delta {v}_{\tau},  \mathcal{E}_{{\rm rank}, 0}(T - \tau + 1)] \nonumber\\
&\leq  v(\rho) -  \sum_{b = 0}^B v(b/\tau) \mathbb{P}\{b_{\tau} = b\} \nonumber \\
&~ + \sum_{s= 1}^2\sum_{b = 0}^B v(b/\tau) \mathbb{P}\{b_{\tau} = b, \mathcal{E}_{{\rm rank}, s}(T - \tau + 1)\}\nonumber\\
&~ + \sum_{b \notin [\rho - \delta, \rho + \delta]}v(b/\tau)\mathbb{P}\{b_{\tau} = b, \mathcal{E}_{{\rm rank}, 0}(T - \tau + 1)\}.
\end{align}
Recall that under UCB-ALP, the remaining budget $b_{\tau}$ follows the hypergeometric distribution. Using the same method as the analysis of Eq.~\eqref{eq:virtual_and_alp}, we have
\begin{eqnarray} \label{eq:correct_rank_regret_part1}
v(\rho) -  \sum_{b = 0}^B v(b/\tau) \mathbb{P}\{b_{\tau} = b\}
\leq (u_1^* - u_J^*) e^{-2\delta^2\tau}.
\end{eqnarray}

In addition,
\begin{eqnarray} \label{eq:correct_rank_regret_part2}
\sum_{b \notin [\rho - \delta, \rho + \delta]}v(b/\tau)\mathbb{P}\{b_{\tau} = b, \mathcal{E}_{{\rm rank}, 0}(T - \tau + 1)\}
\leq \bar{u}^* \sum_{b \notin [\rho - \delta, \rho + \delta]}\mathbb{P}\{b_{\tau} = b\}\leq 2\bar{u}^* e^{-2\delta^2 \tau},
\end{eqnarray}
where $\bar{u}^* = \sum_{j = 1}^J \pi_j u_j^*$ is the expected reward without budget constraint.

Moreover,
\begin{eqnarray} \label{eq:correct_rank_regret_part3}
\sum_{b = 0}^B v(b/\tau) \mathbb{P}\{b_{\tau} = b, \mathcal{E}_{{\rm rank}, s}(T - \tau + 1)\} \leq  \bar{u}^*\mathbb{P}\{\mathcal{E}_{{\rm rank}, s}(T - \tau + 1)\},
\end{eqnarray}

Substituting Eqs.~\eqref{eq:correct_rank_regret_part1} $\sim$ \eqref{eq:correct_rank_regret_part3} into Eq.~\eqref{eq:correct_rank_regret}, we have
\begin{eqnarray}  \label{eq:correct_rank_regret_results}
\mathbb{E}[\Delta {v}_{\tau},  \mathcal{E}_{{\rm rank}, 0}(T - \tau + 1)] \leq   [u_1^* - u_J^* +2\bar{u}^*] e^{-2\delta^2\tau} + \bar{u}^*\sum_{s=1}^2\mathbb{P}\{\mathcal{E}_{{\rm rank}, s}(T - \tau + 1)\}.
\end{eqnarray}

When the rank is wrong, i.e., $1\leq s \leq 2$, since $\Delta {v}_{\tau} \leq v(\rho)$ under any possible ranking results,  we have
\begin{eqnarray}  \label{eq:error_rank_regret_results}
\mathbb{E}[\Delta {v}_{\tau},  \mathcal{E}_{{\rm rank}, s}(T - \tau + 1)] \leq v(\rho) \mathbb{P}\{\mathcal{E}_{{\rm rank}, s}(T - \tau + 1)\}.
\end{eqnarray}

Substituting Eqs.~\eqref{eq:correct_rank_regret_results} and \eqref{eq:error_rank_regret_results} into Eq.~\eqref{eq:correct_rank_single_round_regret}, we have
\begin{eqnarray}  
\mathbb{E}[\Delta {v}_{\tau}] \leq  [u_1^* - u_J^* + 2\bar{u}^*] e^{-2\delta^2\tau}
 + [\bar{u}^* + v(\rho)]\sum_{s=1}^2\mathbb{P}\{\mathcal{E}_{{\rm rank}, s}(T - \tau + 1)\}.\nonumber
\end{eqnarray}

Note that $R_{\rm {UCB-ALP}}^{({\rm c})}(T, B) = \sum_{\tau = 1}^T \mathbb{E}[\Delta {v}_{\tau}]$. Thus
\begin{eqnarray} \label{eq:regret_context_ranking_nonboundary}
R_{\rm {UCB-ALP}}^{({\rm c})}(T, B) \leq  \frac{[u_1^* - u_J^*+2\bar{u}^*] e^{-2\delta^2}}{1 - e^{-2\delta^2}}
+ [\bar{u}^* + v(\rho)]\sum_{s=1}^2\mathbb{E}[T^{(s)}],
\end{eqnarray}
where $T^{(s)} = \sum_{t = 1}^T \mathds{1}(\mathcal{E}_{{\rm rank}, s}(t))$ ($s = 1,2$) is the number of type-$s$ ranking errors.

Next, we bound the expected number of context ranking errors. From Lemma~\ref{thm:context_action_pair_errorprob}, we know that to obtain the correct ordering of two context-action pairs with high probability, the agent needs to execute the suboptimal context-action pair for enough times. Unlike traditional MABs, however,  the context-action pair with the higher UCB in a round might not be executable, as the context of that round could be different. Fortunately, the following lemma will show that if the condition that causes the an context-action pair to be executed with a positive probability appears many times,  the  context-action pair will indeed be executed proportionally with high probability.

\begin{lemma} \label{thm:context_action_pair_errorprob_wContext}
Assume $\mathcal{E}_t$'s, $\hat{\mathcal{E}}_t$'s are events in round $t$ ($1\leq t \leq T$), satisfying $\mathbb{P}\{\mathcal{E}_t| \hat{\mathcal{E}}_t, \mathcal{H}_{1:t-1}\} = \mathbb{P}\{\mathcal{E}_t| \hat{\mathcal{E}}_t\} \geq p> 0$, where $\mathcal{H}_{1:t-1}$ is the filtration from 1 to $t-1$. Let $C(T) = \sum_{t=1}^T \mathds{1}(\mathcal{E}_t)$ and $\widehat{C}(T) = \sum_{t=1}^T \mathds{1}(\hat{\mathcal{E}}_t)$. Then,
\begin{equation}
\mathbb{P}\{C(T) \leq (p-\epsilon) N, \widehat{C}(T) \geq N \} \leq e^{-2 \epsilon^2 N}. \nonumber
\end{equation}
\end{lemma}
\begin{proof}
One may think the proof of this lemma is trivial because  $\mathbb{P}\{C(T) \leq (p-\epsilon) N, \widehat{C}(T) \geq N \} \leq \mathbb{P}\{C(T) \leq (p-\epsilon) N |\widehat{C}(T) \geq N \}$ and we can bound the right-hand-side using Chernoff bound. However, this is incorrect because although $\mathcal{E}_t$ is independent of the history given $\hat{\mathcal{E}}_t$, the event $\hat{\mathcal{E}}_{t+1}$ may depend on $\mathcal{E}(t)$.

We prove this lemma using the coupling argument.
Let $S_t = \mathds{1}(\mathcal{E}_t \cap \hat{\mathcal{E}}_t)$, and $C_S(T) = \sum_{t=1}^T S_t$. Then, we have
\begin{equation}
\mathbb{P}\{C(T) \leq (p-\epsilon) N, \widehat{C}(T) \geq N \} \leq \mathbb{P}\{C_S(T) \leq (p-\epsilon) N, \widehat{C}(T) \geq N \}.
\end{equation}

Now, we show $\mathbb{P}\{C_S(T) \leq (p-\epsilon) N, \widehat{C}(T) \geq N \} \leq e^{-2 \epsilon^2 N}$  using the coupling argument.

First, generate  $W_1, W_2, \ldots, W_T$  i.i.d. according to Bernoulli distribution with $\mathbb{P}\{W_t = 1\} = p$.

Next, generate a sequences  $(V'_t, S'_t, 1\leq t \leq T)$ as follows:

For each $t$, generate $V'_t$ according to Bernoulli distribution with $\mathbb{P}\{V'_t = 1\} = \mathbb{P}\{\hat{\mathcal{E}}_t = 1|\mathds{1}(\hat{\mathcal{E}}_{t'})= V'_{t'},\mathds{1}({\mathcal{E}}_{t'})= S'_{t'} ~1\leq t' \leq t-1\}$. Further, we  generate $S'_t$ conditioned on the value of $V'_t$ and $W_t$. Specifically, let $C_{V'}(t) = \sum_{t'= 1}^t V'_t$.

\textbf{1)} If $V'_t = 1$, generate $S'_t$ conditioned on $W_{C_{V'}(t)}$: \\
{\it \hspace{0.1in}a.} If $W_{C_{V'}(t)} = 1$, then $S'_t = 1$;\\
{\it \hspace{0.1in}b.} If $W_{C_{V'}(t)} = 0$, then generate $S'_t$ according to Bernoulli distribution with
\begin{equation}
\mathbb{P}\{S'_t = 1 | W_{C_{V'}(t)} = 0\} = \frac{\mathbb{P}[\mathds{1}({\mathcal{E}}_t) = 1| \mathds{1}(\hat{\mathcal{E}}_t) = 1]-p}{1-p}. \nonumber\\
\end{equation}
\textbf{2)} If $V'_t = 0$, let $S'_t = 0$.

We can verify that $(V'_t, S'_t, 1\leq t \leq T)$ has the  same distribution as $(\mathds{1}(\hat{\mathcal{E}}_t), S_t, 1\leq t \leq T)$. Hence, $\mathbb{P}\{C_S(T) \leq (p-\epsilon) N, \widehat{C}(T) \geq N \} = \mathbb{P}\{\sum_{t=1}^T S'_t \leq (p-\epsilon)N, \sum_{t=1}^T V'_t \geq N\}$.

On the other hand, from the generation of $S'_t$,  we have $\sum_{t=1}^T S'_t \geq \sum_{t=1}^{C_{V'}(T)} W_t$.
Thus, the event $\{\sum_{t=1}^T S'_t \leq (p-\epsilon)N, \sum_{t=1}^T V'_t \geq N\}$ implies $\{\sum_{t=1}^N W_t \leq (p-\epsilon) N\}$, and
\begin{equation}
\mathbb{P}\{\sum_{t=1}^T S'_t \leq (p-\epsilon)N, \sum_{t=1}^T V'_t \geq N\}\leq \mathbb{P}\{\sum_{t=1}^N W_t \leq (p-\epsilon) N\} \leq e^{-2 \epsilon^2 N}.
\end{equation}
The conclusion of the lemma then follows.
\end{proof}

The following lemma bounds the expected number of context ranking errors.
\begin{lemma}\label{thm:error_ranking}
Given $\pi_j$'s, $u_{j,k}$'s and a fixed $\rho \in (0,1)$, $\rho  \neq q_j$ ($1 \leq j \leq J-1$), under the UCB-ALP algorithm, we have
\begin{align}
&\mathbb{E}[T^{(1)}] \leq \sum_{j = 1}^{\tilde{j}(\rho)}\sum_{k=1}^K\frac{27\log T}{2g_{\tilde{j}(\rho)+1} [\Delta^{(j)}_{\tilde{j}(\rho)+1,k}]^2} + 2K\tilde{j}(\rho) \log T + O(1),\nonumber
\end{align}
\begin{align}
&\mathbb{E}[T^{(2)}] \leq \sum_{j =\tilde{j}(\rho)+2}^{J}\sum_{k=1}^K\frac{27\log T}{2g_j [\Delta^{(\tilde{j}(\rho)+1)}_{j,k}]^2} + 2K[J - \tilde{j}(\rho)-1]\log T + O(1), \nonumber
\end{align}
where
$
g_j  =\min\big\{\pi_j, \frac{1}{2}(\rho - q_{\tilde{j}(\rho)}), \frac{1}{2}(q_{\tilde{j}(\rho)+1} -\rho)\big\}.
$
\end{lemma}

\begin{proof}
We only prove the conclusion for the case of $s = 1$ as the other case can be analyzed similarly.
From Algorithm~\ref{alg:ucb_alp}, we can see that the evolution of the remaining budget also affects the execution of the UCB-ALP algorithm. Under the assumption of known context distribution, it can be verified that Lemma~\ref{thm:alp_to_hypergeo} holds under UCB-ALP, i.e., the remaining budget $b_{\tau}$ follows the hypergeometric distribution and has the properties described in Lemma~\ref{thm:alp_to_hypergeo}. We define an event $\mathcal{E}_{\rm budget, 0}(t)$ as follows,
\begin{eqnarray}
\mathcal{E}_{\rm budget, 0}(t)  = \{(\rho - \delta) \tau \leq b_{\tau} \leq (\rho + \delta) \tau\}, \nonumber
\end{eqnarray}
where $\delta$ is given by
\begin{eqnarray}
\delta = \frac{1}{2}\min\{\rho - q_{\tilde{j}(\rho)}, q_{\tilde{j}(\rho)+1} -\rho\}. \nonumber
\end{eqnarray}
According to Lemma~\ref{thm:alp_to_hypergeo}, we have
\begin{eqnarray}
\mathbb{P}\{\urcorner\mathcal{E}_{\rm budget, 0}(t)\}
= \mathbb{P}\{ b_{\tau} <  (\rho -\delta) \tau\} + \mathbb{P}\{ b_{\tau} >  (\rho +\delta) \tau\} \leq 2e^{-2\delta^2 (T-t + 1)}. \nonumber
\end{eqnarray}

Back to the ranking event $\mathcal{E}_{{\rm rank}, 1}(t)$, we have
\begin{eqnarray}
\mathbb{P}(\mathcal{E}_{{\rm rank}, 1}(t))
\leq  \mathbb{P}(\urcorner\mathcal{E}_{\rm budget, 0}(t))
+\mathbb{P}(\mathcal{E}_{{\rm rank}, 1}(t)\cap \mathcal{E}_{\rm budget, 0}(t)). \nonumber
\end{eqnarray}
Note that the event $\mathcal{E}_{{\rm rank}, 1}(t)$ can be divided as follow:
\begin{eqnarray}
\mathcal{E}_{{\rm rank}, 1}(t) \subseteq  \bigcup_{1\leq j \leq \tilde{j}(\rho), 1\leq k\leq K} \mathcal{E}^{(j,k)}_{{\rm rank}, 1}(t), \nonumber
\end{eqnarray}
where for $1 \leq j \leq \tilde{j}(\rho)$ and $1 \leq k \leq K$,
\begin{eqnarray}
 \mathcal{E}^{(j,k)}_{{\rm rank}, 1}(t) = \big\{\forall j' > \tilde{j}(\rho) +1,\hat{u}^*_{j'}(t) < \hat{u}^*_{\tilde{j}(\rho) +1}(t);
\hat{u}^*_{j}(t) \leq \hat{u}^*_{\tilde{j}(\rho)+1}(t), k_j^*(t) = k\big\}.\nonumber
\end{eqnarray}
Thus,
\begin{eqnarray}\label{eq:error1_partition}
\mathbb{E}[T^{(1)}] &=& \sum_{t = 1}^T \mathbb{P}(\mathcal{E}_{{\rm rank}, 1}(t))\leq \frac{2e^{-2\delta^2}}{1 - e^{-2\delta^2}} + \sum_{j = 1}^{\tilde{j}(\rho)} \sum_{k= 1}^K\mathbb{E}[N^{(1)}_{j,k}(T)],\nonumber\\
\end{eqnarray}
where for $1 \leq j\leq \tilde{j}(\rho)$ and $1 \leq k \leq K$,
\begin{eqnarray}
N^{(1)}_{j,k}(t) = \sum_{t' =1}^t \mathds{1}(\mathcal{E}^{(j,k)}_{{\rm rank}, 1}(t'), \mathcal{E}_{\rm budget, 0}(t')). \nonumber
\end{eqnarray}
Let  $\hat{\ell}^{(j_1)}_{j_2,k} = \frac{2\log T}{g_{j_2} (1- \epsilon)\epsilon^2(\Delta_{j_2, k}^{(j_1)})^2}$, where $g_{j_2} = \min\{\pi_{j_2}, \delta\}$ and $\epsilon \in (0,1)$.
Similar to the analysis of UCB in \cite{Auer2002ML:UCB},  we have
\begin{align}\label{eq:error1_bound_error_obervations}
\mathbb{E}[N^{(1)}_{j,k}(T)] \leq  \hat{\ell}^{(j)}_{\tilde{j}(\rho)-1,k} + \sum_{t = 1}^T \mathbb{P}\big\{\mathcal{E}^{(j,k)}_{{\rm rank}, 1}(t), \mathcal{E}_{\rm budget, 0}(t),  N^{(1)}_{j,k}(t-1) \geq  \hat{\ell}^{(j)}_{\tilde{j}(\rho)+1,k}\big\}.
\end{align}

For each $t$ in the second term, we have
\begin{align}
&\quad\mathbb{P}\{\mathcal{E}^{(j,k)}_{{\rm rank}, 1}(t), \mathcal{E}_{\rm budget, 0}(t),N^{(1)}_{j,k}(t-1) \geq  \hat{\ell}^{(j)}_{\tilde{j}(\rho)+1,k}\} \nonumber \\
&\leq  \mathbb{P}\{\mathcal{E}^{(j,k)}_{{\rm rank}, 1}(t), \mathcal{E}_{\rm budget, 0}(t) |C_{\tilde{j}(\rho)+1,k}(t-1) \geq g_{\tilde{j}(\rho)+1}(1-\epsilon)\hat{\ell}^{(j)}_{\tilde{j}(\rho)+1,k}\} \nonumber \\
& \quad + \mathbb{P}\{C_{\tilde{j}(\rho)+1,k}(t-1) < g_{\tilde{j}(\rho)+1}(1-\epsilon)\hat{\ell}^{(j)}_{\tilde{j}(\rho)+1,k},  N^{(1)}_{j,k}(t-1) \geq  \hat{\ell}^{(j)}_{\tilde{j}(\rho)+1,k}\},~~\nonumber
\end{align}
where $C_{\tilde{j}(\rho)+1,k}(t) = \sum_{t'=1}^t \mathds{1}(X_{t'} = \tilde{j}(\rho)+1,A_{t'} = k)$ is the number that the context-action pair $(\tilde{j}(\rho)+1,k)$ has been executed up to round $t$.

For the first term, we note that the event $\{\mathcal{E}^{(j,k)}_{{\rm rank}, 1}(t), \mathcal{E}_{\rm budget, 0}(t)\}$ implies that $\hat{u}_{j, k_{j}^*}(t) \leq \hat{u}_{\tilde{j}(\rho)+1,k}(t)$. Because ${u}_{j, k_{j}^*} > {u}_{\tilde{j}(\rho)+1,k}$ for all $j \leq \tilde{j}(\rho)$ and $k$, according to Lemma~\ref{thm:context_action_pair_errorprob}, we have
\begin{align}\label{eq:error1_bound_ucb_rank}
& \quad \mathbb{P}\{\mathcal{E}^{(j,k)}_{{\rm rank}, 1}(t), \mathcal{E}_{\rm budget, 0}(t)|C_{\tilde{j}(\rho)+1,k}(t-1) \geq g_{\tilde{j}(\rho)+1}(1-\epsilon)\hat{\ell}^{(j)}_{\tilde{j}(\rho)+1,k}\} \nonumber \\
&\leq \mathbb{P}\{\hat{u}_{j, k_{j}^*}(t) \leq \hat{u}_{\tilde{j}(\rho)+1,k}(t) |C_{\tilde{j}(\rho)+1,k}(t-1) \geq g_{\tilde{j}(\rho)+1}(1-\epsilon)\hat{\ell}^{(j)}_{\tilde{j}(\rho)+1,k}\} \nonumber \\
&\leq 2t^{-1}.
\end{align}

For the second term, we note that since context $\tilde{j}(\rho)+1$ arrives with probability $\pi_{\tilde{j}(\rho)+1}$ independent of the observations, we have
\begin{align}
\mathbb{P}\big\{X_{t} = \tilde{j}(\rho)+1, A_{t} = k | \mathcal{E}^{(j,k)}_{{\rm rank}, 1}(t), \mathcal{E}_{\rm budget, 0}(t)\big\}
 = \min\{\delta, \pi_{\tilde{j}(\rho)+1}\} = g_{\tilde{j}(\rho)+1}. \nonumber
\end{align}
Thus, according to Lemma~\ref{thm:context_action_pair_errorprob_wContext}, we have
\begin{align}\label{eq:error1_bound_pulls}
\mathbb{P}\{C_{\tilde{j}(\rho)+1,k}(t-1) < g_{\tilde{j}(\rho)+1}(1-\epsilon)\hat{\ell}^{(j)}_{\tilde{j}(\rho)+1,k},  N^{(1)}_{j,k}(t-1) \geq  \hat{\ell}^{(j)}_{\tilde{j}(\rho)+1,k}\}  \leq e^{-2\epsilon^2 \hat{\ell}^{(j)}_{\tilde{j}(\rho)+1,k}} \leq T^{-4}.
\end{align}

Substituting Eqs.~\eqref{eq:error1_bound_ucb_rank} and \eqref{eq:error1_bound_pulls} into Eq.~\eqref{eq:error1_bound_error_obervations}, we have
\begin{eqnarray}\label{eq:error1_bound_error_obervations_result}
\mathbb{E}[N^{(1)}_{j,k}(T)]\leq  \hat{\ell}^{(j)}_{\tilde{j}(\rho)+1,k} + \sum_{t=1}^T (2 t^{-1} + T^{-4}) \leq \hat{\ell}^{(j)}_{\tilde{j}(\rho)+1,k} + 2\log T + 2 + T^{-3}.
\end{eqnarray}
Substituting Eq.~\eqref{eq:error1_bound_error_obervations_result} to  Eq.~\eqref{eq:error1_partition} and letting $\epsilon = 2/3$ in $\hat{\ell}^{(j)}_{\tilde{j}(\rho)-1,k}$, we have
\begin{eqnarray}
\mathbb{E}[T^{(1)}]
&\leq& \frac{2e^{-2\delta^2}}{1 - e^{-2\delta^2}} + \sum_{j =1}^{\tilde{j}(\rho)}\sum_{k=1}^K\hat{\ell}^{(j)}_{\tilde{j}(\rho)+1,k} + 2K\tilde{j}(\rho)\log T + O(1) \nonumber \\
&\leq&\sum_{j = 1}^{\tilde{j}(\rho)}\sum_{k=1}^K\frac{27\log T}{2g_{\tilde{j}(\rho)+1} (\Delta^{(j)}_{\tilde{j}(\rho)+1,k})^2} + 2K\tilde{j}(\rho)\log T + O(1).
\end{eqnarray}

\end{proof}

Combining Lemma~\ref{thm:regret_withincontext}, Lemma~\ref{thm:error_ranking}, and Eq.~\eqref{eq:regret_context_ranking_nonboundary}, we have
\begin{eqnarray}
\limsup_{T\to \infty} \frac{R_{\rm{UCB-ALP}}(T, B)}{\log T} \leq \Theta^{\rm (a)} + \Theta^{\rm (c)}_{\rm nb}, \nonumber
\end{eqnarray}
where
\begin{align}
&\Theta^{\rm (a)} = \sum_{j=1}^J \sum_{k \neq k_j^*} \big(\frac{2}{\Delta_{j,k}^{(j)}} + 2\Delta_{j,k}^{(j)}\big),\nonumber\\
&\Theta^{\rm (c)}_{\rm nb} = [\bar{u}^* + v(\rho)]\bigg\{\sum_{j = 1}^{\tilde{j}(\rho)}\sum_{k=1}^K\frac{27}{2g_{\tilde{j}(\rho)+1} [\Delta^{(j)}_{\tilde{j}(\rho)+1,k}]^2} + \sum_{j =\tilde{j}(\rho)+2}^{J}\sum_{k=1}^K\frac{27}{2g_j [\Delta^{(\tilde{j}(\rho)+1)}_{j,k}]^2} + 2KJ \bigg\}.\nonumber
\end{align}
This completes the proof of Part 1 in Theorem~\ref{thm:regret_ucb_alp}.

\textbf{Next, we discuss the bound of $R_{\rm {UCB-ALP}}^{({\rm c})}(T, B)$ for the boundary cases}. The analysis is similar to  the non-boundary cases with slight modification on the threshold.

We note that fundamentally, the context $\tilde{j}(\rho) + 1$ for $\rho \neq q_j$ and the context $\tilde{j}(\rho)$ for $\rho = q_j$ are both the minimum context with positive probability in the static LP problem.
Thus, we can define the  context ranking events $\mathcal{E}_{\rm rank,s}(t)$ $(0 \leq s\leq 2)$ similar to  the analysis of Part~1, with $\tilde{j}(\rho) + 1$ replaced by $\tilde{j}(\rho)$.
Then, we have
\begin{eqnarray}
R_{\rm {UCB-ALP}}^{({\rm c})}(T, B) = \sum_{\tau = 1}^T \mathbb{E}[\Delta {v}_{\tau}], \nonumber
\end{eqnarray}
where
\begin{eqnarray}
\mathbb{E}[\Delta v_\tau]  =  \sum_{s = 0}^2\mathbb{E}[\Delta {v}_{\tau},  \mathcal{E}_{{\rm rank}, s}(T - \tau + 1)]. \nonumber
\end{eqnarray}
For the case of $s = 0$,
\begin{align}
\mathbb{E}[\Delta {v}_{\tau},  \mathcal{E}_{{\rm rank}, 0}(T - \tau + 1)] = \sum_{b = 0}^B \mathbb{E}[\Delta {v}_{\tau}|b_{\tau} = b, \mathcal{E}_{{\rm rank}, 0}(T - \tau + 1)]\mathbb{P}\{b_\tau = b, \mathcal{E}_{{\rm rank}, 0}(T - \tau + 1)\}. \nonumber
\end{align}
When $b/\tau \in [\rho -\delta, \rho + \delta]$ and $\mathcal{E}_{{\rm rank}, 0}(T - \tau + 1)$ occurs, we have $\Delta {v}_{\tau} \leq (u_1^*-u_J^*) |\rho - b/\tau|$. Moreover, $\Delta {v}_{\tau} \leq v(\rho)$ under any condition. Thus,
\begin{align}
&\quad\mathbb{E}[\Delta {v}_{\tau},  \mathcal{E}_{{\rm rank}, 0}(T - \tau + 1)] \nonumber \\
&\leq u_1^* \mathbb{E}[|b_{\tau}/\tau - \rho|] + v(\rho) \sum_{b \notin [\rho -\delta, \rho + \delta]}\mathbb{P}\{b_\tau = b\}\nonumber \\
& \leq u_1^* \sqrt{\frac{{\rm Var}(b_{\tau})}{\tau^2}} + 2v(\rho)e^{-2\delta^2\tau}. \nonumber
\end{align}

For the other cases of $s = 1,2$, we have
\begin{eqnarray}
\mathbb{E}[\Delta {v}_{\tau},  \mathcal{E}_{{\rm rank}, s}(T - \tau + 1)] \leq v(\rho) \mathbb{P}\{\mathcal{E}_{{\rm rank}, s}(T - \tau + 1)\}. \nonumber
\end{eqnarray}

On the other hand, we extend Lemma~\ref{thm:error_ranking} to the boundary cases:
\begin{align}
&\limsup_{T\to \infty}\frac{\mathbb{E}[T^{(1)}]}{\log T} \leq \sum_{j < \tilde{j}(\rho)}\sum_{k=1}^K\frac{27}{2g_{\tilde{j}(\rho)} [\Delta^{(j)}_{\tilde{j}(\rho),k}]^2} + {2K[\tilde{j}(\rho)-1] },\nonumber\\
&\limsup_{T\to \infty}\frac{\mathbb{E}[T^{(2)}]}{\log T} \leq \sum_{j \geq \tilde{j}(\rho)+1}\sum_{k=1}^K\frac{27}{2g_j [\Delta^{(\tilde{j}(\rho))}_{j,k}]^2} + 2K[J - \tilde{j}(\rho)], \nonumber
\end{align}
where
\begin{align}
&\quad g_j  = \min\big\{\pi_j, \frac{1}{2}(\rho - q_{\tilde{j}(\rho)-1}), \frac{1}{2}(q_{\tilde{j}(\rho)+1} -\rho)\big\}.\nonumber
\end{align}

Consequently, we can bound $R_{\rm {UCB-ALP}}^{({\rm c})}(T, B)$ by summing over the entire horizon and using the properties of $T^{(1)}$ and $T^{(2)}$. The conclusion of Part 2 of Theorem~\ref{thm:regret_ucb_alp} then follows by adding the bound of $R_{\rm {UCB-ALP}}^{({\rm a})}(T, B)$ and $R_{\rm {UCB-ALP}}^{({\rm c})}(T, B)$.

%

\section{Two-Context Systems with Unit-Cost} \label{app:two_contex}
As a special case, the oracle algorithm can be obtained for two-context systems with unit costs. When the context distribution and expected rewards are unknown, the oracle algorithm can be combined with the UCB method to achieve logarithmic regret under both boundary and non-boundary cases.

\subsection{Oracle Algorithm: Procrastinate-for-the-Better-context}
When there are only two contexts, the oracle algorithm is trivial.
Under the unit-cost assumption, skipping the worse context does not waste any opportunities if $b_\tau < \tau$.
Thus, the agent can reserve budget for the better context, unless there is sufficient budget; i.e., we have the following algorithm:

\textbf{Procrastinate-for-the-Better (PB):} If $X_t = 1$ and $b_\tau > 0$, or if $b_\tau \geq \tau$, take action  $A_t = k_{X_t}^*$;
otherwise, $A_t = 0$.

We can verify that the above PB algorithm achieves the highest expected reward for any realization of the context arrival process.
Thus, the PB algorithm is optimal in two-context systems. We note that the PB algorithm does not need to know the context distribution and only requires the ordering of the expected rewards. This property allows us to extend it to the case where the context distribution or expected rewards are unknown.
\subsection{UCB-PB: Logarithmic Regret Algorithm for Two-Context Bandits with Unit-Cost}\label{app:ucb_pb}
When the context distribution and expected rewards are unknown, we propose the UCB-based Procrastinate-for-the-Better (UCB-PB) algorithm for solving the constrained contextual bandit problem with two contexts and unit costs.

\begin{algorithm}[htbp]
\caption{UCB-PB}
\label{alg:ucb_dp}
\begin{algorithmic}
\STATE {\bfseries Input:} Time-horizon $T$, budget $B$;

\STATE {\bfseries Init:} Remaining time $\tau = T$, remaining budget $b = B$; \\
~~~~~~$C_{j,k}(0) = 0$, $\bar{u}_{j,k}(0) = 0$, $\hat{u}_{j,k}(0) = 1$, for all $j \in \mathcal{X}$ \\
~~~~~~and $k\in \mathcal{A}$; $\hat{u}^*_{j}(0) = 1$ for all $j \in \mathcal{X}$;

\FOR{$t = 1$ {\bfseries to} $T$}
\STATE{$k^{*}_j(t) \gets \argmax_{k} \hat{u}_{j,k}(t), ~\forall j$;}
\STATE{$\hat{u}_{j}^*(t) \gets \hat{u}_{j,k_{j}^*(t)}^*(t), ~\forall j$;}
\STATE{${j}^*(t) \gets \argmax_{j} \hat{u}_{j}^*(t)$;}
\IF{$b \geq \tau$ \OR  ($0 < b < \tau$ \AND $X_t = j^*(t)$)}
\STATE{ Take action $k^{*}_{X_t}(t)$;}
\ENDIF
\STATE{Update $\tau$, $b$, $C_{j,k}(t)$, $\bar{u}_{j,k}(t)$, and $\hat{u}_{j,k}(t)$;}
\ENDFOR
\end{algorithmic}
\end{algorithm}

As shown in Algorithm~\ref{alg:ucb_dp}, the agent maintains UCB estimates $\hat{u}_{j,k}(t)$'s for the expected rewards of all context-action pairs. In each round, the agent implements the PB algorithm based on these estimates.

Next, we study the regret of the UCB-PB algorithm.  We  show that the UCB-PB algorithm achieves logarithmic regret for any given $\rho \in (0,1)$.
\begin{theorem}\label{thm:regret_ucb_dp}
For a constrained contextual bandit with unit-cost and two contexts, the UCB-PB algorithm achieves  logarithmic regret as $T$ goes to infinity, i.e.,
\begin{eqnarray}
\limsup_{T \to \infty} \frac{R_{\text{UCB-PB}}(T, B)}{\log T} \leq \sum_{k=1}^K \bigg[\frac{27}{2\pi_2\Delta^{(1)}_{2,k}} + 2\Delta^{(1)}_{2,k} \bigg] + \sum_{j=1}^2 \sum_{k \neq k_j^*}\bigg[\frac{2}{\Delta_{j,k}^{(j)}} + 2\Delta_{j,k}^{(j)}\bigg]. \nonumber
\end{eqnarray}
\end{theorem}
\begin{proof}
The proof of Theorem~\ref{thm:regret_ucb_dp} is similar to that of Theorem~\ref{thm:regret_ucb_alp}, while the analysis on the error events is even simpler.
Note that the regret is defined as the difference between the expected total rewards achieved by the UCB-PB algorithm and the oracle algorithm. For the oracle algorithm, let $C_j^*(t) = \sum_{t' = 1}^t \mathds{1}\{X_{t'} = j, A_{t'} = k_j^*\}$ be the number of times that the context-action pair $(j, k_j^*)$ has been executed up to round $t$. For the UCB-PB algorithm, Recall that $C_{j,k}(t) = \sum_{t' = 1}^t \mathds{1}\{X_{t'} = j, A_{t'} = k_j\}$ is the number of times that the context-action pair $(j,k)$ has been executed up to round $t$, and let $C_j(t) = \sum_{k=1}^K C_{j,k}(t)$.
Then the regret of UCB-PB can be expressed as
\begin{eqnarray} \label{eq:regret_ucb_dp_details}
&&R_{\text{UCB-PB}}(T, B) \nonumber \\
&=& \sum_{j = 1}^J u_j^* \mathbb{E}[C_j^*(T)] - \sum_{j = 1}^J \sum_{k =1}^K u_{j,k} \mathbb{E}[C_{j,k}(T)] \nonumber\\
& =& \sum_{j=1}^J \sum_{k=1}^K \Delta^{(j)}_{j,k}\mathbb{E}[C_{j,k}(T)] + \sum_{j = 1}^J u_j^* \mathbb{E}\big[C_j^*(T) -\sum_{k=1}^K C_{j,k}(T)\big]\nonumber \\
& =& R_{\text{UCB-PB}}^{\rm (a)}(T, B) + R_{\text{UCB-PB}}^{\rm (c)}(T, B),
\end{eqnarray}
where $R_{\text{UCB-PB}}^{\rm (a)}(T, B)$ is the regret due to action-ranking errors, i.e.,
\begin{eqnarray}
R_{\text{UCB-PB}}^{\rm (a)}(T, B) &=& \sum_{j=1}^J \sum_{k\neq k_j^*} \Delta^{(j)}_{j,k}\mathbb{E}[C_{j,k}(T)], \nonumber
\end{eqnarray}
and $R_{\text{UCB-PB}}^{\rm (c)}(T, B)$ is the regret due to context-ranking errors, i.e.,
\begin{eqnarray} \label{eq:regret_ucb_dp_context}
R_{\text{UCB-PB}}^{\rm (c)}(T, B) &=& \sum_{j = 1}^J u_j^* \mathbb{E}\big[C_j^*(T) -\sum_{k=1}^K C_{j,k}(T)\big] =(u_1^* - u_2^*) \mathbb{E}\big[C_1^*(T) - C_1(T)\big]. 
\end{eqnarray}
The expression of $R_{\text{UCB-PB}}^{\rm (c)}(T, B)$ uses the fact that both the oracle algorithm and UCB-PB  will exhaust their entire budget, i.e., $\sum_{j=1}^J C_j^*(T) = \sum_{j=1}^J C_j(T) = B$.

For $R_{\text{UCB-PB}}^{\rm (a)}(T, B)$, we note that  Lemma~\ref{thm:regret_withincontext} also holds under UCB-PB, i.e.,
\begin{eqnarray} \label{eq:regret_ucb_dp_action}
R_{\text{UCB-PB}}^{\rm (a)}(T, B)
\leq \sum_{j=1}^J \sum_{k \neq k_j^*} \bigg[\big(\frac{2}{\Delta_{j,k}^{(j)}} + 2\Delta_{j,k}^{(j)}\big) \log T + 2 \Delta_{j,k}^{(j)}\bigg].
\end{eqnarray}

Next, we show that $R_{\text{UCB-PB}}^{\rm (c)}(T, B)$ is also of order $O(\log T)$.
Let $(\hat{X}_t, \hat{A}_t)$ be the context-action pair that has the highest UCB in round $t$. Moreover, let $\hat{C}_j(t)$ be the number of events that context $j$ has the maximum index up to round $t$, i.e., $\hat{C}_j(t) = \sum_{t' = 1}^{t}\mathds{1}(\hat{X}_t = j)$, and $\hat{C}_{j,k}(t)$ be the number of events that the context-action pair $(j,k)$ has the highest UCB up to round $t$, i.e., $\hat{C}_{j,k}(t) = \sum_{t' = 1}^{t}\mathds{1}(\hat{X}_t = j, \hat{A}_t = k)$. We show that the UCB-PB algorithm mistakes the suboptimal context as the optimal context for at most $O(\log T)$ times, i.e., $\mathbb{E} [\hat{C}_2(T)] = O(\log T)$, and then $\mathbb{E}\big[C_1^*(T) - C_1(T)\big] \leq \mathbb{E} [\hat{C}_2(T)] = O(\log T)$.

Specifically, consider the suboptimal context $j =  2$.
For  $1 \leq k \leq K$, we have
\begin{align}
\mathbb{E}\big[\hat{C}_{2,k}(T)\big] \leq \hat{\ell}_{2,k}^{(1)}+ \sum_{t = 1}^T \mathbb{P}\{\hat{X}_t = 2, \hat{A}_t = k, b_{\tau} > 0, \hat{C}_{2,k}(t-1) \geq \hat{\ell}_{2,k}^{(1)}\}, \nonumber
\end{align}
where $\hat{\ell}_{2,k}^{(1)} = \frac{2\log T}{\pi_2(1-\epsilon)\epsilon^2(\Delta_{2,k}^{(1)})^2}$, and $\epsilon  \in (0,1)$.

Based on Lemma~\ref{thm:context_action_pair_errorprob_wContext},  we have
\begin{align} \label{eq:lowerbound_pull}
\mathbb{P}\{{C}_{2,k}(t-1) < \pi_2(1 - \epsilon)\hat{\ell}_{2,k}^{(1)}, b_\tau > 0, \hat{C}_{2,k}(t-1) \geq \hat{\ell}_{2,k}^{(1)}\} \leq e^{-2\epsilon^2\hat{\ell}_{2,k}}\leq T^{-4}.
\end{align}
Thus,
\begin{align}
&\quad\mathbb{P}\{\hat{X}_t = 2, \hat{A}_t = k, \hat{C}_{2,k}(t-1) \geq \hat{\ell}_{2,k}^{(1)}, b_\tau > 0\} \nonumber \\
&\leq  \mathbb{P}\{\hat{X}_t = 2, \hat{A}_t = k, {C}_{2,k}(t-1) \geq \pi_2(1 - \epsilon)\hat{\ell}_{2,k}^{(1)}\} \nonumber\\
&\quad+ \mathbb{P}\{{C}_{2,k}(t-1) < \pi_2(1 - \epsilon)\hat{\ell}_{2,k}^{(1)}, \hat{C}_{2,k}(t-1) \geq \hat{\ell}_{2,k}^{(1)}, b_\tau > 0\} \nonumber \\
&\leq  \mathbb{P}\{\hat{u}_{2,k}^*(t) > \hat{u}_{1,k_1^*}(t)| {C}_{2,k}(t-1) \geq \pi_2(1 - \epsilon)\hat{\ell}_{2,k}^{(1)}\} \nonumber \\
&\quad+ \mathbb{P}\{{C}_{2,k}(t-1) < \pi_2(1 - \epsilon)\hat{\ell}_{2,k}^{(1)}, \hat{C}_{2,k}(t-1) \geq \hat{\ell}_{2,k}^{(1)}, b_\tau >0\} \nonumber \\
&\leq 2t^{-1} + T^{-4}, \nonumber
\end{align}
where the last inequality results from Lemma~\ref{thm:context_action_pair_errorprob} (note that for $j=2$, $\hat{u}_{2,k}(t) < \hat{u}_{1,k_1^*}(t) \leq \hat{u}_1^*(t)$) and  Eq.~\eqref{eq:lowerbound_pull}.

Summing over all actions,  we have
\begin{eqnarray}
\mathbb{E}[\hat{C}_2(T)]   &= & \sum_{k = 1}^K\mathbb{E}\big[\hat{C}_{2,k}(T)\big]
  \leq   \sum_{k = 1}^K\hat{\ell}_{2,k}^{(1)} + \sum_{t = 1}^T\sum_{k = 1}^K(2t^{-1} +T^{-4}) \nonumber \\
&=&\sum_{k = 1}^K \bigg[\frac{2}{\pi_2(1-\epsilon)\epsilon^2(\Delta_{2,k}^{(1)})^2} + 2\bigg]\log T + O(1). \nonumber
\end{eqnarray}


Consequently,
\begin{eqnarray} \label{eq:regret_ucb_dp_context_num}
\mathbb{E}[C_1^*(T) - C_1(T)]
&\leq & \sum_{k = 1}^K \bigg[\frac{2}{\pi_2(1-\epsilon)\epsilon^2(\Delta_{2,k}^{(1)})^2} + 2\bigg]\log T + O(1) \nonumber \\
& = & \sum_{k = 1}^K \bigg[\frac{27}{2\pi_2(\Delta_{2,k}^{(1)})^2} + 2\bigg]\log T + O(1).
\end{eqnarray}
The last equality is obtained by letting $\epsilon = 2/3$.
Combining Eqs.~\eqref{eq:regret_ucb_dp_action}, \eqref{eq:regret_ucb_dp_context}, and \eqref{eq:regret_ucb_dp_context_num},  and using the fact that $u_1^* - u_2^* \leq u_1^* - u_{2,k}$ for all $k$, we can obtain the conclusion of Theorem~\ref{thm:regret_ucb_dp}.
\end{proof}

\section{Constrained Contextual Bandits with Unknown Context Distribution} \label{app:unknown_dist}
In this section, we relax the assumption of known context distribution and study unit-cost systems with unknown context distribution. Since the arrival of contexts is independent of the actions taken by the agent.
a natural idea is to implement the ALP or UCB-ALP algorithm based on the empirical distribution as follows:

\textbf{EALP and UCB-EALP Algorithms:} the agent maintains  the empirical distribution of the contexts, denoted by $\hat{\boldsymbol{\pi}}_t = (\hat{\pi}_{1,t}, \hat{\pi}_{2,t}, \ldots, \hat{\pi}_{J,t})$, where $\hat{\pi}_{j,t} = \frac{1}{t}\sum_{t' = 1}^t \mathds{1}(X_{t'} = j)$. In each round, the agent executes the ALP (when the expected rewards are known) or UCB-ALP (when the expected rewards are unknown) algorithms with the context distribution $\boldsymbol{\pi}$ in $\mathcal{LP}_{\tau, b}$ replaced by $\hat{\boldsymbol{\pi}}_t$. These algorithms are referred to as Empirical ALP (EALP) and UCB-EALP, respectively.

As we can see from the numerical simulations in Appendix~\ref{sec:sim_results}, the above EALP and UCB-EALP algorithms have similar performance as ALP and UCB-ALP, respectively. However, the regret analysis for these algorithms is challenging because the empirical distribution introduces complex temporal dependency since the empirical distribution depends on the context arrivals in all the past rounds, which makes it difficult to analyze the evolution of the remaining budget. Thus, we focus on the non-boundary cases and consider truncated version of EALP and UCB-EALP. Specifically, we study algorithms that stop updating the empirical distribution from the $T_1$-th (will be defined later) round and use the fixed estimate $\hat{\boldsymbol{\pi}}_{T_1}$ for the remaining rounds, which are  referred to as EALP2 (as shown in Algorithm~\ref{alg:ealp2}) and UCB-EALP2, respectively. We focus on the EALP2 algorithm for the case where the expected rewards are known, while the properties of UCB-EALP2 can be obtained by similar techniques in the analysis of UCB-ALP combined with the properties of EALP2.
\begin{algorithm}[htbp]
\caption{EALP2}
\label{alg:ealp2}
\begin{algorithmic}
\STATE {\bfseries Input:} Time horizon $T$, budget $B$, learning stage length $T_1$, and expected rewards $u_j^*$'s;

\STATE {\bfseries Init:} $\tau = T$; $b = B$; $\hat{\pi}_{j,0} = 0$, $\forall j$;

\FOR{$t = 1$ {\bfseries to} $T$}
\IF{$t \leq T_1$}
\STATE{$\hat{\pi}_{j,t} \gets \frac{(t-1)\hat{\pi}_{j,t-1} + \mathds{1}(X_t = j)}{t}$, $\forall j$;}
\ENDIF
\IF{$b > 0$}
\STATE{Obtain the probabilities $p_j(b/\tau)$'s by solving $\mathcal{LP}_{\tau,b}$ with $\boldsymbol{\pi}$ replaced by $\hat{\boldsymbol{\pi}}_t$.}
\STATE{Take action $k_{X_t}^*$ with probability $p_{X_t}(b/\tau)$.}
\ENDIF
\ENDFOR
\end{algorithmic}
\end{algorithm}

Now we show that for a sufficiently large $T$ and an appropriate chosen $T_1$, the EALP2 algorithm achieves similar performance as ALP in the non-boundary cases. Let $\delta = \min\{q_{\tilde{j}(\rho) + 1} - \rho, \rho - q_{\tilde{j}(\rho)}\}$ be the gap between $\rho$ and the boundaries. The following theorem shows that EALP2 achieves $O(1)$ regret with appropriately chosen $T$ and $T_1$.
\begin{theorem} \label{thm:ealp2_regret}
Given a fixed $\rho \in (0,1)$, $\rho \neq q_j$, $j = 1,2,\ldots, J-1$. If $T_1 = 16 J^2\log^3 T/\delta^2$ and $T$ satisfies $\log^3 T/T \leq \delta^3/(64J^2)$, then the regret of EALP2 satisfies $R_{\rm EALP2}(T,B) = O(1)$.
\end{theorem}

Note that here we assume $\delta$ is known for the simplicity of presentation. When considering practical scenarios where $\delta$ is unknown, we can obtain a lower confidence bound of $\delta$ as follows. At round $t$, let $\hat{q}_{j,t} = \sum_{j'=1}^j \hat{\pi}_{j,t}$ be the empirical estimate of the cumulative probability. Further, let $\tilde{j}'_t(\rho) = \max \{j: \hat{q}_{j,t} \leq \rho\}$ be the threshold under the empirical estimate. Let $\hat{\delta}_{t} = \min\{\hat{q}_{\tilde{j}'_t(\rho) + 1} - \rho, \rho - \hat{q}_{\tilde{j}'_t(\rho)}\}$ and $\check{\delta}_{t} = \frac{1}{2}\hat{\delta}_{t}$. Then $\check{\delta}_{t}$ is a lower confidence bound of $\delta$ with $\mathbb{P}\{\delta \geq  \check{\delta}_{t}\} \geq 1 - e^{-2 \check{\delta}_{t}^2 t}$. We choose $T_1$ which is the smallest $t$ such that $e^{-2 \check{\delta}_{t}^2 t} \leq \frac{1}{T^2}$ and $t \geq 16 J^2\log^3 T/\check{\delta}_{t}^2$. Then the following analysis holds, while the regret due to the event that $\delta \geq  \check{\delta}_{t}$ will be $O(1)$. Moreover, such a $t$ will appear with high probability after $64 J^2\log^3 T/\delta^2$ rounds for the non-boundary cases, and not appear with high probability for the boundary cases.

Similar to the non-boundary cases in Theorem~\ref{thm:alp_regret}, the key idea of proving Theorem~\ref{thm:ealp2_regret} is to show that under EALP2, the average remaining budget $b_\tau/\tau$ will not cross the boundaries with high probability. To achieve this, we examine the expectation of $b_\tau/\tau$ and its concentration properties under EALP2.

\textbf{Step 1: Estimation error of $\hat{\boldsymbol{\pi}}_{T_1}$.} Let $\alpha = \delta/(4J\log T)$. According to Hoeffding-Chernoff bound, we have
\begin{eqnarray}
\mathbb{P}\{|\hat{\pi}_{j,T_1} - {\pi}_j| \leq \alpha, \forall j\} \geq 1 - \frac{2J}{T^2}.
\end{eqnarray}

\textbf{Step 2: Bound on the expectation of $b_{\tau}/\tau$.}
\begin{lemma} \label{thm:bounded_exp}
Assume $|\hat{\pi}_{j,T_1} - \pi_j| \leq \alpha$ for all $j$. Then, for all $T_1 \leq t \leq T$, the expectation of the average remaining budget satisfies
\begin{eqnarray}
|\mathbb{E}[b_{\tau}/\tau] - \rho| \leq \frac{\delta}{2},
\end{eqnarray}
where $\tau = T- t + 1$.
\end{lemma}
\begin{proof}
First, we note that the average remaining budget $b_{\tau}/\tau$ is close to the initial value $\rho$ at round $T_1$, because we can verify that for all $t \leq T_1$,
\begin{eqnarray} \label{eq:empi_dist_error}
\rho - \frac{\delta}{4}\leq \frac{B - T_1}{T-T_1+1} \leq \frac{b_{\tau}}{\tau} \leq \frac{B}{T-T_1+1} \leq \rho + \frac{\delta}{4}.
\end{eqnarray}

Now we show by mathematical induction that, if $|\hat{\pi}_{j,T_1} - \pi_j| \leq \alpha$ for all $j$, then $|\mathbb{E}[b_{\tau}/\tau] - \rho| \leq J\alpha\sum_{\tau' = T-T_1 + 1}^\tau\frac{1}{\tau'} + \frac{\delta}{4}$ for $\tau \leq T - T_1 + 1$ (i.e., $t \geq T_1$).

Specifically, for $t = T_1$, we have $\big|\mathbb{E}[\frac{b_{T-T_1 + 1}}{T-T_1+1}] - \rho\big|  \leq \frac{\delta}{4}$ according to Eq.~\eqref{eq:empi_dist_error}. For any given $t \geq T_1$, we have $\tau = T- t + 1$, and
\begin{eqnarray}
&&\mathbb{E}[b_{\tau - 1}|b_{\tau} = b] \nonumber \\
 &=& b - \bigg(\frac{b/\tau - \sum_{j\leq \tilde{j}(b/\tau)}\hat{\pi}_{j, T_1}}{\hat{\pi}_{\tilde{j}(b/\tau)+1, T_1}} \pi_{\tilde{j}(b/\tau)+1} + \sum_{j\leq \tilde{j}(b/\tau)} \pi_j\bigg) \nonumber \\
&=& b- b/\tau + \frac{b/\tau - \sum_{j\leq \tilde{j}(b/\tau)}\hat{\pi}_{j, T_1}}{\hat{\pi}_{\tilde{j}(b/\tau)+1, T_1}} (\hat{\pi}_{\tilde{j}(b/\tau)+1, T_1} - \pi_{\tilde{j}(b/\tau)+1}) +  \sum_{j\leq \tilde{j}(b/\tau)} (\hat{\pi}_{j,T_1} - \pi_j).\nonumber
\end{eqnarray}
Note that $0 < \frac{b/\tau - \sum_{j\leq \tilde{j}(b/\tau)}\hat{\pi}_{j, T_1}}{\hat{\pi}_{\tilde{j}(b/\tau)+1, T_1}} < 1$. Thus, $\big|\mathbb{E}[b_{\tau - 1}|b_{\tau} = b] - \frac{b(\tau-1)}{\tau}\big| \leq J\alpha$, implying that $\big|\mathbb{E}[\frac{b_{\tau - 1}}{\tau-1}|b_{\tau} = b] - \frac{b}{\tau}\big| \leq \frac{J\alpha}{\tau-1}$.
If $\big|\mathbb{E}[b_{\tau}/\tau] - \rho| \leq J\alpha\sum_{\tau' = T-T_1 + 1}^\tau\frac{1}{\tau'} + \frac{\delta}{4}$ for $2 \leq \tau \leq T - T_1 + 1$, then $\big|\mathbb{E}[b_{\tau-1}/(\tau-1)] - \rho| \leq J\alpha\sum_{\tau' = T-T_1 + 1}^\tau\frac{1}{\tau'} + \frac{J\alpha}{\tau-1} +   \frac{\delta}{4}$.

The conclusion of Lemma~\ref{thm:bounded_exp} then follows because $|\mathbb{E}[b_{\tau}/\tau] - \rho| \leq J\alpha\sum_{\tau' = T-T_1 + 1}^\tau\frac{1}{\tau'} + \frac{\delta}{4} \leq J\alpha \log T + \frac{\delta}{4} \leq \frac{\delta}{2}$.
\end{proof}

\textbf{Step 3: Concentration of $b_{\tau}/\tau$.} The next lemma shows the concentration of the average remaining budget $b_\tau/\tau$.
\begin{lemma} \label{thm:concentration_aver_remain_budget}
Assume $|\hat{\pi}_{j,T_1} - \pi_j| \leq \alpha$ for all $j$. The average remaining budget $b_\tau/\tau$ in round $t = T- \tau + 1$ satisfies
\begin{eqnarray}\label{eq:concentration_aver_remain_budget}
\mathbb{P}\big\{|\frac{b_\tau}{\tau} - \mathbb{E}[\frac{b_\tau}{\tau}]| \geq \delta/4\big\} \leq 2\exp\big(-\frac{\delta^2\tau}{32}\big).
\end{eqnarray}
\end{lemma}

To show the concentration of $b_{\tau}/\tau$, we first use the coupling argument to show the following lemma and then use the method of averaged bounded differences \cite{Dubhashi2009Concentration}.
\begin{lemma} \label{thm:bounded_diff}
Assume $|\hat{\pi}_{j,T_1} - \pi_j| \leq \alpha$ for all $j$. The remaining budget $b_\tau$ in round $t = T- \tau + 1$ satisfies
\begin{eqnarray}
\big|\mathbb{E}[b_\tau | \boldsymbol{Z}_{t'-1}, Z_{t'} =1] - \mathbb{E}[b_{\tau}|\boldsymbol{Z}_{t'-1}, Z_{t'} = 0]\big| \leq 2\big(\frac{\tau}{T- t'}\big)^{1-\sigma}, ~~T_1 \leq t' < t.
\end{eqnarray}
where $\boldsymbol{Z}_{t'-1} = (Z_1, Z_2, \ldots, Z_{t'-1})$ and $\sigma = 1- \min_j \frac{\pi_j}{\pi_j + \alpha}$.
\end{lemma}
\begin{proof}
We bound the difference by constructing a coupling $\mathcal{M}$ of the two conditional distributions $(\cdot|\boldsymbol{Z}_{t'-1}, Z_{t'} =1)$ and $(\cdot|\boldsymbol{Z}_{t'-1}, Z_{t'} =0)$. Let $\zeta_{t'+1}, \zeta_{t'+2}, \ldots, \zeta_{T-\tau}$ and $\zeta'_{t'+1}, \zeta'_{t'+2}, \ldots, \zeta'_{T-\tau}$ be the pair of random variables in the coupling $\mathcal{M}$. We construct the coupling as follows:

\textbf{Coupling:} We generate the value of $\zeta_{t''}$'s and $\zeta'_{t''}$'s sequentially. For each $t'' > t'$, let $\tilde{b}_{T-t''+1} = B - 1-\sum_{s = 1}^{t'-1}Z_{s} - \sum_{s = t'+1}^{t''-1}\zeta_{s}$ and $\tilde{b}'_{T-t''+1} = B -\sum_{s = 1}^{t'-1}Z_{s} - \sum_{s = t'+1}^{t''-1}\zeta'_{s}$ be the remaining budgets in round $t''$ corresponding to the pair of random variables. For $\zeta_{t''}$, We pick its value randomly with distribution $\mathbb{P}\{\zeta_{t''} = 1\} = h\big(\frac{\tilde{b}_{T-t''+1}}{T-t''+1}\big)$ and $\mathbb{P}\{\zeta_{t''} = 0\} = 1-h\big(\frac{\tilde{b}_{T-t''+1}}{T-t''+1}\big)$, where $h(\rho)$ is the probability that one unit of budget will be consumed under EALP2 when the average remaining budget is $\rho$, i.e.,
\begin{eqnarray} \label{eq:pull_prob_est}
h(\rho) = \frac{\rho - \sum_{j' \leq \tilde{j}(\rho)}\hat{\pi}_{j,T_1}}{\hat{\pi}_{\tilde{j}(\rho)+1, T_1}}\pi_{\tilde{j}(\rho)+1} + \sum_{j' \leq \tilde{j}(\rho)}\pi_j.
\end{eqnarray}
For $\zeta'_{t''}$, we generate its value conditioned on $\zeta_{t''}$. If $\tilde{b}'_{T-t''+1} = \tilde{b}_{T-t''+1}$, then $\zeta'_{t''} = \zeta_{t''}$. If $\tilde{b}'_{T-t''+1} = \tilde{b}_{T-t''+1} + 1$, then
\begin{eqnarray}
&&\mathbb{P}\big\{\zeta'_{t''} = \zeta_{t''}|\tilde{b}'_{T-t''+1} = \tilde{b}_{T-t''+1}\big\} = 1, \nonumber\\
&&\mathbb{P}\big\{\zeta'_{t''} = 1|\tilde{b}'_{T-t''+1} = \tilde{b}_{T-t''+1}+1, \zeta_{t''} = 1\big\} = 1, \nonumber \\
&&\mathbb{P}\big\{\zeta'_{t''} = 1|\tilde{b}'_{T-t''+1} = \tilde{b}_{T-t''+1}+1, \zeta_{t''} = 0\big\} = \frac{h\big(\frac{\tilde{b}'_{T-t''+1}}{T-t''+1}\big) - h\big(\frac{\tilde{b}_{T-t''+1}}{T-t''+1}\big)}{1- h\big(\frac{\tilde{b}_{T-t''+1}}{T-t''+1}\big)}.\nonumber
\end{eqnarray}
Note that according to the above construction, $\tilde{b}'_{T-t''+1}- \tilde{b}_{T-t''+1}$ could only be 0 or 1. We can verify that the marginals satisfy
\begin{eqnarray}
(\zeta_{t''}, t'' > t')\sim (Z_{t''}, t'' > t'|\boldsymbol{Z}_{t'-1}, Z_{t'} =1), \nonumber
\end{eqnarray}
and
\begin{eqnarray}
(\zeta'_{t''}, t'' > t')\sim (Z_{t''}, t'' > t'|\boldsymbol{Z}_{t'-1}, Z_{t'} =0). \nonumber
\end{eqnarray}

From the construction of the coupling, we know that $\tilde{b}'_{\tau}- \tilde{b}_{\tau} = 1$ if and only if $\zeta_{t''} = \zeta'_{t''}$ for all $t' <t''  \leq T-\tau$. Thus,
\begin{eqnarray}\label{eq:bounded_diff_prod}
&&\big|\mathbb{E}[b_\tau | \boldsymbol{Z}_{t'-1}, Z_{t'} =1] - \mathbb{E}[b_{\tau}|\boldsymbol{Z}_{t'-1}, Z_{t'} = 0]\big| \nonumber\\
&=& \mathbb{P}\big\{\zeta_{t''} = \zeta'_{t''},  t' <t''  \leq T-\tau\big\} \nonumber\\
&=& \prod_{t'' = t'+1}^{T-\tau}\mathbb{P}\big\{\zeta_{t''} = \zeta'_{t''}|\zeta_{s} = \zeta'_{s},  t' < s  \leq t''-1\big\}.
\end{eqnarray}

We show that each term in Eq.~\eqref{eq:bounded_diff_prod} can be bounded as follows.
\begin{lemma} \label{thm:bounded_diff_single_term}
The coupling $\mathcal{M}$ satisfies
\begin{eqnarray}
\mathbb{P}\big\{\zeta_{t''} = \zeta'_{t''}|\zeta_{s} = \zeta'_{s},  t' < s  \leq t''-1\big\} \leq 1 - \frac{1-\sigma}{T-t''+1}, \nonumber
\end{eqnarray}
where $\sigma = 1-\min_j \frac{\pi_j}{\pi_j + \alpha}$.
\end{lemma}
\begin{proof}
Conditioned on $\zeta_{s} = \zeta'_{s},  t' < s  \leq t''-1$, we have $\tilde{b}'_{T-t''+1} = \tilde{b}_{T-t''+1} + 1$, and $\zeta_{t''} \neq \zeta'_{t''}$ i.f.f. $\zeta'_{t''} = 0$ and $\zeta_{t''} = 1$. Thus,
\begin{align}
\mathbb{P}\big\{\zeta_{t''} = \zeta'_{t''}|\zeta_{s} = \zeta'_{s},  t' < s  \leq t''-1\big\}
&=1- \frac{h\big(\frac{\tilde{b}'_{T-t''+1}}{T-t''+1}\big) - h\big(\frac{\tilde{b}_{T-t''+1}}{T-t''+1}\big)}{1- h\big(\frac{\tilde{b}_{T-t''+1}}{T-t''+1}\big)}\bigg[1 - h\big(\frac{\tilde{b}_{T-t''+1}}{T-t''+1}\big)\bigg] \nonumber\\
&= 1- \bigg[h\big(\frac{\tilde{b}'_{T-t''+1}}{T-t''+1}\big) - h\big(\frac{\tilde{b}_{T-t''+1}}{T-t''+1}\big) \bigg].
\end{align}
To prove Lemma~\ref{thm:bounded_diff_single_term}, it suffices to show that for any $b$ and $\tau$ satisfying $b \leq \tau -1$, we have $h\big(\frac{b+1}{\tau}\big) - h\big(\frac{b}{\tau}\big) \geq \gamma/\tau$, where $\gamma = \min_j \frac{\pi_j}{\pi_j + \alpha}$.

Specifically, from the definition of $h(\cdot)$ in Eq.~\eqref{eq:pull_prob_est}, we know that if $\tilde{j}\big(\frac{b+1}{\tau}\big) = \tilde{j}\big(\frac{b}{\tau}\big)$, we have $h\big(\frac{b+1}{\tau}\big) - h\big(\frac{b}{\tau}\big) = \frac{\pi_{\tilde{j}(\frac{b}{\tau})}}{\hat{\pi}_{\tilde{j}(\frac{b}{\tau}), T_1}}\cdot \frac{1}{\tau} \geq \gamma/\tau$.


If $\tilde{j}\big(\frac{b+1}{\tau}\big)  > \tilde{j}\big(\frac{b}{\tau}\big)$,  we have  $\frac{b}{\tau} - \sum_{j\leq \tilde{j}(\frac{b}{\tau})} \hat{\pi}_{j,T_1} < \hat{\pi}_{\tilde{j}(\frac{b}{\tau})+1,T_1}$ and $\frac{b+1}{\tau} - \sum_{j\leq \tilde{j}(\frac{b+1}{\tau})} \hat{\pi}_{j,T_1} > 0$. Therefore,
\begin{align}
&\quad h\big(\frac{b+1}{\tau}\big) - h\big(\frac{b}{\tau}\big) \nonumber\\
&= \sum_{j = \tilde{j}(\frac{b}{\tau}) + 1}^{\tilde{j}(\frac{b+1}{\tau})}\pi_{j} + \frac{\frac{b+1}{\tau} - \sum_{j\leq \tilde{j}(\frac{b+1}{\tau})} \hat{\pi}_{j,T_1}}{\hat{\pi}_{\tilde{j}(\frac{b+1}{\tau})+1,T_1}}\pi_{\tilde{j}(\frac{b+1}{\tau})+1} - \frac{\frac{b}{\tau} - \sum_{j\leq \tilde{j}(\frac{b}{\tau})} \hat{\pi}_{j,T_1}}{\hat{\pi}_{\tilde{j}(\frac{b}{\tau})+1,T_1}}\pi_{\tilde{j}(\frac{b}{\tau})+1}\nonumber\\
&= \sum_{j = \tilde{j}(\frac{b}{\tau})+2}^{\tilde{j}(\frac{b+1}{\tau})}\pi_{j} +\frac{\pi_{\tilde{j}(\frac{b}{\tau})+1}}{\hat{\pi}_{\tilde{j}(\frac{b}{\tau})+1,T_1}}\bigg[\hat{\pi}_{\tilde{j}(\frac{b}{\tau})+1,T_1} - \big(\frac{b}{\tau} - \sum_{j\leq \tilde{j}(\frac{b}{\tau})} \hat{\pi}_{j,T_1}\big) \bigg]+
\frac{\pi_{\tilde{j}(\frac{b+1}{\tau})+1}}{\hat{\pi}_{\tilde{j}(\frac{b+1}{\tau})+1,T_1}}\bigg[\frac{b+1}{\tau} - \sum_{j\leq \tilde{j}(\frac{b+1}{\tau})} \hat{\pi}_{j,T_1}\bigg]\nonumber\\
&\geq \sum_{j = \tilde{j}(\frac{b}{\tau})+2}^{\tilde{j}(\frac{b+1}{\tau})}\frac{\pi_j}{\hat{\pi}_{j,T_1}}\hat{\pi}_{j,T_1} + \gamma \bigg(\frac{1}{\tau} -  \sum_{j = \tilde{j}(\frac{b}{\tau})+2}^{\tilde{j}(\frac{b+1}{\tau})}\hat{\pi}_{j,T_1}\bigg) \geq \gamma/\tau.
\end{align}
\end{proof}
Using Lemma~\ref{thm:bounded_diff_single_term}, we have
\begin{eqnarray}
\big|\mathbb{E}[b_\tau | \boldsymbol{Z}_{t'-1}, Z_{t'} =1] - \mathbb{E}[b_{\tau}|\boldsymbol{Z}_{t'-1}, Z_{t'} = 0]\big| &\leq& \prod_{t'' = t'+1}^{T-\tau}\big(1 - \frac{1-\sigma}{T-t''+1}\big) \nonumber\\
&\overset{(\rm a)}{=}&\frac{\tau + \sigma}{T-t'}\prod_{s = 1}^{T-t'-\tau-1}\big(1+\frac{\sigma}{\tau+s} \big)\nonumber\\
&\overset{(\rm b)}{\leq}&\frac{\tau + \sigma}{T-t'}\big(\frac{T-t'-1}{\tau} \big)^{\sigma} \nonumber \\
&\leq &2\big(\frac{\tau}{T-t'}\big)^{1-\sigma}. \nonumber
\end{eqnarray}
Equality (a) is obtained by merging the numerator of each term with the denominator of the next term. Inequality (b) is true because $\sigma < 1$, and
\begin{eqnarray}
\log \prod_{s = 1}^{T-t'-\tau-1}\big(1+\frac{\sigma}{\tau+s} \big) = \sum_{s = 1}^{T-t'-\tau-1}\log \big(1+\frac{\sigma}{\tau+s} \big) \leq\sum_{s = 1}^{T-t'-\tau-1}\frac{\sigma}{\tau+s}  \leq \sigma \log\big(\frac{T-t'-1}{\tau} \big). \nonumber
\end{eqnarray}
\end{proof}
To use the method of averaged bounded differences  \cite{Dubhashi2009Concentration}, we note that
\begin{eqnarray}
\sum_{t' = T_1}^{T-\tau}\bigg[2\big(\frac{\tau}{T- t'}\big)^{1-\sigma}\bigg]^2 \leq 4\tau^{2-2\alpha}\cdot\big[\frac{1}{\tau^{1-2\sigma}} - \frac{1}{(T-T_1+1)^{1-2\sigma}}\big]\leq 4\tau.
\end{eqnarray}

Then, according to  Corollary~5.1 in \cite{Dubhashi2009Concentration} and Lemma~\ref{thm:bounded_diff}, we have
\begin{eqnarray}
\mathbb{P}\big\{|b_\tau - \mathbb{E}[b_\tau]| \geq \delta\tau/4\big\} \leq 2\exp\big(-\frac{2(\delta\tau/4)^2}{4\tau}\big)= 2\exp\big(-\frac{\delta^2\tau}{32}\big). \nonumber
\end{eqnarray}
implying Eq.~\eqref{eq:concentration_aver_remain_budget} in Lemma~\ref{thm:concentration_aver_remain_budget}.

\textbf{Step 4: Upper bound of $R_{\rm EALP2}(T,B)$.} Now we bound the regret $R_{\rm EALP2}(T,B)$ using the results obtained in the previous steps.  We analyze the event of ``boundary-crossing'' in round $t$, denoted as $\mathcal{E}_{{\rm cross},t}$, which is the event that $\tilde{j}(b_\tau/\tau) \neq \tilde{j}(\rho)$. The event of ``boundary-crossing'' may happen when the estimates of empirical distribution is inaccurate or the average remaining budget $b_\tau/\tau$ deviates far from $\rho$. We study the probability of $\mathcal{E}_{{\rm cross},t}$ for $t \leq T_1$ and $t > T_1$, respectively.

For $t \leq T_1$, the average remaining budget satisfies $\rho - \delta/4 \leq b_\tau/\tau \leq \rho + \delta/4$, as discussed in Step~2. The event  $\mathcal{E}_{{\rm cross},t}$ may occur only when there is some $j$ such that $|\hat{\pi}_{j,t} - \pi_j| \geq \delta/(4J)$. Thus,
\begin{equation}
\mathbb{P}\{\mathcal{E}_{{\rm cross},t}\} \leq \mathbb{P}\{\exists j,  |\hat{\pi}_{j,t} - \pi_j| \geq \delta/(4J)\}\leq 2J\exp(-\delta^2t/8J^2), ~~t \leq T_1.
\end{equation}

For $t > T_1$, if the empirical distribution $|\hat{\pi}_{j,T_1} - \pi_j| \leq \alpha$ ($< \delta/(4J)$ for sufficiently large $T$) for all $j$, then the average remaining budget satisfies $\mathbb{P}\big\{|\frac{b_\tau}{\tau} - \mathbb{E}[\frac{b_\tau}{\tau}]| \geq 3\delta/4\big\} \leq 2\exp\big(-\frac{\delta^2\tau}{32}\big)$ due to Lemma~\ref{thm:bounded_exp} and Lemma~\ref{thm:concentration_aver_remain_budget}. Thus,
\begin{eqnarray}
\mathbb{P}\{\mathcal{E}_{{\rm cross},t}\} &\leq& \mathbb{P}\{\exists j,  |\hat{\pi}_{j,T_1} - \pi_j| \geq \alpha\} + \mathbb{P}\big\{|\frac{b_\tau}{\tau} - \mathbb{E}[\frac{b_\tau}{\tau}]| \geq 3\delta/4|\forall j,  |\hat{\pi}_{j,T_1} - \pi_j| \geq \alpha\big\}\nonumber\\
&\leq& \frac{2J}{T^2} + 2\exp\big(-\frac{\delta^2\tau}{32}\big), ~~t > T_1.
\end{eqnarray}

Now we bound the expectation of $C_j(T)$, i.e., the number of executions under context $j$.

For $j \leq \tilde{j}(\rho)$,
\begin{eqnarray}
\mathbb{E}[C_j(T)]& =& \mathbb{E}\bigg[\sum_{t=1}^T \mathds{1}(X_t=j, A_t=k_j^*)\bigg] \nonumber\\
&\geq& \sum_{t=1}^T \mathbb{P}\{X_t=j, A_t=k_j^*|\urcorner\mathcal{E}_{{\rm cross},t}\} \mathbb{P}\{\urcorner\mathcal{E}_{{\rm cross},t}\} \nonumber \\
&= & \sum_{t=1}^T\pi_j\big(1- \mathbb{P}\{\mathcal{E}_{{\rm cross},t}\}\big)\nonumber\\
&\geq& \pi_j T - \sum_{t=1}^{T_1}2J\exp(-\delta^2t/8J^2) - \sum_{t=T_1+1}^{T}\big[\frac{2J}{T^2} + 2\exp\big(-\frac{\delta^2\tau}{32}\big)\big] = \pi_j T + O(1), \nonumber
\end{eqnarray}
and
\begin{eqnarray}
\mathbb{E}[C_j(T)]\leq \pi_j T, \nonumber
\end{eqnarray}
Similarly, for $j > \tilde{j}(\rho)+1$, we have
\begin{eqnarray}
\mathbb{E}[C_j(T)] \leq \sum_{t=1}^T \mathbb{P}\{X_t=j, A_t=k_j^*|\mathcal{E}_{{\rm cross},t}\} \mathbb{P}\{\mathcal{E}_{{\rm cross},t}\} = O(1). \nonumber\\
\end{eqnarray}
For $j = \tilde{j}(\rho)+1$, we have
\begin{eqnarray}
\mathbb{E}[C_j(T)] = \mathbb{E}[B-b_0] - \sum_{j \neq \tilde{j}(\rho)+1}\mathbb{E}[C_j(t)] \geq B-T\sum_{j \leq \tilde{j}(\rho)}\pi_j -O(1) = \big(\rho - \sum_{j \leq \tilde{j}(\rho)}\pi_j\big)T - O(1). \nonumber
\end{eqnarray}
We complete the proof of Theorem~\ref{thm:ealp2_regret} by summing over all contexts:
\begin{eqnarray}
U_{\rm EALP2}(T,B) = \sum_{j=1}^J\mu_j^*\mathbb{E}[C_j(T)]\geq T\tilde{v}(\rho) - O(1) = \widehat{U}(T,B) - O(1). \nonumber
\end{eqnarray} 

\section{Constrained Contextual Bandits with Heterogeneous Costs} \label{app:het_cost}

In this section, we consider the case where the cost for each action $k$ under context $j$ is fixed at $c_{j,k}$, which may be different for different $j$ and $k$. We discuss how to use the insight from unit-cost systems in  heterogeneous-cost systems.

\subsection{Approximation of the Oracle Algorithm}
Similar to unit-cost systems, we first study the case with known statistics. We generalize the upper bound and the ALP algorithm in Section~\ref{sec:oracle_solution} to general-cost systems.

\subsubsection{Upper Bound}
 With known statistics, the agent knows the context distribution $\pi_j$'s, the costs $c_{j,k}$'s, and the expected rewards $u_{j,k}$'s.
In heterogeneous-cost systems, the quality of a context-action pair $(j,k)$ is roughly captured by the normalized reward, denoted by $\eta_{j,k} = u_{j,k}/c_{j,k}$.
However, unlike the unit-cost case, the agent \emph{cannot} only focus on the ``best'' action with highest normalized reward, i.e., $k_j^* = \argmax_k \eta_{j,k}$, when making a decision under context $j$. This is because there may exist another action $k$ such that $\eta_{j,k} < \eta_{j,k_{j}^*}$, but $u_{j,k} > u_{j,k_j^*}$ (and of course, $c_{j,k} > c_{j, k_j^*}$). If there is sufficient budget allocated for context $j$, then the agent may take action $k$ to maximize the expected reward.
Therefore, the agent needs to consider all actions under each context. Let $p_{j,k}$ be the probability that action $k$ is taken under context $j$. We define the following LP problem:
\begin{align}
({\mathcal{LP}}'_{T,B})~~\text{maximize} &~~ \sum_{j = 1}^J \pi_j \sum_{k = 1}^K p_{j,k} u_{j,k},  \label{eq:LP_general_obj} \\
\text{subject to} &~~ \sum_{j = 1}^J \pi_j \sum_{k = 1}^K p_{j,k} c_{j,k} \leq B/T, \label{eq:LP_general_cost_const}\\
&~~\sum_{k = 1}^K p_{j,k} \leq 1, ~\forall j, \label{eq:LP_intra_cont_const} \\
&~~ p_{j,k} \in [0,1].\nonumber
\end{align}
The above LP problem ${\mathcal{LP}}'_{T,B}$ can be solved efficiently by optimization tools. Let $\hat{v}(\rho)$ be the maximum value of ${\mathcal{LP}}'_{T,B}$.
Similar to Lemma~\ref{thm:upper_bound}, we can show that $T\hat{v}(\rho)$ is an upper bound of the expected total reward, i.e., $T\hat{v}(\rho) \geq U^*(T,B)$.

To obtain insight from the solution of ${\mathcal{LP}}'_{T,B}$, we derive an explicit representation for the solution by analyzing the structure of ${\mathcal{LP}}'_{T,B}$. Note that there are two types of (non-trivial) constraints in ${\mathcal{LP}}'_{T,B}$, one is the ``inter-context'' budget constraint \eqref{eq:LP_general_cost_const}, the other is the ``intra-context'' constraint \eqref{eq:LP_intra_cont_const}. These constraints can be decoupled by first allocating budget for each context, and then solving a subproblem with the allocated budget constraint for each context. Specifically, let $\rho_j$ be the budget allocated to context $j$, then ${\mathcal{LP}}'_{T,B}$ can be decomposed as follows:
\begin{align}
~~\text{maximize} &~~ \sum_{j = 1}^J \pi_j \hat{v}_j(\rho_j),  \nonumber\\
\text{subject to} &~~ \sum_{j = 1}^J \pi_j \rho_j \leq  B/T,  \nonumber
\end{align}
where
\begin{align}
(\mathcal{SP}_j)~~v_j(\rho_j) = \text{maximize} &~ \sum_{k = 1}^K p_{j,k} u_{j,k},  \\
\text{subject to} &~~ \sum_{k = 1}^K p_{j,k} c_{j,k} \leq \rho_j, \\
&~~\sum_{k = 1}^K p_{j,k} \leq 1,   \\
&~~ p_{j,k} \in [0,1].\nonumber
\end{align}
Next, by analyzing sub-problem $\mathcal{SP}_j$, we show that some actions can be deleted without affecting the performance, i.e., the probability is 0 in the optimal solution.
\begin{lemma} \label{thm:intra_context_LP}
For any given $\rho_j \geq 0$, there exists an optimal solution of $\mathcal{SP}_j$, i.e.,  $\boldsymbol{p}^*_j = (p^*_{j, 1}, p^*_{j, 2}, \ldots, p^*_{j,K})$, satisfies:\\
(1) For $k_1$, if there exists another action $k_2$, such that $\eta_{j, k_1} \leq \eta_{j, k_2}$ and $u_{k_1} \leq u_{k_2}$, then $p^*_{j, k_1} = 0$;\\
(2) For $k_1$, if there exists two actions $k_2$ and $k_3$, such that $\eta_{j, k_2} \leq \eta_{j, k_1} \leq \eta_{j, k_3}$, $u_{j,k_2}\geq u_{j,k_1} \geq u_{j,k_3}$, and $\frac{u_{j, k_1} - u_{j,k_3}}{c_{j, k_1} - c_{j,k_3}} \leq \frac{u_{j, k_2} - u_{j,k_3}}{c_{j, k_2} - c_{j,k_3}}$, then $p^*_{j, k_1} = 0$.
\end{lemma}
Intuitively, the first part of Lemma~\ref{thm:intra_context_LP} shows that if an action has small normalized and original expected reward, then it can be removed. The second part of Lemma~\ref{thm:intra_context_LP} shows that if an action has small normalized expected reward and medium original expected reward, but the increasing rate is smaller than another action with larger expected reward, then it can also be removed.
\begin{proof}
The key idea of this proof is that, if the conditions is satisfied, and there is a feasible solution $\boldsymbol{p}_j = (p_{j,1}, p_{j,2}, \ldots, p_{j,K})$ such that $p_{j,k_1} > 0$, then we can construct another feasible solution $\boldsymbol{p}'_j$ such that $p'_{j,k_1} = 0$, without reducing the objective value $v_j(\rho_j)$.

We first prove part (1). Under the conditions of part $(1)$, if $\boldsymbol{p}_j$ is a feasible solution of $\mathcal{SP}_j$ with $p_{j,k_1} > 0$, then consider another solution $\boldsymbol{p}'_j$, where $p'_{j,k} = p_{j,k}$ for $k \notin \{k_1, k_2\}$, $p'_{j, k_1} = 0$, and $p'_{j, k_2} = p_{j,k_2} + p_{j,k_1} \min\{\frac{c_{j,k_1}}{c_{j,k_2}}, 1\}$. Then, we can verify that $\boldsymbol{p}'_j$ is a feasible solution of $(\mathcal{SP}_j)$, and the objective value under $\boldsymbol{p}'_j$ is no less than that under $\boldsymbol{p}_j$.

For the second part, if the conditions are satisfied and $p_{j,k_1} > 0$, then we construct a new solution $\boldsymbol{p}'_j$ by re-allocating the budget consumed by action $k_1$ to actions $k_2$ and $k_3$, without violating the constraints. Specifically, we set the probability the same as the original solution for other actions, i.e., $p'_{j,k} = p_{j,k}$ for $k \notin \{k_1, k_2, k_3\}$, and set $p'_{j,k_1} = 0$ for action $k_1$. For $k_2$ and $k_3$, to maximize the objective function, we would like to allocate as much budget as possible to $k_3$ unless there is remaining budget. Therefore, we set
$p'_{j,k_2} = p_{j,k_2}$ and $p'_{j,k_3} = p_{j,k_3} + \frac{p_{j,k_1}c_{j,k_1}}{c_{j,k_3}}$, if $\sum_{k \neq k_1} p_{j,k} + \frac{p_{j,k_1}c_{j,k_1}}{c_{j,k_3}} \leq 1$; or,
$p'_{j,k_2} = p_{j,k_2} + \frac{p_{j,k_1}c_{j,k_1} - (1 - \sum_{k\neq k_1}p_{j,k})c_{j,k_3}}{c_{j,k_2} - c_{j,k_3}}$ and $p'_{j,k_3} = p_{j,k_3} + \frac{(1 - \sum_{k\neq k_1}p_{j,k})c_{j,k_2} - p_{j,k_1}c_{j,k_1}}{c_{j,k_2} - c_{j,k_3}}$, if $\sum_{k \neq k_1} p_{j,k} + \frac{p_{j,k_1}c_{j,k_1}}{c_{j,k_3}} > 1$. We can verify that $\boldsymbol{p}_j$ satisfies the constraints of $(\mathcal{SP}_j)$ but the objective value is no less than that under $\boldsymbol{p}_j$.
\end{proof}

With Lemma~\ref{thm:intra_context_LP}, the agent can ignore some actions that will obviously be allocated with zero probability under a given context $j$. We call the set of the remaining actions as \emph{candidate set} for context $j$, denoted as $\mathcal{A}_j$. We propose an algorithm to construct the candidate action set for context $j$, as shown in Algorithm~\ref{alg:find_candidates}.

\begin{algorithm}[htbp]
\caption{Find Candidate Set for Context $j$}
\label{alg:find_candidates}
\begin{algorithmic}
\STATE{\bfseries Input:} $c_{j,k}$'s, $u_{j,k}$'s, for all $1 \leq k \leq K$;\\
\STATE{\bfseries Output:} $\mathcal{A}_j$;\\
\STATE{\bfseries Init:} $\mathcal{A}_j = \{1,2,\ldots, K\}$ ;\\
\STATE{Calculate normalized rewards: $\eta_{j,k} = u_{j,k}/c_{j,k}$}; \\
\STATE{Sort actions in descending order of their normalized rewards:}
 \begin{equation}
 \eta_{j,k_{1}} \geq \eta_{j,k_{2}} \geq  \ldots \geq \eta_{j,k_{K}}. \nonumber
 \end{equation}
\FOR{$a  = 2$ {\bfseries to} $K$}
\IF{$\exists a' < a$ such that $u_{j,k_a} \leq u_{j,{k_{a'}}}$}
\STATE{$\mathcal{A}_j = \mathcal{A}_j \backslash \{k_{a}\}$;}
\ENDIF
\ENDFOR
\STATE{$a = 1$;}
\WHILE{$a \leq K - 1$}
\STATE{Find the action with highest increasing rate:
\begin{equation}
a^* = \argmax_{a': a' > a, k_{a'} \in \mathcal{A}_j} \frac{u_{j, k_{a'}} - u_{j, k_{a}}}{c_{j, k_{a'}} - c_{j, k_{a}}}. \nonumber
\end{equation}}
\STATE{Remove the actions in between:
\begin{equation}
\mathcal{A}_j = \mathcal{A}_j \backslash \{k_{a'}: a < a' < a^*\}.\nonumber
\end{equation} }
\STATE{Move to the next candidate action: $a = a^*$; }
\ENDWHILE
\end{algorithmic}
\end{algorithm}

For context $j$, assume that the candidate set $\mathcal{A}_j = \{k_{j,1}, k_{j,2}, \ldots, k_{j, K_j}\}$ has been sorted in descending order of their normalized rewards, i.e., $\eta_{j,k_{j,1}} \geq \eta_{j,k_{j,2}} \geq \ldots \geq \eta_{j, k_{j, K_j}}$. From Algorithm~\ref{alg:find_candidates}, we know that $u_{j, k_{j,1}} < u_{j, k_{j,2}} < \ldots < u_{j, k_{j, K_j}}$, and $ c_{j, k_{j,1}} < c_{j, k_{j,2}} < \ldots < c_{j, k_{j, K_j}}$.

The agent now only needs to consider the actions in the candidate set $\mathcal{A}_j$. To decouple the ``intra-context'' constraint \eqref{eq:LP_intra_cont_const}, we introduce the following transformation:
\begin{eqnarray}
p_{j, k_{j,a}} =
\begin{cases}
\tilde{p}_{j, k_{j,a}} - \tilde{p}_{j, k_{j, a+1}},&\text{if $ 1 \leq a \leq K_j - 1$},\\
\tilde{p}_{j, k_{j, K_j}}, &\text{if $a = K_j$},
\end{cases}\nonumber
\end{eqnarray}
where $\tilde{p}_{j, k_{j,a}} \in [0, 1]$, and $\tilde{p}_{j, k_{j,a}} \geq \tilde{p}_{j, k_{j, a+1}}$ for $ 1 \leq a \leq K_j - 1$.
Substituting the transformations into $({\mathcal{SP}}_j)$ and reorganize it as
\begin{align}
(\widetilde{\mathcal{SP}}_j)~~ \text{maximize} &~ \sum_{a = 1}^{K_j} \tilde{p}_{j,k_{j,a}} \tilde{u}_{j,k_{j,a}}, \nonumber \\
\text{subject to} &~~ \sum_{a = 1}^{K_j} \tilde{p}_{j,k_{j,a}} \tilde{c}_{j,k_{j, a}} \leq \rho_j, \nonumber \\
&~~ \tilde{p}_{j, k_{j,a}} \geq \tilde{p}_{j, k_{j, a+1}}, ~1 \leq a \leq K_j - 1, \label{eq:virtual_lp_ineff}\\
&~~ \tilde{p}_{j,k_{j,a}} \in [0,1], ~\forall a,\nonumber
\end{align}
where
\begin{eqnarray}
\tilde{u}_{j,k_{j,a}} =
\begin{cases}
u_{j, k_{j,1}},&\text{if $a = 1$},\\
u_{j, k_{j,a}} - u_{j, k_{j, a-1}}, &\text{if $2 \leq a \leq K_j$},
\end{cases}\nonumber
\end{eqnarray}
\begin{eqnarray}
\tilde{c}_{j,k_{j, a}} =
\begin{cases}
c_{j, k_{j, 1}},&\text{if $a = 1$},\\
c_{j, k_{j, a}} - c_{j, k_{j, a-1}}, &\text{if $2 \leq a \leq K_j$}.
\end{cases}\nonumber
\end{eqnarray}

Next, we show that the constraint \eqref{eq:virtual_lp_ineff} can indeed be removed.
For each $k_{j,a}$, we can view $\tilde{c}_{j,k_{j,a}}$ and $\tilde{u}_{j,k_{j,a}}$ as the cost and expected reward of a virtual action.
Let $\tilde{\eta}_{j,k_{j, a}} = \tilde{u}_{j,k_{j,a}}/\tilde{c}_{j, k_{j, a}}$ be the normalized expected reward of virtual action $k_{j,a}$.
For $a = 1$, using $\frac{u_{j, k_{j,1}}}{c_{j, k_{j,1}}} \geq \frac{u_{j, k_{j,2}}}{c_{j, k_{j,2}}}$, we can show that $\tilde{\eta}_{j,k_{j,1}} \geq \tilde{\eta}_{j,k_{j,2}}$. For $2 \leq a \leq K_j -1$,  using $\frac{u_{j, k_{j,a}} - u_{j, k_{j,a-1}}}{c_{j, k_{j,a}} - c_{j, k_{j,a-1}}} \geq \frac{u_{j, k_{j,a+1}} - u_{j, k_{j,a-1}}}{c_{j, k_{j,a+1}} - c_{j, k_{j,a-1}}}$, we can show that $\tilde{\eta}_{j,k_{j,a}} \geq \tilde{\eta}_{j,k_{j,a+1}}$. In other words, we can verify that $\tilde{\eta}_{j,k_{j,1}} \geq \tilde{\eta}_{j,k_{j,2}} \geq \ldots \geq \tilde{\eta}_{j, k_{j,K_j}}$. Thus, without constraint \eqref{eq:virtual_lp_ineff},  the optimal solution $\tilde{\boldsymbol{p}}^*_j = [\tilde{p}^*_{j, k_1}, \tilde{p}^*_{j, k_2}, \ldots, \tilde{p}^*_{j, k_{K_j}}]$
automatically satisfies 
$\tilde{p}^*_{j, k_1} \geq \tilde{p}^*_{j, k_2} \geq  \ldots \geq \tilde{p}^*_{j, k_{K_j}}$. Hence, we can remove the constraint \eqref{eq:virtual_lp_ineff}, and thus decouple the probability constraint under a context.

With the above transformations, we can thus rewrite the global LP problem
\begin{align}
(\widetilde{{\mathcal{LP}}}'_{T,B})~~\text{maximize} &~~ \sum_{j = 1}^J \sum_{a = 1}^{K_j}\pi_j \tilde{p}_{j,k_{j,a}} \tilde{u}_{j,k_{j,a}}, \nonumber \\
\text{subject to} &~~ \sum_{j = 1}^J \sum_{a = 1}^{K_j}\pi_j \tilde{p}_{j,k_{j,a}} \tilde{c}_{j,k_{j,a}} \leq B/T, \nonumber \\
&~~ \tilde{p}_{j,k_{j,a}} \in [0,1], ~\forall j, ~\text{and $1\leq a \leq K_j$}.\nonumber
\end{align}
The solution of $\widetilde{{\mathcal{LP}}}'_{T,B}$ follows a threshold structure.
We sort all context-(virtual-)action pairs $(j,k_a)$ in descending order of their normalized expected reward. Let $j^{(i)}, k^{(i)}$ be the context index and action index of the $i$-th pair, respectively. Namely, $\tilde{\eta}_{j^{(1)}, k^{(1)}} \geq \tilde{\eta}_{j^{(2)}, k^{(2)}} \geq \ldots \geq \tilde{\eta}_{j^{(M)}, k^{(M)}}$, where $M = \sum_{j=1}^J K_j$ is the total number of candidate actions for all contexts.
Define a threshold corresponding to $\rho = B/T$,
\begin{eqnarray} 
\tilde{i}(\rho) = \max\{i: \sum_{i' = 1}^i\pi_{j^{(i')}} \tilde{c}_{j^{(i')},k^{(i')}} \leq \rho\},
\end{eqnarray}
where $\rho = B/T$ is the average budget.
We can verify that the following solution is optimal for $\widetilde{{\mathcal{LP}}}'_{T,B}$:
\begin{eqnarray}
\tilde{p}_{j^{(i)}, k^{(i)}}(\rho) =
\begin{cases}
1, &\text{if $1 \leq i \leq  \tilde{i}(\rho)$},\\
\frac{\rho - \sum_{i' = 1}^{\tilde{i}(\rho)}\pi_{j^{(i')}} \tilde{c}_{j^{(i')},k^{(i')}}}{\pi_{j^{(\tilde{i}(\rho)+1)}} \tilde{c}_{j^{(\tilde{i}(\rho)+1)},k^{(\tilde{i}(\rho)+1)}}}, & \text{if $i = \tilde{i}(\rho)+1$},\\
0, & \text{if $i > \tilde{i}(\rho) + 1$}.\\
\end{cases}\nonumber
\end{eqnarray}
Then, the optimal solution of $\widetilde{{\mathcal{LP}}}'_{T,B}$ can be calculated using the reverse transformation from
$\tilde{p}_{j, k}(\rho)$'s to ${p}_{j, k}(\rho)$'s

\subsubsection{ALP Algorithm}
Similar to unit-cost systems, the ALP algorithm replaces the average constraint $B/T$ in ${\mathcal{LP}}'_{T,B}$ with the average remaining budget $b_\tau/\tau$, and obtains probability $p_{j, k}(b_\tau/\tau)$. Under context $j$, the ALP algorithm take action $k$ with probability $p_{j, k}(b_\tau/\tau)$.

Unlike unit-cost systems, the remaining budget $b_{\tau}$ does not follow any classic distribution in heterogeneous-cost systems. However, we can show that the concentration property still holds for this general case by using the method of averaged bounded differences \cite{Dubhashi2009Concentration}.
\begin{lemma}\label{thm:alp_concentrate_general_cost}
For  $0 < \delta < 1$, there exists a positive number $\kappa$, such that under the ALP algorithm, the remaining budget $b_{\tau}$ satisfies
\begin{eqnarray}
\mathbb{P}\{b_{\tau} > (\rho+\delta)\tau \} \leq e^{-\kappa\delta^2 \tau}\nonumber, \\
\mathbb{P}\{b_{\tau} < (\rho-\delta)\tau \} \leq e^{-\kappa\delta^2 \tau}\nonumber.
\end{eqnarray}
\end{lemma}
\begin{proof}
We prove the lemma using the method of averaged bounded differences \cite{Dubhashi2009Concentration}. The process is similar to Section 7.1 in \cite{Dubhashi2009Concentration}, except that we consider the remaining budget and the successive differences of the remaining budget are bounded by $c_{\max}$.

Specifically, let $\tilde{c}_{t'}$, $1 \leq t' \leq T$ be the budget consumed under ALP, and let $\tilde{\boldsymbol{c}}_{t'} = (\tilde{c}_{1}, \tilde{c}_{2}, \ldots, \tilde{c}_{t'})$. Then the remaining budget at round $t$ (the remaining time $\tau = T-t+1$), i.e.,  $b_{T-t+1}$ is a function of $\tilde{\boldsymbol{c}}_t$. We note that under ALP, the expectation of the ratio between the remaining budget and the remaining time does not change, i.e., for any $b\leq \sum_{j = 1}\pi_j c_j^*$ (here $c_j^* = \max_k c_{j,k}$), if $b_{\tau} = b$, then $\mathbb{E}[b_{\tau-1}/(\tau-1)] =b/\tau$. Thus, we can verify that for any $1\leq t' \leq t$, we have
\begin{eqnarray}
\mathbb{E}[b_{T-t+1}|\tilde{\boldsymbol{c}}_{t'}] = b_{T - t' + 1} - \frac{b_{T - t' + 1}}{T-t'+1}(t - t').
\end{eqnarray}
Note that $\Delta b = b_{T - t' + 2} - b_{T - t' + 1}  \leq c_{\max}$ and $b_{T - t' + 2} \geq -c_{\max}$, we have
\begin{eqnarray}
&&\big|\mathbb{E}[b_{T-t+1}|\tilde{\boldsymbol{c}}_{t'}] - \mathbb{E}[b_{T-t+1}|\tilde{\boldsymbol{c}}_{t'-1}]\big| \nonumber \\
&\leq &  \max_{0 \leq \Delta b \leq c_{\max}}\bigg\{\big|\Delta b - \frac{b_{T - t' + 2}}{T - t' + 2}\big|  \bigg\} \frac{T - t+1}{T - t' + 1}\nonumber \\
&\leq & \frac{2 c_{\max}(T - t+1)}{T - t' + 1}.
\end{eqnarray}
Moreover,
\begin{eqnarray}
&&\sum_{t' = 1}^t \big[\frac{2 c_{\max}(T - t+1)}{T - t' + 1}\big]^2 \nonumber \\
& =&  4 c_{\max}^2(T - t+1)^2\sum_{t' = 1}^t\frac{1}{(T - t' + 1)^2}\nonumber \\
& =&  4 c_{\max}^2(T - t+1)^2\sum_{\tau' = T - t+1}^{T}\frac{1}{(\tau')^2}\nonumber \\
&\approx &4 c_{\max}^2(T - t+1)^2\int_{T - t+1}^{T}\frac{1}{(\tau')^2}{\rm d}\tau'\nonumber\\
&= & 4 c_{\max}^2(T - t+1)\frac{t-1}{T}.
\end{eqnarray}
According to Theorem~5.3 in \cite{Dubhashi2009Concentration}, and noting $\tau = T-t+1$, $\mathbb{E}[b_{\tau}] = \rho\tau$, we have
\begin{eqnarray}
\mathbb{P}\{b_{\tau} > \mathbb{E}[b_{\tau}] + \delta \tau \}\leq  e^{-\frac{2T (\delta \rho\tau)^2}{4c_{\max}^2(T-t+1)(t-1)}}\leq e^{-\frac{T \delta^2B^2\tau}{2c_{\max}^2T^2(t-1)}}
\leq e^{-\frac{\delta^2\rho^2}{2c_{\max}^2}\tau},
\end{eqnarray}
and similarly,
\begin{eqnarray}
\mathbb{P}\{b_{\tau} < \mathbb{E}[b_{\tau}] - \delta \tau \}\leq e^{-\frac{\delta^2\rho^2}{2c_{\max}^2}\tau},
\end{eqnarray}
Choosing $\kappa = \frac{\rho^2}{2c_{\max}^2}$ concludes the proof.
\end{proof}
Then, using similar methods in Section~\ref{sec:oracle_solution}, we can show that the generalized ALP algorithm achieves $O(1)$ regret in non-boundary cases, and $O(\sqrt{T})$ regret in boundary cases, where the boundaries are now defined as $Q_i = \sum_{i' = 1}^i\pi_{j^{(i')}} \tilde{c}_{j^{(i')},k^{(i')}}$.

%

\subsection{$\epsilon$-First ALP Algorithm}
When the expected rewards are unknown, it is difficult to combine UCB method with the proposed ALP for general systems.
As a special case, when  all actions have the same cost under a given context, i.e., $c_{j,k} = c_j$ for all $k$ and $j$, the normalized expected reward $\eta_{j,k}$ represents the quality of action $k$ under context $j$. In this case, the candidate set for each context only contains one action, which is the action with the highest expected reward.
Thus, the ALP algorithm for the known statistics case is simple. When the expected rewards are unknown, we can extend the UCB-ALP algorithm by managing the UCB for the normalized expected rewards.

When the costs for different actions under the same context are heterogeneous, it is difficult to combine ALP with the UCB method since the ALP algorithm in this case not only requires the ordering of $\eta_{j,k}$'s, but also the ordering of $u_{j,k}$'s and the ratios $\frac{u_{j,k_1} - u_{j, k_2}}{c_{j,k_1} - c_{j, k_2}}$.
We propose an $\epsilon$-First ALP Algorithm that explores and exploits separately: the agent takes actions under all contexts in the first $\epsilon(T)$ rounds to estimate the expected rewards, and runs ALP based on the estimates in the remaining $T-\epsilon(T)$ rounds.

\begin{algorithm}[htbp]
\caption{$\epsilon$-First ALP}
\label{alg:eps_first_alp}
\begin{algorithmic}
\STATE{\bfseries Input:} Time horizon $T$, budget $B$, exploration stage length $\epsilon(T)$, and $c_{j,k}$'s, for all $j$ and $k$;\\
\STATE{\bfseries Init:} Remaining budget $b = B$;\\
$C_{j,k} = 0$, $\bar{u}_{j,k} = 0$;
\FOR{$t  = 1$ {\bfseries to} $\epsilon(T)$}
\IF{$b > 0$}
\STATE{Take action $A_t = \argmin_{k\in \mathcal{A}} C_{X_t,k}$ (with random tie-breaking);}
\STATE{Observe the reward $Y_{A_t,t}$;}
\STATE{Update counter $C_{X_t,A_t} = C_{X_t,A_t} + 1$; update remaining budget $b = b - c_{X_t,A_t}$;}
\STATE{Update the reward estimate:
\begin{equation}
\bar{u}_{X_t,A_t} = \frac{(C_{X_t,A_t} -1) \bar{u}_{X_t,A_t} + Y_{A_t,t}}{C_{X_t,A_t}}.\nonumber
\end{equation} }
\ENDIF
\ENDFOR
\FOR{$t = \epsilon(T)+1$ {\bfseries to} $T$}
\STATE{Remaining time $\tau = T- t + 1$;}
\IF{$b > 0$}
\STATE{Obtain the probabilities ${p}_{j,k}(b/\tau)$'s by solving the problem $({\mathcal{LP}}'_{\tau,b})$ with $u_{j,k}$ replaced by $\bar{u}_{j,k}$;}
\STATE{Take action $k$ with probability ${p}_{X_t,k}(b/\tau)$;}
\STATE{Remaining budget $b = b - c_{X_t,A_t}$;}
\ENDIF
\ENDFOR
\end{algorithmic}
\end{algorithm}

For the ease of exposition, we assume $c_{j,k_1} \neq c_{j,k_2}$ for any $j$ and $k_1 \neq k_2$ \footnote{For the case with $c_{j,k_1} = c_{j,k_2}$ for some $j$ and $k_1 \neq k_2$ (and ${u}_{j,k_1} \neq {u}_{j,k_2}$), we can correctly remove the suboptimal action with high probability by comparing their empirical rewards $\bar{u}_{j,k_1} = \bar{u}_{j,k_2}$.}, and let $\Delta^{(c)}_{\min}$ be the minimal difference, i.e., 
\begin{equation}
\Delta^{(c)}_{\min} = \min_{\overset{j \in \mathcal{X}}{k_{1},k_{2} \in \{0\}\cup\mathcal{A}}} \{|c_{j,k_1} - c_{j,k_2}|\}.\nonumber
\end{equation}
Let $\xi_{j,k_1,k_2} = \frac{u_{j,k_1} - u_{j,k_2}}{c_{j,k_1} - c_{j,k_2}}$ for $j \in \mathcal{X}$, $k_1,k_2 \in \{0\}\cup\mathcal{A}$, and $k_1 \neq k_2$ (recall that $u_{j,0} = 0$ and $c_{j,0} = 0$ for the dummy action), $\bar{\xi}_{j,k_1,k_2}$ be its estimate at the end of the exploration stage, i.e.,  $\bar{\xi}_{j,k_1,k_2} = \frac{\bar{u}_{j,k_1} - \bar{u}_{j,k_2}}{c_{j,k_1} - c_{j,k_2}}$. Let $\Delta^{(\xi)}_{\min}$ be the minimal difference between any $\xi_{j_1, k_{11}, k_{12}}$ and $\xi_{j_2, k_{21}, k_{22}}$, i.e.,
\begin{equation}
\Delta^{(\xi)}_{\min} = \min_{\overset{j_1,j_2 \in \mathcal{X}}{ k_{11},k_{12},k_{21},k_{22} \in \{0\}\cup\mathcal{A}}} \{|\xi_{j_1, k_{11}, k_{12}} - \xi_{j_2, k_{21}, k_{22}}|\}.\nonumber
\end{equation}
Moreover, let $\pi_{\min} = \min_{j\in \mathcal{X}}\pi_j$ and let $\Delta^* = \Delta^{(c)}_{\min} \Delta^{(\xi)}_{\min}$.   Then, the following lemma states that under $\epsilon$-First ALP with a sufficiently large $\epsilon(T)$, the agent will obtain a correct ordering of $\xi_{j,k_1,k_2}$'s with high probability at the end of the exploration stage.
\begin{lemma}\label{thm:eps_first_estimate}
Let $0 < \delta < 1$. Under $\epsilon$-First ALP, if
\begin{equation}
\epsilon(T) = \bigg\lceil \frac{K}{(1-\delta)\pi_{\min}} + \log T \max\big\{\frac{1}{\delta^2}, \frac{16 K}{(1 - \delta)\pi_{\min}(\Delta^*)^2}\big\}\bigg\rceil, \nonumber
\end{equation}
then for any contexts $j_1,j_2 \in \mathcal{X}$, and actions $k_{11}, k_{12}, k_{21}, k_{22} \in \{0\}\cup\mathcal{A}$, if $\xi_{j_1,k_{11},k_{12}} < \xi_{j_2,k_{21},k_{22}}$, then at the end of the $\epsilon(T)$-th round, we have
\begin{equation}
\mathbb{P}\bigg\{\bar{\xi}_{j_1,k_{11},k_{12}} \geq \bar{\xi}_{j_2,k_{21},k_{22}} \bigg\} \leq (J + 4)T^{-2}.\nonumber
\end{equation}
Moreover, the agent ranks all the $\xi_{j,k_1,k_2}$'s correctly with probability no less than $1 - (4K + 1)JT^{-2}$.
\end{lemma}

\begin{proof}
We first analyze the number of executions for each context-action pair $(j,k)$ in the exploration stage.
Let $N_j = \sum_{t = 1}^{\epsilon(T)} \mathds{1}(X_t = j)$ be the number of occurrences of context $j$ up to round $\epsilon(T)$. Recall that the contexts $X_t$ arrive i.i.d.~in each round. Thus, using Hoeffding-Chernoff Bound for each context $j$, we have
\begin{align}
&\quad \mathbb{P}\bigg\{\forall j \in \mathcal{X}, N_j \geq (1-\delta) \pi_j \epsilon(T)\bigg\} \nonumber \\
& \geq 1 - \sum_{j = 1}^J \mathbb{P}\bigg\{N_j < (1-\delta) \pi_j \epsilon(T)\bigg\} \nonumber \\
& \geq 1 - J e^{-2\delta^2 \epsilon(T)} \nonumber \\
& \geq  1 - J e^{-2 \log T} \nonumber \\
& = 1 - J T^{-2}
\end{align}
On the other hand, the lower bound $(1-\delta) \pi_j \epsilon(T)\geq K + \frac{16K\log T}{(\Delta^*)^2}$. From the implementation of the exploration stage in Algorithm~\ref{alg:eps_first_alp}, we know that if $N_j \geq (1-\delta) \pi_j \epsilon(T)$, then
\begin{align}
C_{j,k} \geq \big\lfloor 1 + \frac{16\log T}{(\Delta^*)^2}\big\rfloor \geq \frac{16\log T}{(\Delta^*)^2},~~\forall k \in \mathcal{A}.
\end{align}
Therefore,
\begin{align}
&\quad \mathbb{P}\bigg\{\forall j \in \mathcal{X}, \forall k \in \mathcal{A}, C_{j,k} \geq  \frac{16\log T}{(\Delta^*)^2} \bigg\} \nonumber \\
& \geq 1 - J T^{-2}
\end{align}

Next, we study the relationship between the estimates $\bar{\xi}_{j_1,k_{11},k_{12}}$ and $\bar{\xi}_{j_2,k_{21},k_{22}} $ at the end of the exploration stage. We note that
\begin{align}
&\quad \bar{\xi}_{j_1,k_{11},k_{12}} \geq \bar{\xi}_{j_2,k_{21},k_{22}}  \nonumber \\
&\Leftrightarrow \big( \bar{\xi}_{j_1,k_{11},k_{12}} - \xi_{j_1,k_{11},k_{12}} - \frac{\xi_{j_2,k_{21},k_{22}}- \xi_{j_1,k_{11},k_{12}}}{2} \big) \nonumber\\
&~~- \big(\bar{\xi}_{j_2,k_{21},k_{22}} -\xi_{j_2,k_{21},k_{22}} + \frac{\xi_{j_2,k_{21},k_{22}}- \xi_{j_1,k_{11},k_{12}}}{2} \big) \geq 0 \nonumber\\
&\Leftrightarrow \big( \frac{\bar{u}_{j_1,k_{11}} - u_{j_1,k_{11}}}{c_{j_1,k_{11}} - c_{j_1,k_{12}}} - \frac{\xi_{j_2,k_{21},k_{22}}- \xi_{j_1,k_{11},k_{12}}}{4} \big) \nonumber\\
&\quad -\big(\frac{\bar{u}_{j_1,k_{12}} - u_{j_1,k_{12}}}{c_{j_1,k_{11}} - c_{j_1,k_{12}}} + \frac{\xi_{j_2,k_{21},k_{22}}- \xi_{j_1,k_{11},k_{12}}}{4} \big) \nonumber\\
& \quad - \big( \frac{\bar{u}_{j_2,k_{21}} - u_{j_2,k_{21}}}{c_{j_2,k_{21}} - c_{j_2,k_{22}}} + \frac{\xi_{j_2,k_{21},k_{22}}- \xi_{j_1,k_{11},k_{12}}}{4} \big) \nonumber\\
& \quad + \big(\frac{\bar{u}_{j_2,k_{22}} - u_{j_2,k_{22}}}{c_{j_2,k_{21}} - c_{j_1,k_{22}}} - \frac{\xi_{j_2,k_{21},k_{22}}- \xi_{j_1,k_{11},k_{12}}}{4} \big) \geq 0.\nonumber\\
\end{align}
Thus, for the event $\bar{\xi}_{j_1,k_{11},k_{12}} \geq \bar{\xi}_{j_2,k_{21},k_{22}}$ to be true, we require that at least one term (with the sign) in the last inequation above is no less than zero. Conditioned on $C_{j,k} \geq \frac{16\log T}{(\Delta^*)^2}$, we can bound the probability of each term according to the Hoeffding-Chernoff bound. For example, for the first term, we have
\begin{align}
&\quad \mathbb{P}\big\{ \frac{\bar{u}_{j_1,k_{11}} - u_{j_1,k_{11}}}{c_{j_1,k_{11}} - c_{j_1,k_{12}}} - \frac{\xi_{j_2,k_{21},k_{22}}- \xi_{j_1,k_{11},k_{12}}}{4} \geq 0 \nonumber \\
& \quad\quad\quad |C_{j_1,k_{11}} \geq \frac{16\log T}{(\Delta^*)^2} \big\} \nonumber \\
& \leq \mathbb{P}\big\{\bar{u}_{j_1,k_{11}} \geq u_{j_1,k_{11}}+  \frac{\Delta^*}{4} |C_{j_1,k_{11}} \geq \frac{16\log T}{(\Delta^*)^2}\big\}\nonumber \\
& \leq e^{-2\log T} = T^{-2}. \nonumber
\end{align}
The conclusion then follows by considering the event  $\big\{C_{j,k} \geq \frac{16\log T}{(\Delta^*)^2}, \forall j \in \mathcal{X}, \forall k \in \mathcal{X}\big\}$ and its negation.
\end{proof}

\begin{theorem}\label{thm:eps_first_alp}
Let $0 < \delta < 1$. Under $\epsilon$-First ALP, if
\begin{equation}
\epsilon(T) \geq \frac{K}{(1-\delta)\pi_{\min}} + \log T \max\bigg\{\frac{1}{\delta^2}, \frac{16 K}{(1 - \delta)\pi_{\min}(\Delta^*)^2}\bigg\}, \nonumber
\end{equation}
then the regret of $\epsilon$-First ALP satisfies:\\
1) if $\rho = B/T \neq Q_i$, then $R_{\epsilon-\rm First ALP}(T,B) = O(\log T)$;

2) if $\rho = B/T = Q_i$, then $R_{\epsilon-\rm First ALP}(T,B) = O(\sqrt{T})$.
\end{theorem}

\proof{(Sketch) The key idea of proving this theorem is considering the event where the $\xi_{j,k_1,k_2}$'s are ranked correctly and its negation. When the $\xi_{j,k_1,k_2}$'s are ranked correctly, we can use the properties of the ALP algorithm with modification on the time horizon and budget (subtracting the time and budget in the exploration stage, which is $O(\log T)$); otherwise, if the agent obtains a wrong ranking results, the regret is bounded as $O(1)$ because the probability is $O(T^{-2})$ and the reward in each round is bounded. }

\subsection{Deciding $\epsilon(T)$ without Prior Information}
In Theorem~\ref{thm:eps_first_alp}, the agent requires the value of $\Delta^*$ (in fact $\Delta^{(\xi)}_{\min}$ because $\Delta^{(c)}_{\min}$ is known) to calculate $\epsilon(T)$. This is usually impractical since the expected rewards are unknown {\it a priori}.
Thus, without the knowledge of  $\Delta^{(\xi)}_{\min}$, we propose a Confidence Level Test (CLT) algorithm  for deciding when to end the exploration stage.

Specifically, assume $\Delta^{(\xi)}_{\min} > 0$ and is unknown by the agent. In each round of the exploration stage, the agent tries to solve the problem $({\mathcal{LP}}'_{\tau,b})$ with $u_{j,k}$ replaced by $\bar{u}_{j,k}$ using comparison, i.e., using Algorithm~\ref{alg:find_candidates} and sorting the virtual actions. For each comparison, the agent tests the confidence level according to Algorithm~\ref{alg:clt}. If all comparisons pass the test, i.e., \texttt{flagSucc = true} for all comparisons, then the agent ends the exploration stage and starts the exploitation stage.

\begin{algorithm}[htbp]
\caption{Confidence Level Test (CLT)}
\label{alg:clt}
\begin{algorithmic}
\STATE{\bfseries Input:} Time horizon $T$, estimates $\bar{\xi}_{j_1,k_{11}, k_{12}}$, $\bar{\xi}_{j_2,k_{21}, k_{22}}$, number of executions $C_{j_1,k_{11}}$, $C_{j_1,k_{12}}$, $C_{j_2,k_{21}}$, and $C_{j_2,k_{22}}$;\\
\STATE{\bfseries Output:} \texttt{flagSucc};\\
\STATE{\bfseries Init:} \texttt{flagSucc = false}, \\~~~~~~~~$\Delta' = \frac{\Delta^{(c)}_{\min}(\bar{\xi}_{j_1,k_{11}, k_{12}} - \bar{\xi}_{j_2,k_{21}, k_{22}})}{2}$;
\IF{$e^{-2(\Delta')^2 \min\{C_{j_1,k_{11}}, C_{j_1,k_{12}}\}} \leq T^{-2}$ \& $e^{-2(\Delta')^2 \min\{C_{j_2,k_{21}}, C_{j_2,k_{22}}\}} \leq T^{-2}$}
\STATE{\texttt{flagSucc = true};}
\ENDIF
\STATE{\bfseries return} \texttt{flagSucc};
\end{algorithmic}
\end{algorithm}

Next, we show that the $\epsilon$-First policy with CLT will achieve $O(\log T)$ regret except for the boundary cases, where it achieves $O(\sqrt{T})$ regret. On one hand, according to Hoeffding-Chernoff bound, if all comparisons pass the confidence level test, then with probability at least $1 - JK^2 T^{-2}$, the algorithm obtains the correct rank and provide a right solution for the problem $({\mathcal{LP}}'_{\tau,b})$. On the other hand, because $\Delta^* > 0$, from the analysis in the previous section, we know that the exploration stage will end within $O(\log T)$ rounds with high probability. Therefore, the expected regret is the same as that in the case with known $\Delta^{(\xi)}_{\min}$. 

\section{Numerical Experiments} \label{sec:sim_results}

In this section, we evaluate the regret of the proposed algorithms through numerical simulations.
We study the performance of the proposed algorithms here for unit-cost systems as the parameter setting is relatively simple to control while providing  us useful insights. The performance in heterogeneous-cost systems is similar as we have shown theoretically, and omitted here.
In the case with known statistics, we compare the proposed PB (two-context case) and ALP algorithms with Fixed LP (FLP) algorithm that uses a fixed average budget constraint $B/T$ since both \cite{Badanidiyuru2014COLT} and \cite{Agrawal2014EC} use fixed average budget constraint. Then, the UCB-based FLP, i.e., UCB-FLP, is evaluated in the case without knowledge of expected rewards. We also evaluate algorithms for the case without knowledge of context distribution. When the context distribution is unknown to the agent, we use the Empirical ALP (EALP) algorithm, that uses the empirical distribution (histogram) of context for making decisions, in the case with known expected rewards. Then, the UCB-based EALP is proposed for the case without knowledge of expected rewards.
The results are averaged from 5,000 independent runs of the simulations.

\subsection{Two-Context Systems}

\begin{figure*}[thbp]
\begin{center}
    \begin{minipage}[b]{1.0\linewidth}
        \begin{center}
        \subfigure[]{\includegraphics[angle = 0,width = 0.32\linewidth]{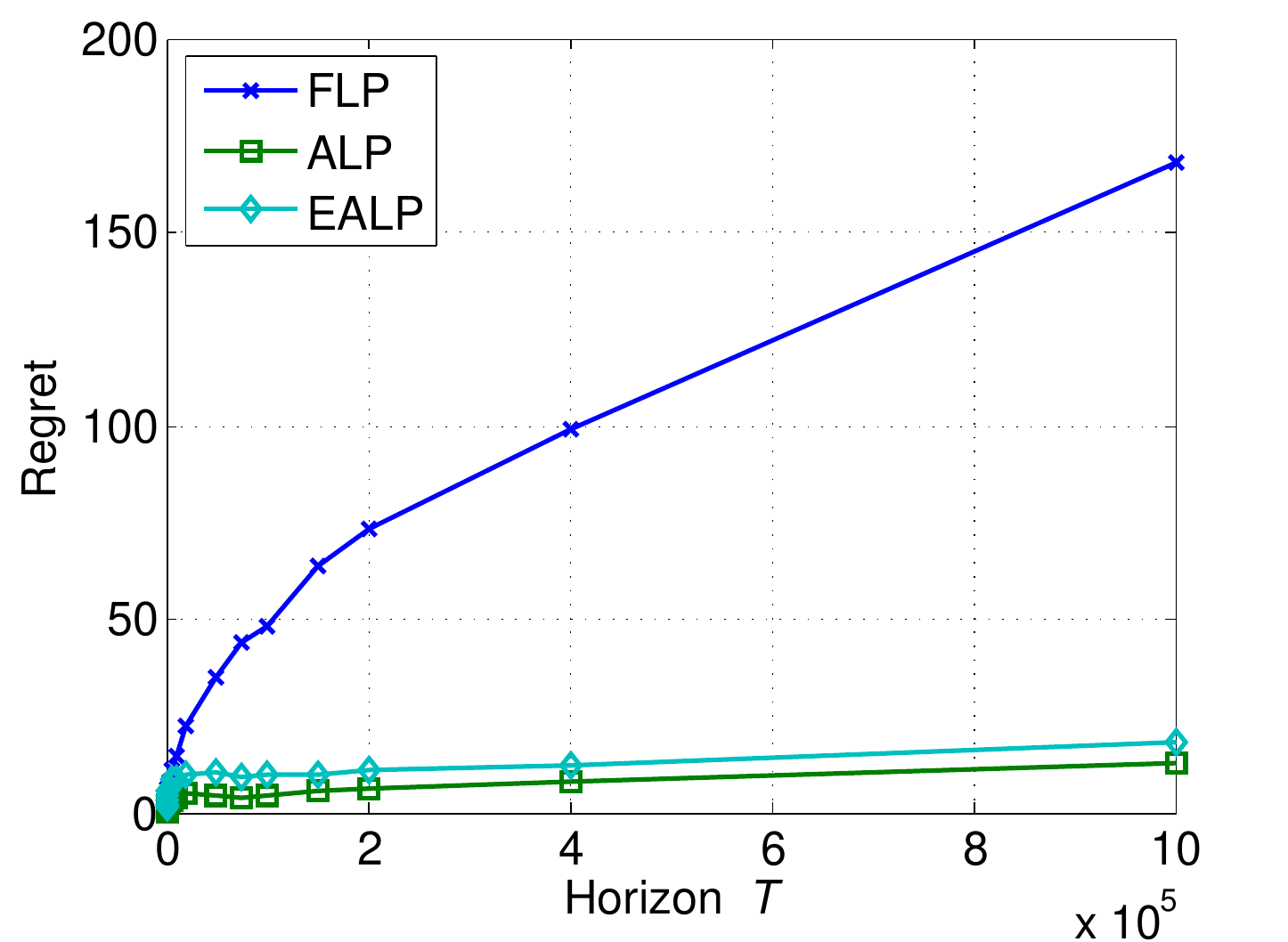}
        \label{fig:twocont_regret_known_rho0p39}}
        \subfigure[]{\includegraphics[angle = 0,width = 0.32\linewidth]{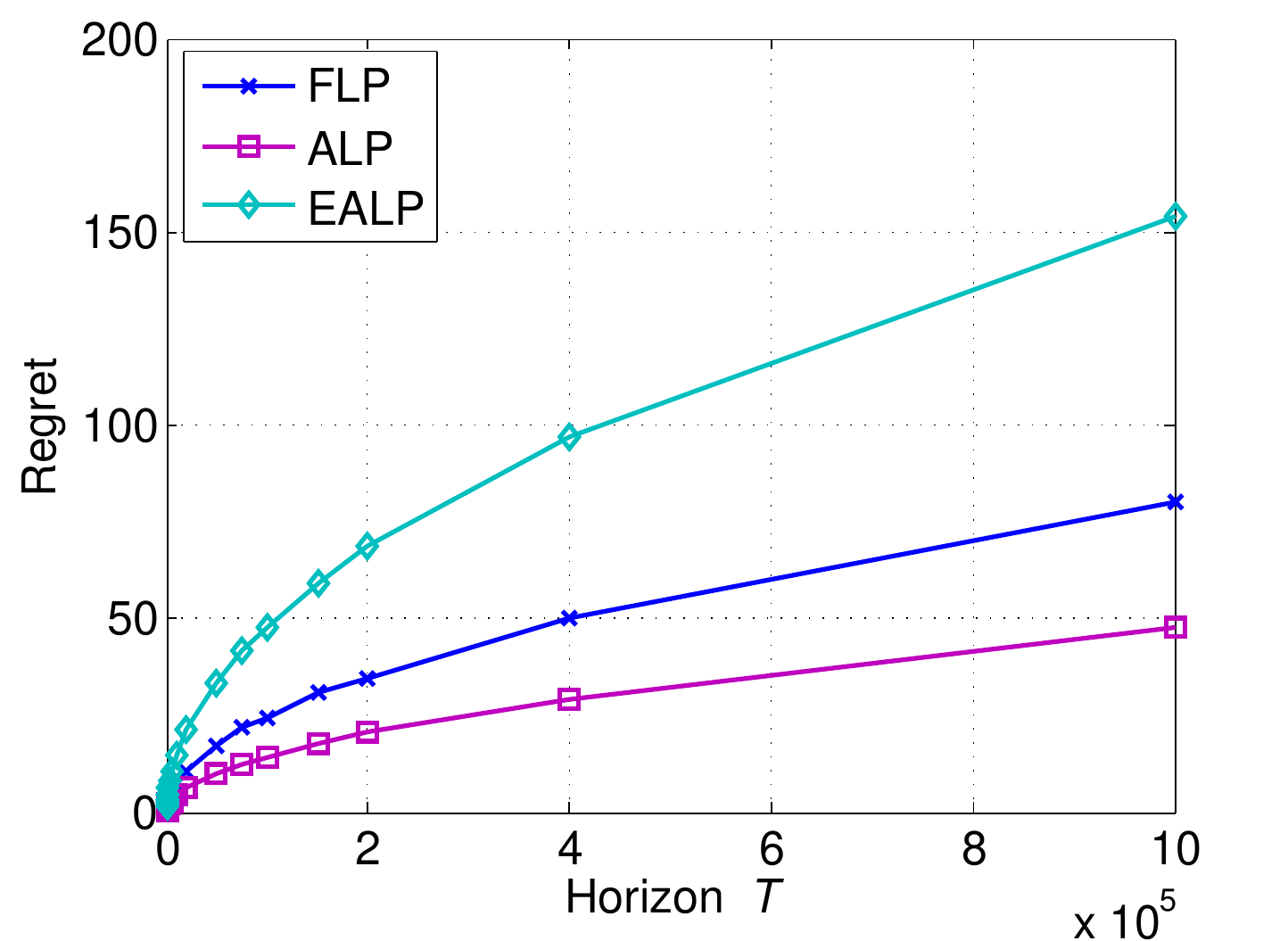}
        \label{fig:twocont_regret_known_rho0p40}}
        \subfigure[]{\includegraphics[angle = 0,width = 0.32\linewidth]{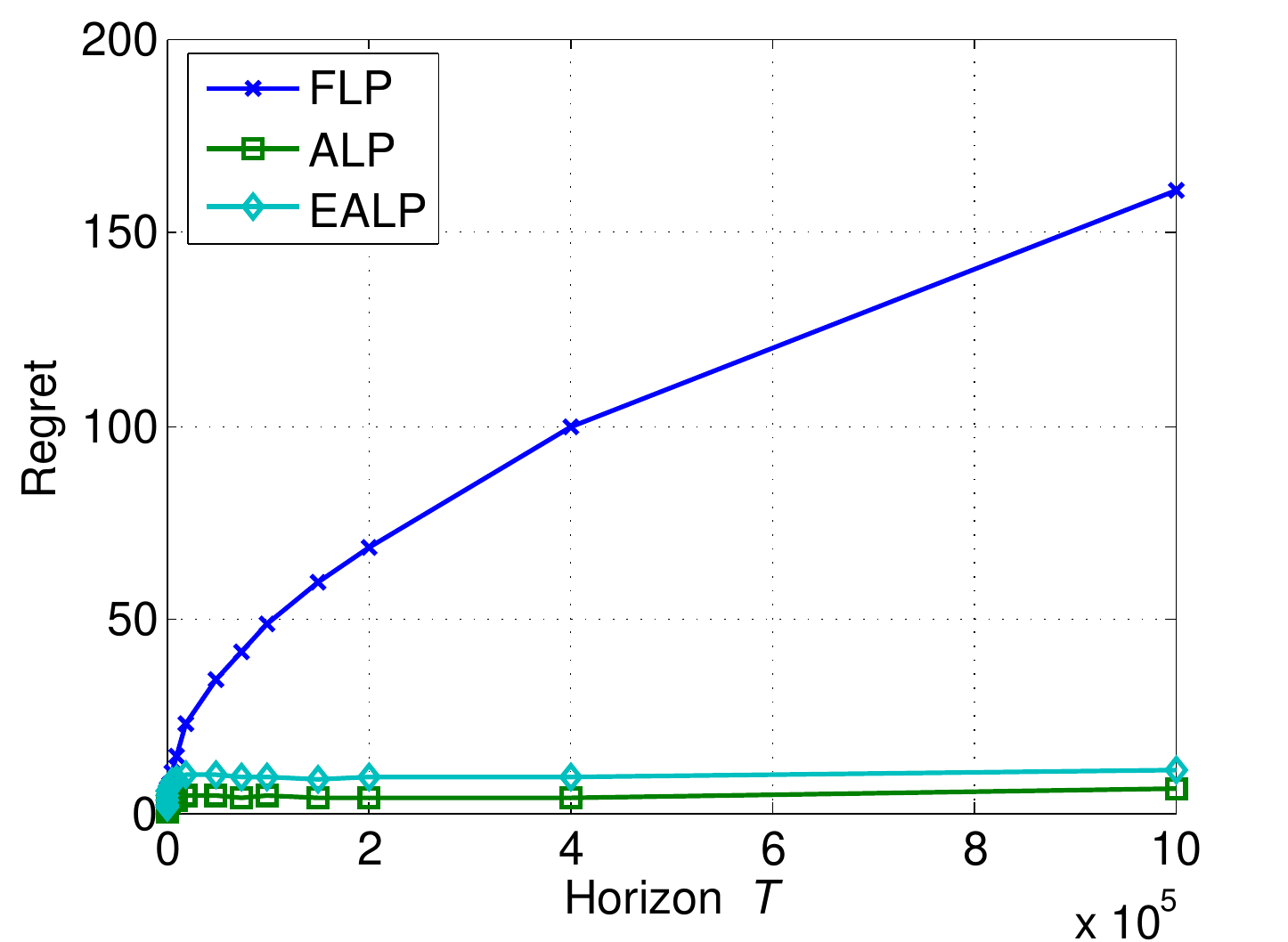}
        \label{fig:twocont_regret_known_rho0p41}}
        \caption{Comparison of algorithms for the two-context systems with perfect knowledge ($\pi_1 = 0.4, \pi_2 = 0.6$), (a) $\rho = 0.39$,
        (b) $\rho = 0.4$,
        (c) $\rho = 0.41$.}
        \label{fig:twocont_known}
        \end{center}
    \end{minipage}
\end{center}
\vspace{-0.5cm}
\end{figure*}

We first consider a two-context scenario with $K = 3$ arms and Bernoulli rewards: the context distribution vector is $\boldsymbol{\pi} = [0.4, 0.6]$, the expected rewards are $\boldsymbol{u}_1 = 0.8 \times [1/3,2/3,1]$ for context 1, and $\boldsymbol{u}_2 = 0.4 \times [1/3,2/3,1]$  for context 2. The boundary is $q_1 = \pi_1 = 0.4$ and we study the cases with normalized budget $\rho = 0.39, 0.4$, and $0.41$, respectively.

Figure~\ref{fig:twocont_known} shows the regret of different algorithms in the case with known expected rewards. In the non-boundary cases (i.e., $\rho = 0.39, 0.41$), the ALP algorithm achieves near optimal performance. Even without the knowledge of context distribution, the EALP algorithm performs much better than FLP. In the boundary case, i.e., $\rho = 0.4$, the regret of ALP increases with $T$ but is still lower than that of FLP. The EALP algorithm achieves higher regret than ALP and FLP due to the empirical distribution errors.

\begin{figure*}[thbp]
\begin{center}
    \begin{minipage}[b]{1.0\linewidth}
        \begin{center}
        \subfigure[]{\includegraphics[angle = 0,width = 0.32\linewidth]{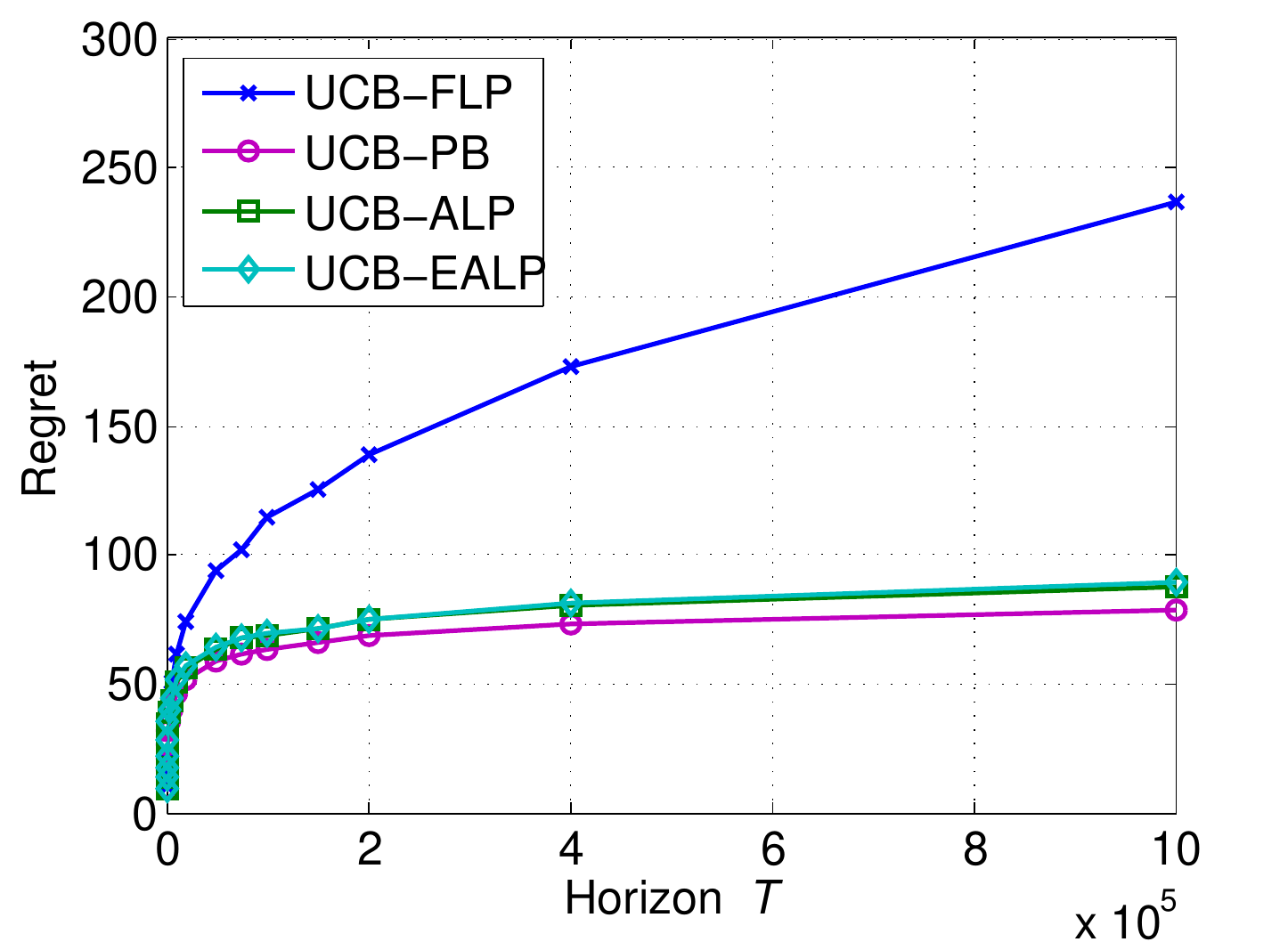}
        \label{fig:twocont_regret_unknown_rho0p39}}
        \subfigure[]{\includegraphics[angle = 0,width = 0.32\linewidth]{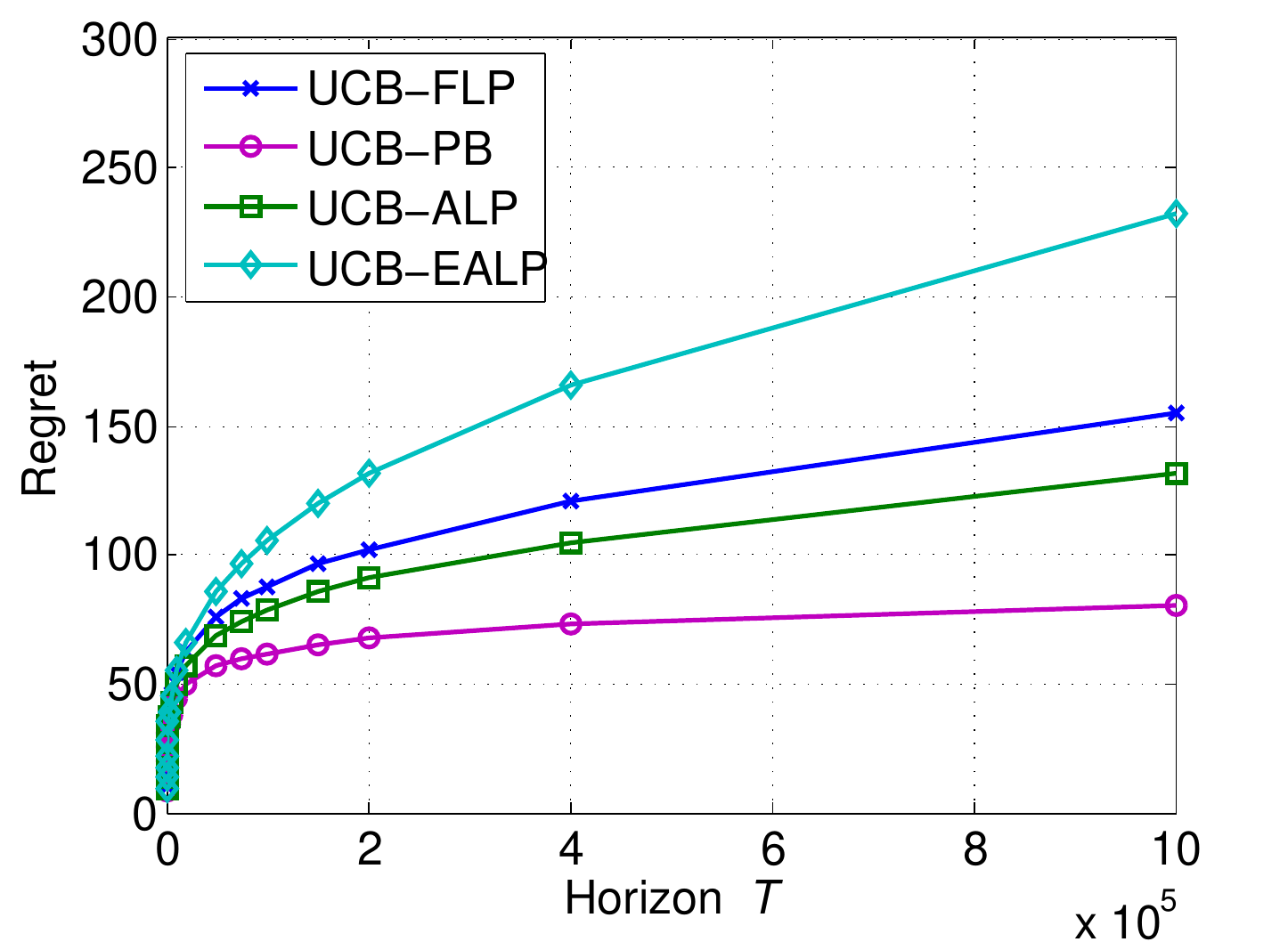}
        \label{fig:twocont_regret_unknown_rho0p40}}
        \subfigure[]{\includegraphics[angle = 0,width = 0.32\linewidth]{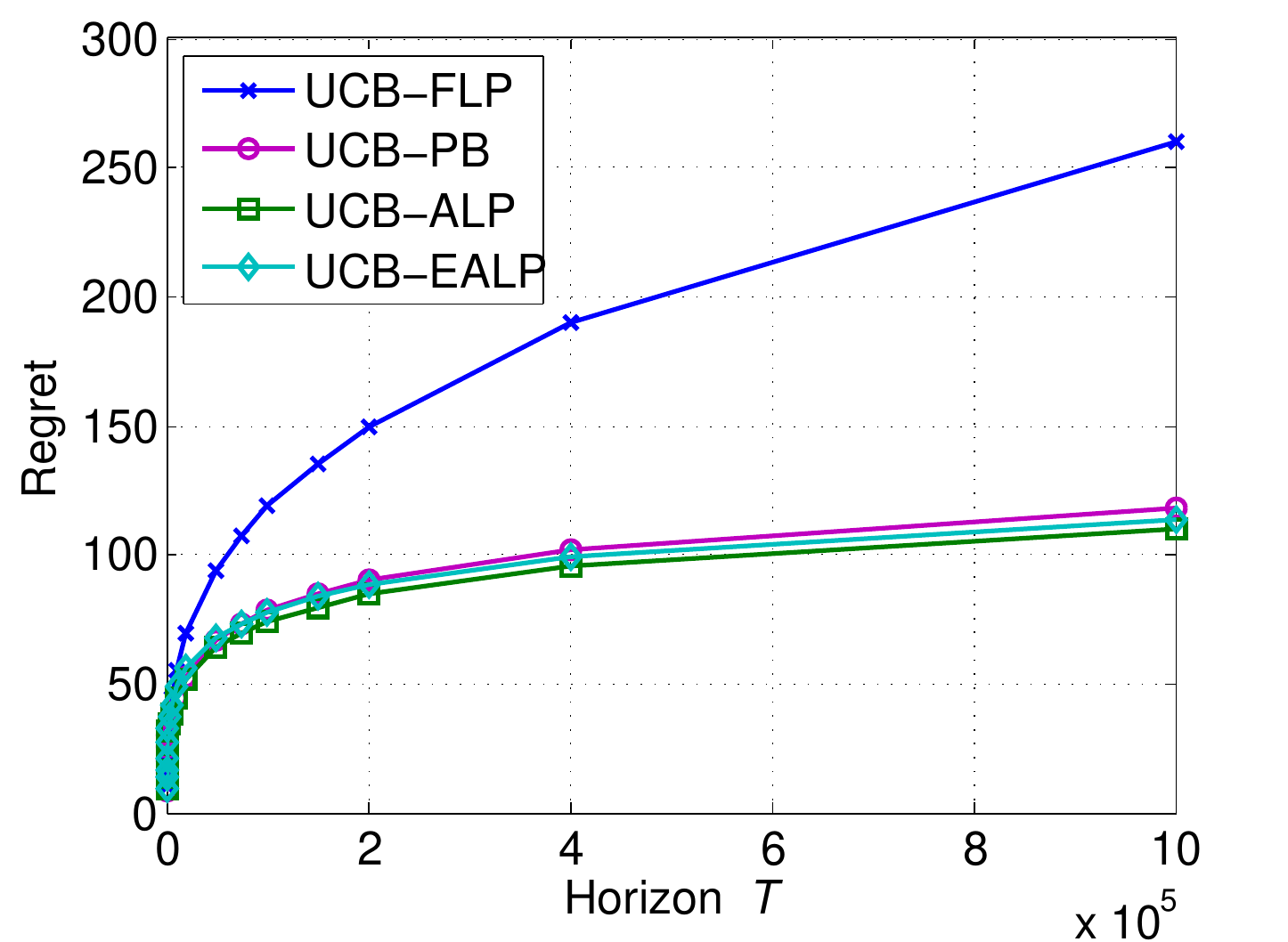}
        \label{fig:twocont_regret_unknown_rho0p41}}
        \caption{Comparison of algorithms for the two-context systems without perfect knowledge ($\pi_1 = 0.4, \pi_2 = 0.6$), (a) $\rho = 0.39$,
        (b) $\rho = 0.4$,
        (c) $\rho = 0.41$. }
        \label{fig:twocont_unknown}
        \end{center}
    \end{minipage}
\end{center}
\vspace{-0.5cm}
\end{figure*}

Figure~\ref{fig:twocont_unknown} shows the regret of different algorithms in the case without knowledge of expected rewards. We can see that in the non-boundary cases, UCB-ALP and UCB-EALP achieves regret that is very close to UCB-PB and outperforms UCB-FLP. Interestingly, we can even see that UCB-ALP achieves slightly lower regret than UCB-PB in the case with $\rho = 0.41$. This is because under UCB-PB, the better context may be skipped and wasted if it does not have the highest UCB. In contrast, the UCB-ALP algorithm may allocate certain resource to the better context, even when it does not have the highest UCB. On the boundary case, the regrets of UCB-ALP  and UCB-EALP become larger than that of UCB-PB, but are still sublinear in $T$.

\subsection{Multi-Context Systems}
Next, we study a multi-context scenario with $J = 10$ contexts, $K = 5$ arms, and Bernoulli rewards. Specifically,  the context distribution vector is $\boldsymbol{\pi} = [0.025, 0.05, 0.075, 0.15, 0.2, 0.2, 0.15, 0.075,  0.05, 0.025]$. The expected reward of action $k$ under context $j$ is ${u}_{j,k} =\frac{jk}{JK}$. One boundary in this system is $q_5 = 0.5$. We study the cases with average budget $\rho = 0.49, 0.5$, and $0.51$, respectively. In this case, it is difficult to calculate the expected total reward obtained by the oracle solution. Thus, we calculate the regret by comparing with the upper bound, i.e., $\widehat{U}(T,B) = Tv(\rho)$.

\begin{figure*}[thbp]
\begin{center}
    \begin{minipage}[b]{1.0\linewidth}
        \begin{center}
        \subfigure[]{\includegraphics[angle = 0,width = 0.32\linewidth]{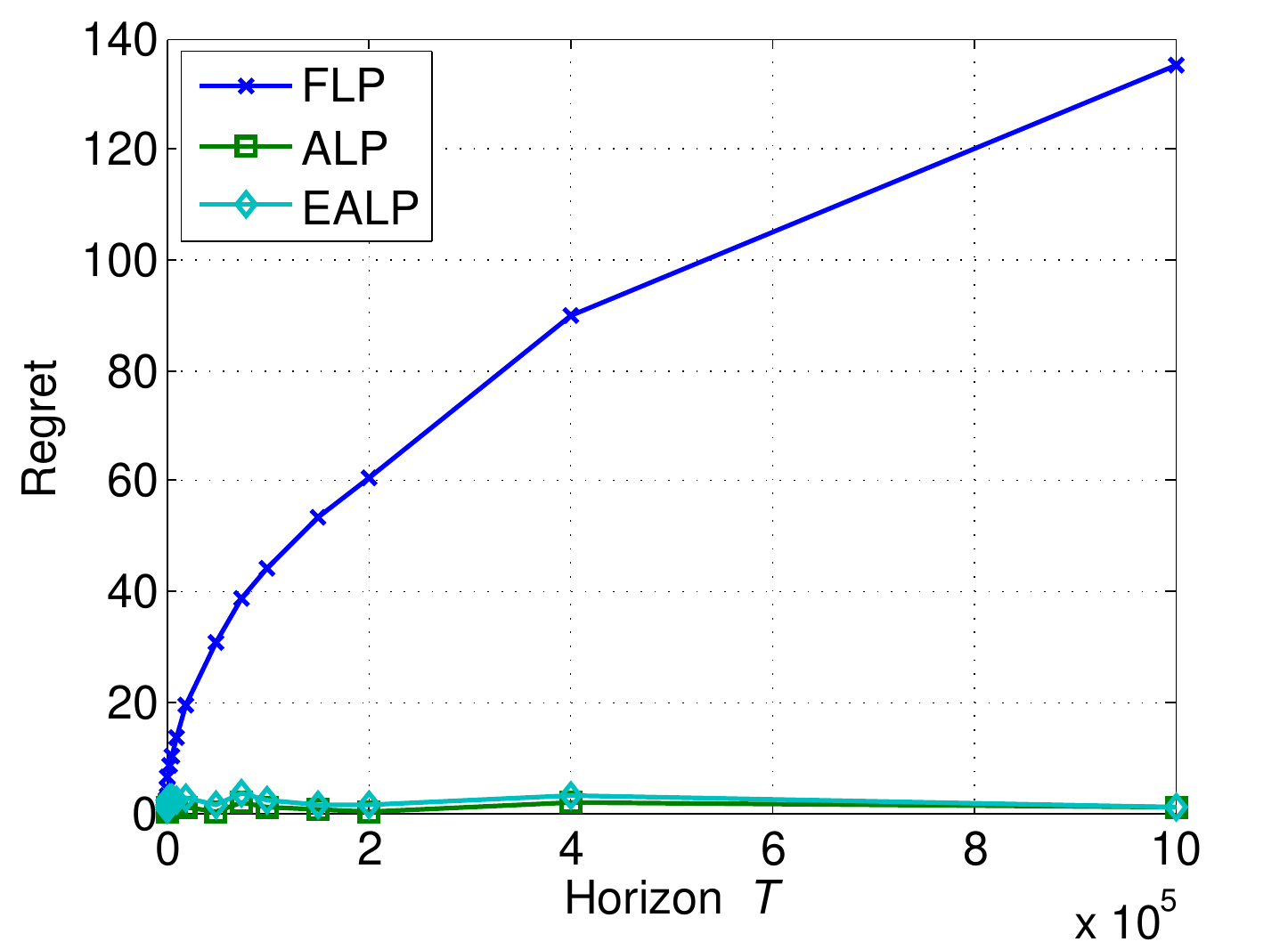}
        \label{fig:multicont_regret_known_rho0p49}}
        \subfigure[]{\includegraphics[angle = 0,width = 0.32\linewidth]{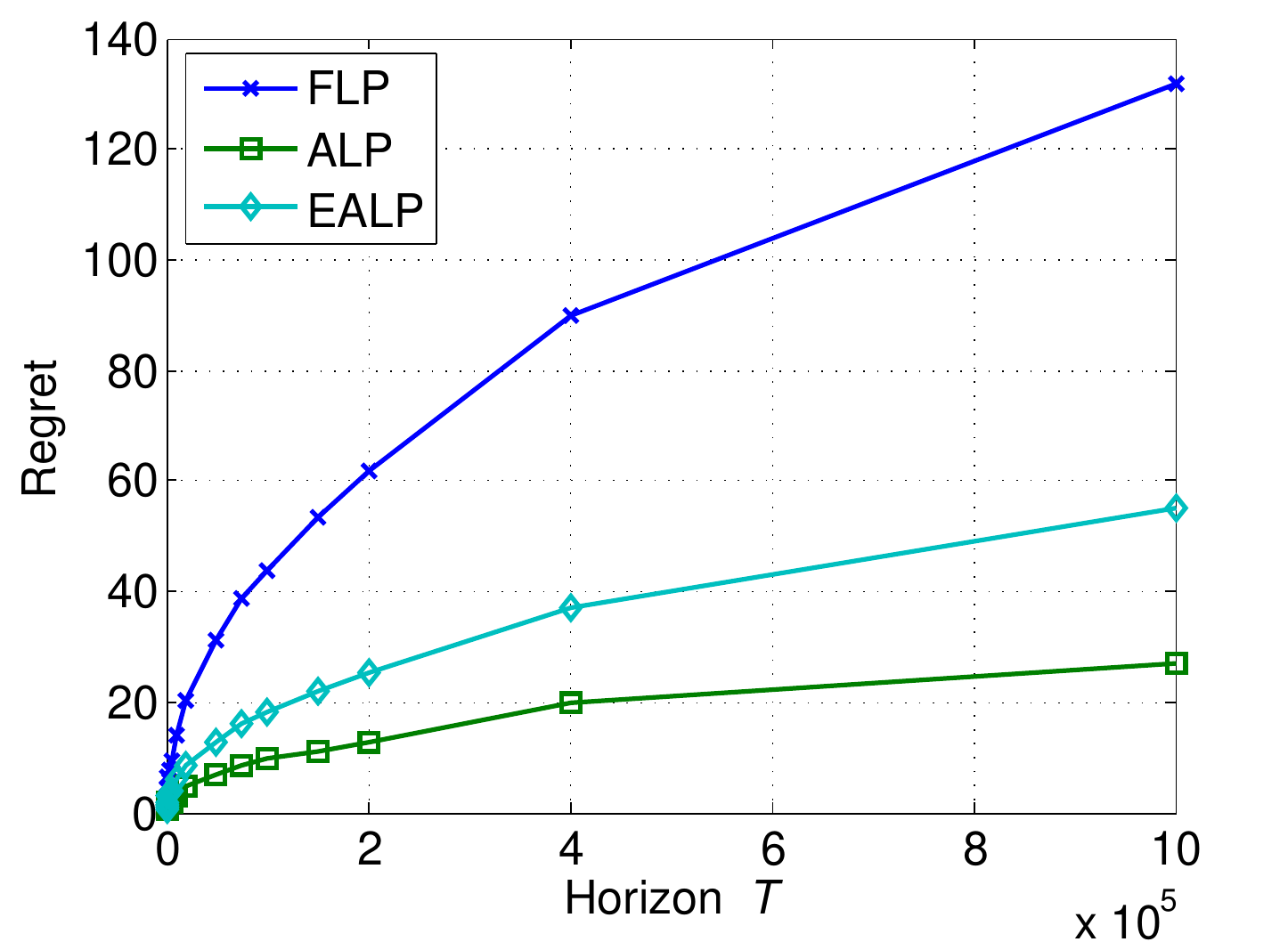}
        \label{fig:multicont_regret_known_rho0p50}}
        \subfigure[]{\includegraphics[angle = 0,width = 0.32\linewidth]{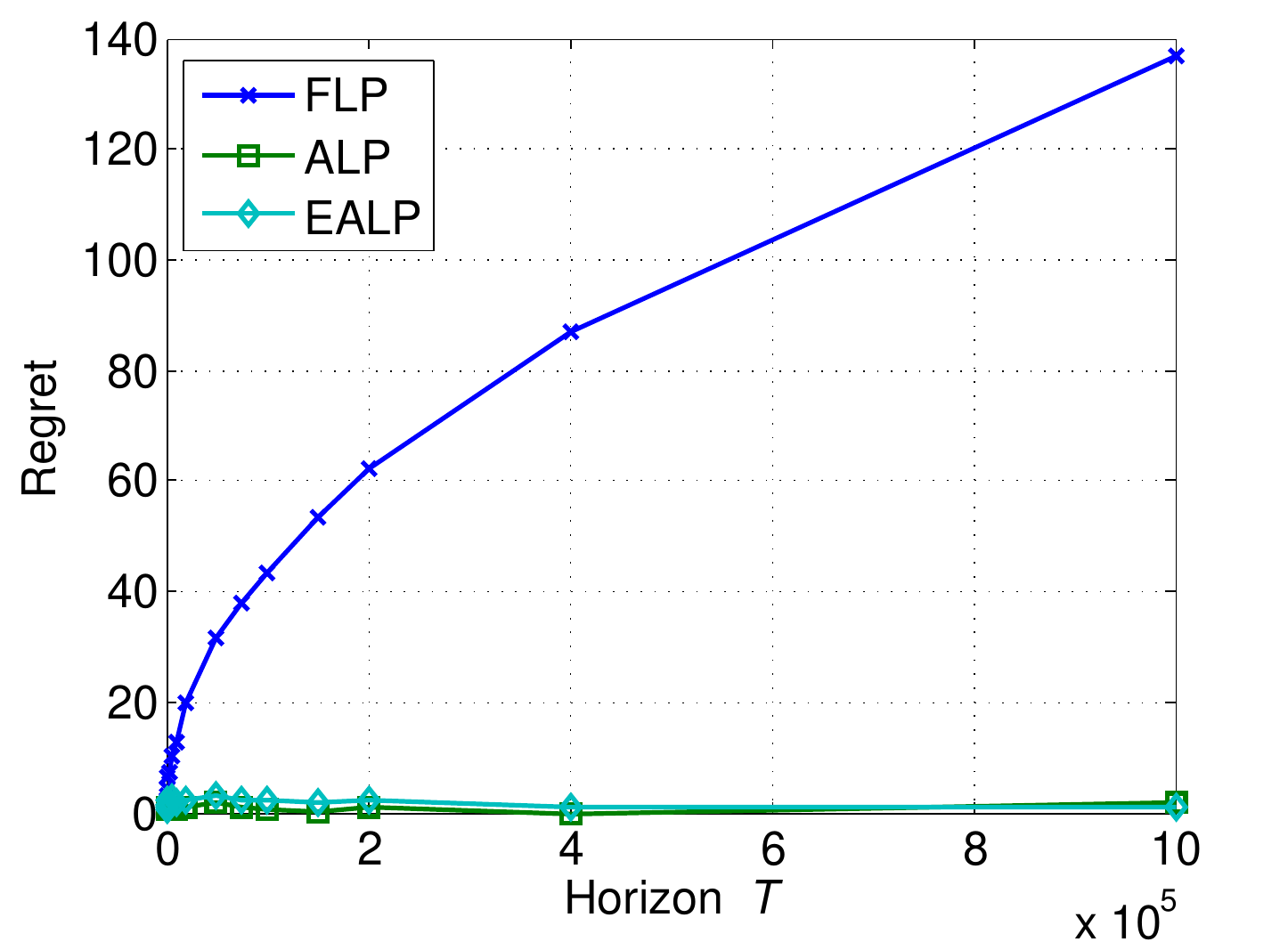}
        \label{fig:multicont_regret_known_rho0p51}}
        \caption{Comparison of algorithms for the multi-context systems with perfect knowledge ($Q_5 = 0.5$), (a) $\rho = 0.49$,
        (b) $\rho = 0.5$,
        (c) $\rho = 0.51$.}
        \label{fig:multicont_known}
        \end{center}
    \end{minipage}
\end{center}
\vspace{-0.5cm}
\end{figure*}

Figure~\ref{fig:multicont_known} shows the regret of different algorithms in the case with known expected rewards.
In the non-boundary cases, both the ALP and EALP algorithm  achieve similar performance as in the two-context case. The regret of EALP is even lower than FLP in the boundary case, since the ratio of contexts that are executed with correct probability is higher than that in the two-context systems.

\begin{figure*}[thbp]
\begin{center}
    \begin{minipage}[b]{1.0\linewidth}
        \begin{center}
        \subfigure[]{\includegraphics[angle = 0,width = 0.32\linewidth]{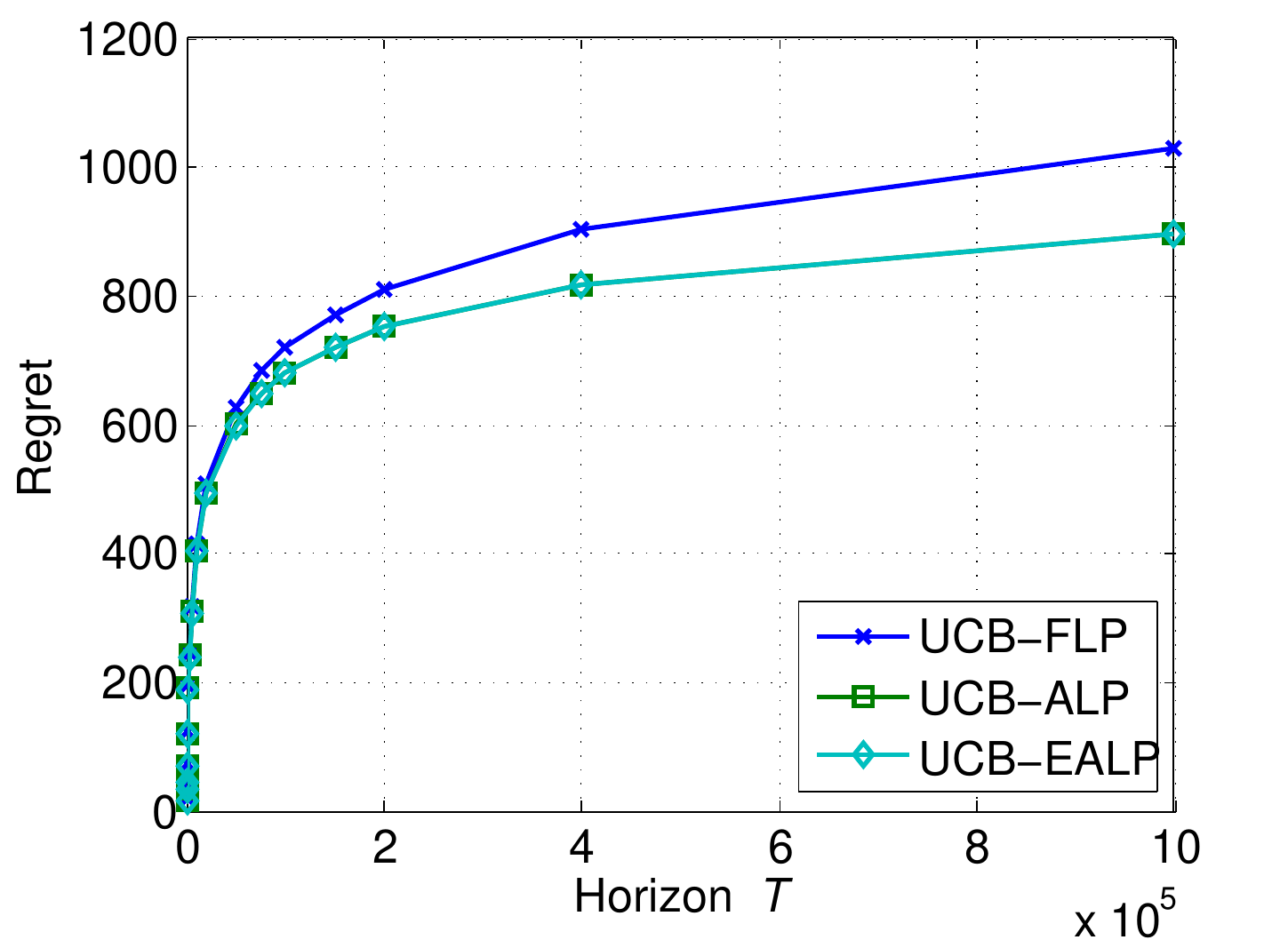}
        \label{fig:multicont_regret_unknown_rho0p49}}
        \subfigure[]{\includegraphics[angle = 0,width = 0.32\linewidth]{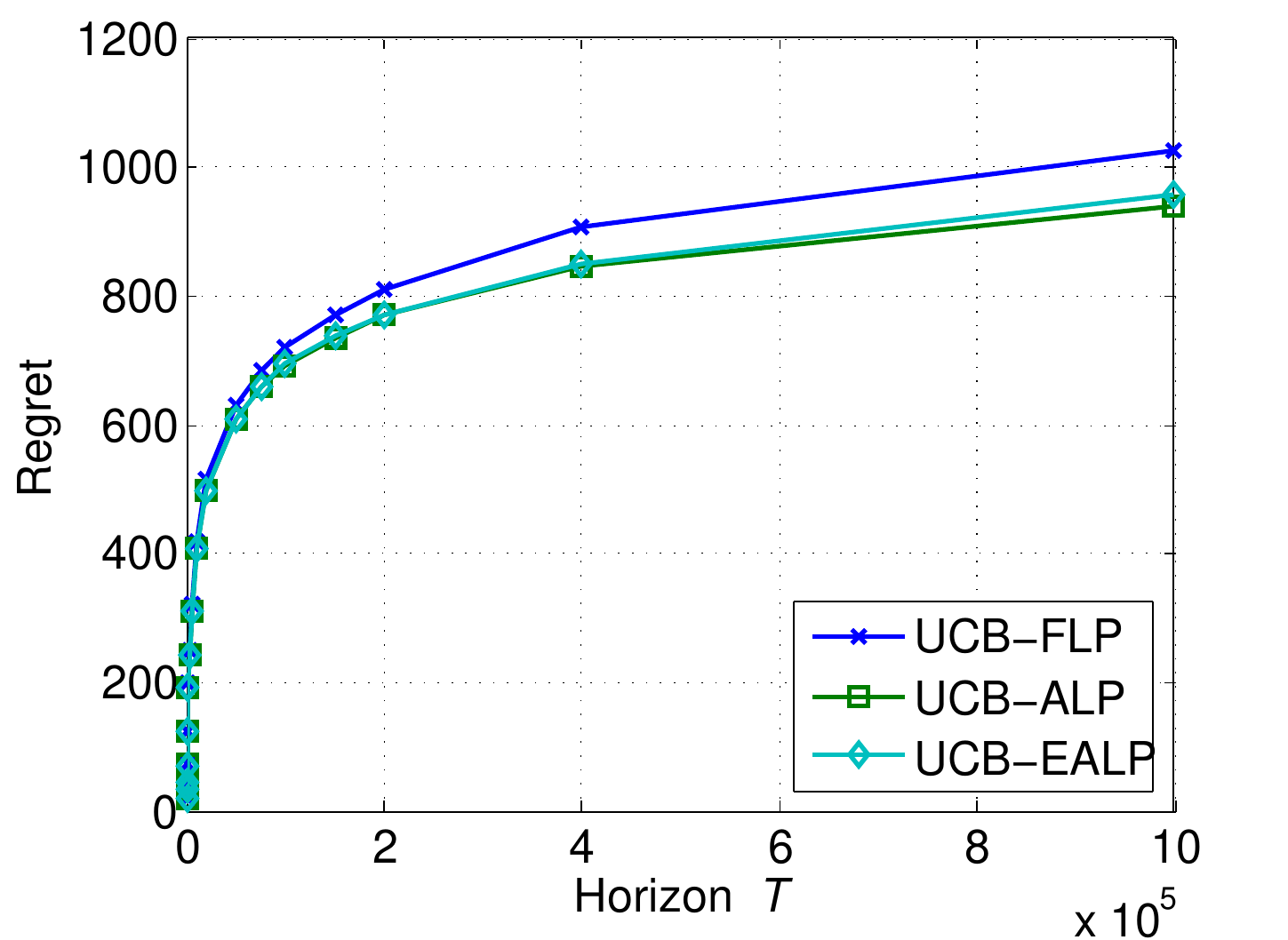}
        \label{fig:multicont_regret_unknown_rho0p50}}
        \subfigure[]{\includegraphics[angle = 0,width = 0.32\linewidth]{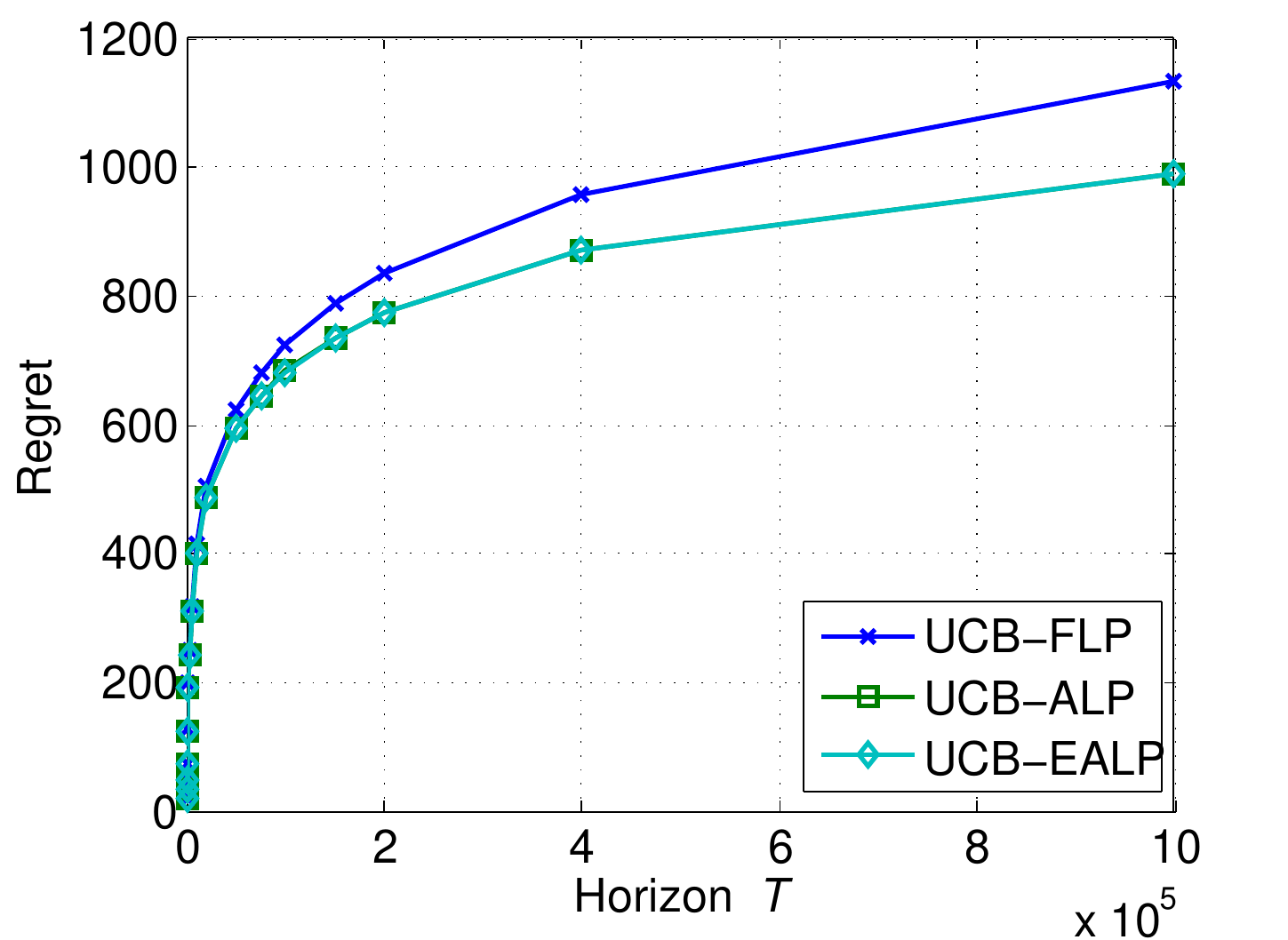}
        \label{fig:multicont_regret_unknown_rho0p51}}
        \caption{Comparison of algorithms for the multi-context systems without perfect knowledge ($Q_5 = 0.5$), (a) $\rho = 0.49$,
        (b) $\rho = 0.5$,
        (c) $\rho = 0.51$.}
        \label{fig:multicont_unknown}
        \end{center}
    \end{minipage}
\end{center}
\vspace{-0.5cm}
\end{figure*}
Figure~\ref{fig:multicont_unknown} shows the regret of different algorithms in the case without  knowledge of expected rewards.
We can see that all algorithms achieve sublinear regret, but the difference between the non-boundary cases and the boundary case is small. This is rooted in the fact that when the number of contexts and the number of actions are large, it requires more time to learn the expected rewards. Hence, the constant in the $\log T$ term is much larger than that in the $\sqrt{T}$ term, and the  $\log{T}$ term dominates the regret and the impact of the $\sqrt{T}$ term could be small. Exploring the structure of the reward function in contextual bandits, e.g., similarity \cite{Slivkins2014JMLR:ContMAB} and linearity \cite{Li2010WWW:LinUCB}, to reduce the exploration time is part of our future work.

\end{document}